\documentclass[english,final,3p,times]{elsarticle}

\usepackage{amsmath,amsfonts,amssymb,epsfig}
\newcommand{\argmin}[1]{\underset{#1}{\mbox{arg}\min}}
\usepackage{subfigure}
\usepackage{makeidx}  
\usepackage{url}      
\usepackage[latin1]{inputenc}
\usepackage{babel}

\usepackage{graphicx}
\usepackage{algorithm}
\usepackage{algpseudocode}
\usepackage{array}

\newcommand{\conv}{\ast\mathcal{G}_\sigma}
\renewcommand{\choose}[2]{\genfrac{(}{)}{0pt}{}{#1}{#2}}

\begin{document}

\begin{frontmatter}


\title{Prostate biopsy tracking with deformation estimation}
\author[timc,koelis]{Michael Baumann}
\ead{baumann@koelis.com}
\author[pitie]{Pierre Mozer}
\author[koelis]{Vincent Daanen}
\author[timc]{Jocelyne Troccaz}
\ead{jocelyne.troccaz@imag.fr}
\address[timc]{Université J.~Fourier, TIMC Laboratory, CNRS, UMR 5525, Grenoble, France }
\address[pitie]{La Pitié-Salpêtrière Hospital, Urology Dpt., 75651 Paris Cedex 13, France}
\address[koelis]{Koelis SAS, 5. av. du Grand Sablon, 38700 La Tronche, France}

\begin{abstract}
Transrectal biopsies under 2D ultrasound (US) control are the current clinical standard for prostate cancer diagnosis. The isoechogenic nature of prostate carcinoma makes it necessary to sample the gland systematically, resulting in a low sensitivity. Also, it is difficult for the clinician to follow the sampling protocol accurately under 2D US control and the exact anatomical location of the biopsy cores is unknown after the intervention. Tracking systems for prostate biopsies make it possible to generate biopsy distribution maps for intra- and post-interventional quality control and 3D visualisation of histological results for diagnosis and treatment planning. They can also guide the clinician toward non-ultrasound targets. In this paper, a volume-swept 3D US based tracking system for fast and accurate estimation of prostate tissue motion is proposed. The entirely image-based system solves the patient motion problem with an a priori model of rectal probe kinematics. Prostate deformations are estimated with elastic registration to maximize accuracy. The system is robust with only 17 registration failures out of 786 (2$\%$) biopsy volumes acquired from 47 patients during biopsy sessions. Accuracy was evaluated to 0.76$\pm$0.52 mm using manually segmented fiducials on 687 registered volumes stemming from 40 patients. A clinical protocol for assisted biopsy acquisition was designed and implemented as a biopsy assistance system, which allows to overcome the draw-backs of the standard biopsy procedure.
\end{abstract}

\begin{keyword}
image-based tracking \sep prostate tracking \sep prostate biopsy maps \sep guided prostate biopsies \sep 3D ultrasound registration




\end{keyword}

\end{frontmatter}


\section{Introduction}

\subsection{Prostate biopsies}
\begin{figure}
\centering
\subfigure[]{\includegraphics[width=0.4\textwidth]{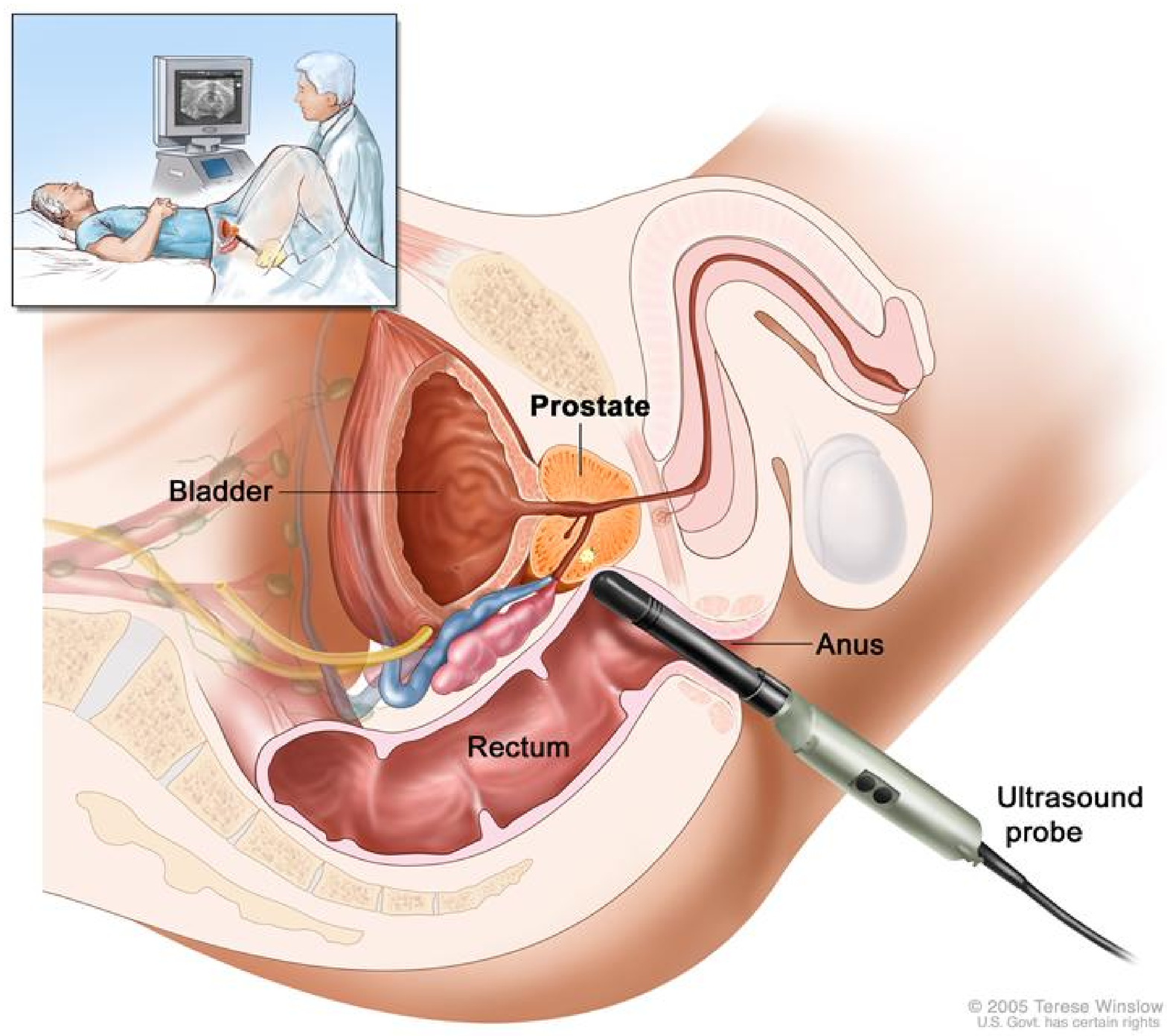}}\hfill
\subfigure[]{\includegraphics[width=0.28\textwidth]{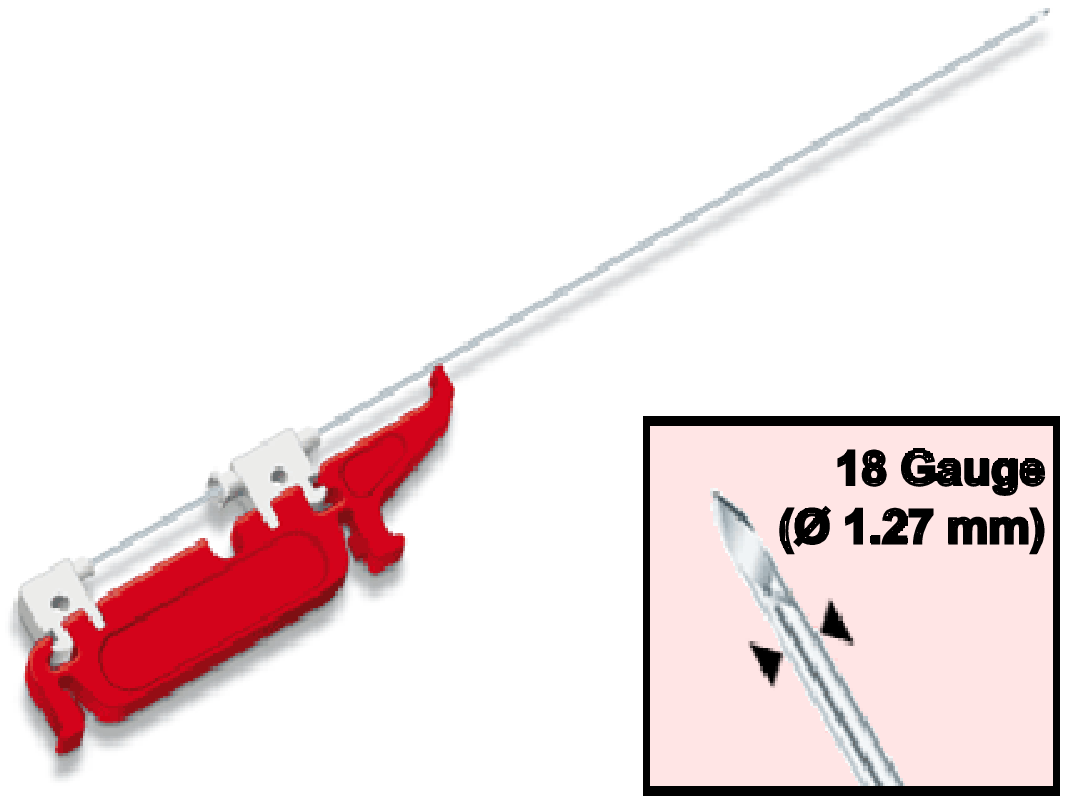}}\hfill
\subfigure[]{\includegraphics[width=0.28\textwidth]{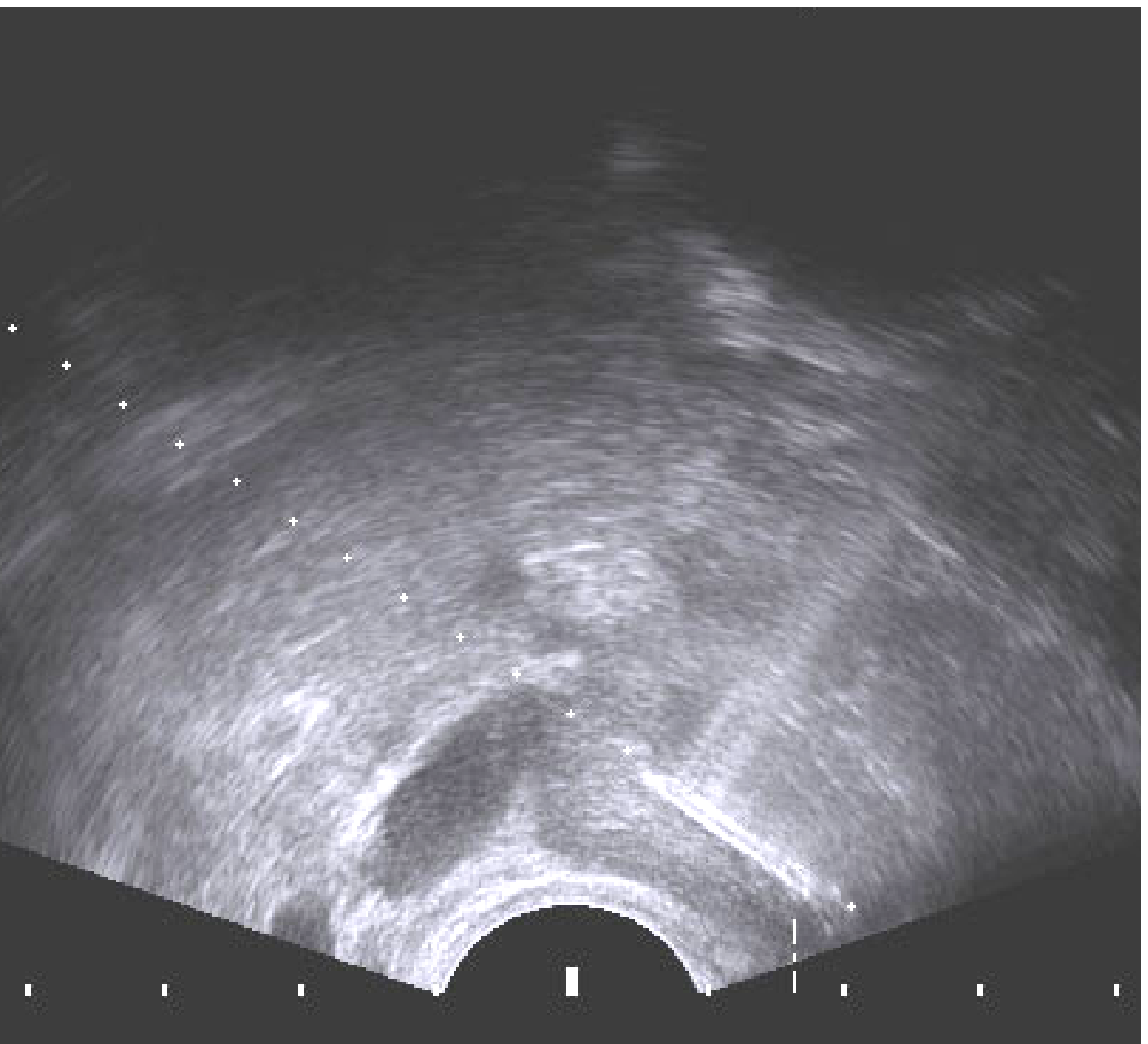}}
\caption{2D-TRUS guided prostate biopsies. The patient is in dorsal or lateral decubitus and locally anesthetized. The 2D TRUS probe is inserted into the rectum to position the needle. A rigidly attached needle guide ensures that the puncture trajectory lies in the US plane. Fig.~(b) illustrates an 18 Gauge (diameter of 1.27 mm) biopsy needle spring gun. Fig.~(c) shows the 2D TRUS image containing the needle (bright line with reflection artefacts) and the virtual puncture trajectory known from guide calibration (dotted line). Fig.~(a) was found on the web site of the National Cancer Institute (\textit{www.cancer.gov}). Fig.~(c) was found on the web site of BK Medical (\textit{www.bkmed.com}).}
\label{fig:clinicalcontext}
\end{figure}

\begin{figure}
\centering
\subfigure[]{\includegraphics[width=0.24\textwidth]{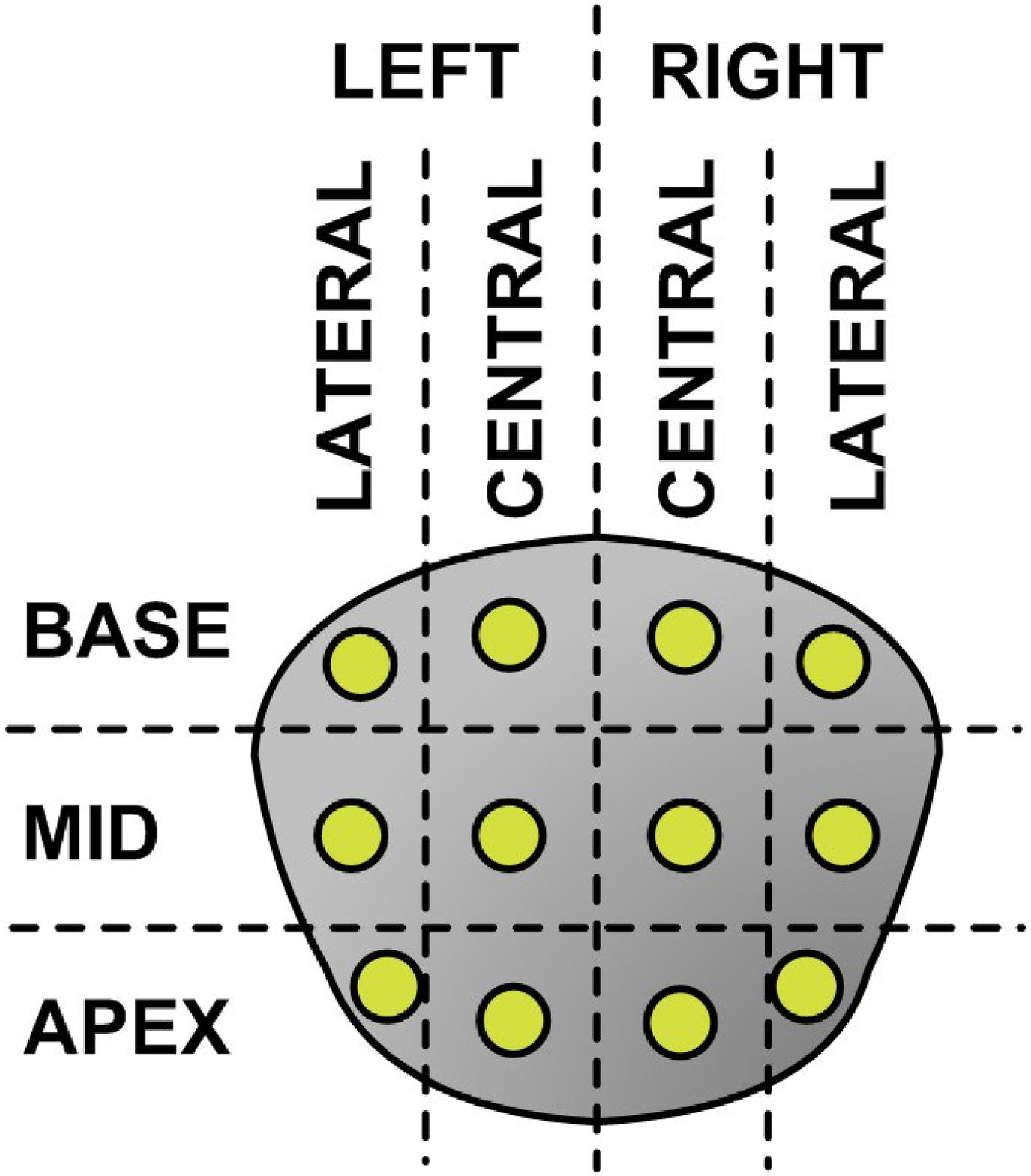}}
\subfigure[]{\includegraphics[width=0.74\textwidth]{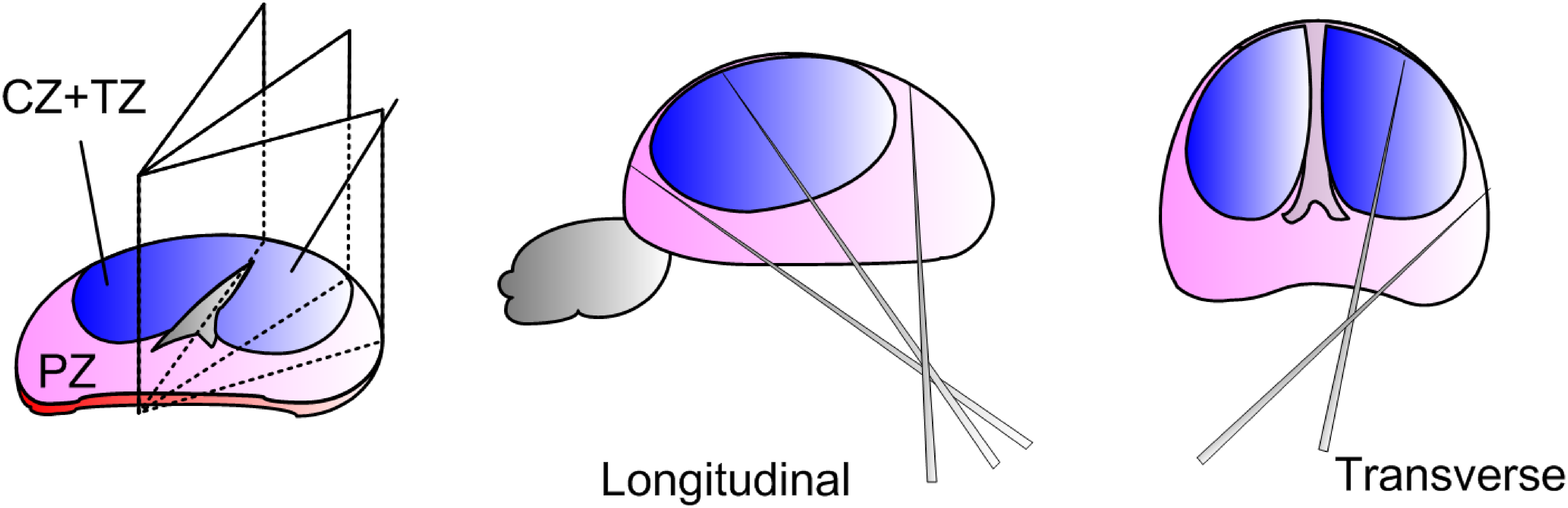}}
\caption{12-core protocol. Fig.~(a) shows the schematized protocol in the coronal plane of the prostate. Fig.~(b) illustrates the protocol in 3D. PZ is the peripheral zone, TZ the transition, and CZ the central zone of the prostate.}
\label{fig:standardprotocol}
\end{figure}

Today, prostate biopsies are the only definitive way to confirm a prostate cancer hypothesis. The current clinical standard is to perform prostate biopsies under 2D transrectal ultrasound (TRUS) control. The US probe is equipped with a needle guide for transrectal access to the prostate, cf. Fig.~\ref{fig:clinicalcontext}. The guide aligns the needle trajectory with the end-fire US image plane, which makes it possible to visualize the trajectory on the image for needle placement control. However, early- and mid-stage carcinoma are mostly isoechogenic, i.e. not visible in US images, which makes it necessary to sample the gland according to a systematic pattern. The standard protocol consists in the acquisition of 10-12 cores and takes roughly into account that most tumors (about 70$\%$) develop in the peripheral zone of the gland (see Fig.~\ref{fig:standardprotocol}). 

\subsection{Clinical issues}
\label{sec:clinicalissues}
The current standard biopsy procedure has several shortcomings: first, it is difficult for the clinician to reach systematic targets accurately because he has to move the probe continuously to place the needle; a constant visual reference is lacking \cite{long07trus}. Second, performing a non-exhaustive systematic search for an invisible target implies that the target can be missed. Negative results leave the clinician in a dilemma when the cancer hypothesis cannot be discarded: his only option is to repeat the biopsy series. Furthermore, the location of the acquired samples with respect to the patient anatomy is only very approximately known after the intervention. Uncertainty about tumor location is a major reason why prostate cancer therapy generally consists in the treatment of the entire gland. Incontinence and impotence are frequent side-effects of whole gland treatments that considerably reduce the quality of life. For this reason a rapidly increasing number of research groups is currently investigating methods that allow the implementation of focal therapy strategies \cite{ahmed07nature}.

\subsection{Proposed solutions}

Several authors have proposed to perform biopsy under MR-guidance to overcome the tumor visibility problem. Hata et al \cite{hata01mriguidance}, Susil et al. \cite{susil04mriguidance} and Krieger et al. \cite{krieger05mriguidance} propose an MR compatible end-effector for transperineal biopsy core acquisition, while Beyersdorff et al. \cite{beyersdorff05mriguidance} propose an end-effector for transrectal access. Additionally, Stoianovici et al. \cite{stoianovici07mrbot} propose a fully actuated transrectal biopsy robot. MR imaging is, however, a costly and sparse resource; it is thus unlikely that MR-guidance will become a standard for the millions of prostate biopsies performed in the US and the EU alone every year. MR-guidance is, however, an interesting option if a biopsy series needs to be repeated because of inconclusive results.

Baumann et al. \cite{baumann07tracking} and Xu et al. \cite{xu07protrack} proposed to acquire a US volume before the intervention and to use it as an anatomical reference. The stream of US control images acquired during the intervention is then registered with the reference volume. This allows the projection of targets defined in the reference volume into the control images, and, conversely, the biopsy trajectory, known in control image space, into the reference volume. This technique makes it possible to improve biopsy distribution accuracy by showing the current trajectory in a fixed reference together with the trajectories of previously acquired biopsy cores. It also allows the clinician to aim for targets defined in the reference volume during a planning phase and to obtain the precise biopsy positions for post-interventional analysis. An example for non-US targets are suspicious lesions found in MR volumes that are multi-modally registered with the US reference volume. It is also possible to derive targets from more sophisticated statistical atlases \cite{shen04atlas} or, in the case of repeated biopsies, they could consist of previously unsampled regions. After the intervention, the biopsy trajectories in the reference volume can be combined with the histological results and used for therapy planning.

Xu et al. \cite{xu07protrack} acquire a freehand 3D US volume and use 2D control images during the intervention. The control images are tracked in operating room space with a magnetic sensor mounted on the probe. In a second step, image-based registration is performed in about 15s to compensate for small organ and patient movements. A similar approach was proposed by Bax et al. \cite{bax08biopsyguidance} and Cool et al. \cite{cool08medphys}, who use an articulated arm for 2D US beam tracking and to acquire a 3D reference volume. Bax does not compensate for patient and gland movements. However, pain-related pelvis movements are frequent, since the patient is not under total anesthesia. In this case, both methods risk to lose track of the gland because the US beam is followed in operating room space and not in organ space, requiring the acquisition of a new reference volume. The acquisition of a 3D volume with a 2D transducer is time-consuming and the reconstruction process inevitably introduces some distortion in the volume. Furthermore, the US probe continually deforms the gland during needle placement, and it is difficult to estimate these non-linear deformations on 2D images. This can lead to inaccurate estimations of the puncture trajectory. Envisioneering Medical Technologies commercialize the TargetScan system \cite{andriole07targetscan} which uses a side-fire probe, flexible biopsy needles and a motorized encoded stepper to place and track the needles. This device shares the draw-backs of the system proposed in \cite{bax08biopsyguidance,cool08medphys}. In \cite{baumann07tracking}, we proposed to use a US probe with an articulated, motorized transducer array that allows to acquire 3D images without moving the probe, which reduces the distortions in the reconstructed volume. The sweep duration of 0.5-5~s makes it possible to acquire volumes also during the biopsy core acquisition phase without delaying the procedure noticeably. The rich spatial information in these 3D control volumes makes it feasible to design a purely image-registration based tracking method that can cope with most types of patient movements during the procedure. In \cite{baumann09miccai} a deformation estimation step was added to the registration pipeline, which reduced the target registration error (TRE) below 1 mm, with a registration time of 6~s. Recently, Karnik et al. \cite{karnik10medphys} reported a TRE of 6.1$\pm$2.0~mm (RMS) with maximum errors of more than 10 mm for transformations obtained with the articulated arm of the system discussed in \cite{bax08biopsyguidance,cool08medphys} ("pre-registration error"). To achieve a more clinically acceptable error, which they define to be a TRE of less than 2.5mm, they evaluated the accuracy of non-linear image-based registration using reconstructed 3D TRUS volumes. They obtained a TRE of 1.5$\pm$0.8~mm with a computation time of 5-90~s. The proposed registration algorithm is based on local optimization and is initialized with transformations obtained from the encoded arm in operating room space. A new reference volume has to be acquired when the patient moves outside the capture range of the registration algorithm.

In this paper, we present a purely US image registration based prostate tracking system capable of deformation estimation. The system does not require beam tracking and reduces hence hardware requirements. It is more accurate and faster than other systems that achieve clinically acceptable accuracy as defined in \cite{karnik10medphys}, and also less sensitive to patient motion. The registration framework, previously presented in \cite{baumann07tracking,baumann09miccai}, is extended and improved: a method is outlined that allows to increase the number of levels in a multi-resolution registration framework by reducing the information loss on coarse levels. This makes it possible to reduce registration time significantly. The kinematic model \cite{baumann07tracking}, used to estimate plausible positions of the US probe with respect to the gland, is extended to cope with varying probe insertion depths. The similarity measure introduced for deformation estimation in \cite{baumann09miccai}, capable of dealing efficiently with the frequently occurring local intensity shifts caused by probe pressure variations and changing US beam angles, is analyzed in more detail. The measure is less complex than the correlation coefficient and therefore requires less data to yield statistically robust estimates, which makes it possible to use it on coarser resolution levels. In a second part we will present our novel clinical prostate biopsy application which is able to compute precise biopsy and cancer maps. It also allows for guidance towards non-US targets and provides the facility to visualize the location of samples acquired during a previous biopsy session. Finally, the different steps of the tracking algorithm are validated on a large set of patient data and an in-depth discussion of the clinically acceptable tracking error is presented. 

\section{3D ultrasound-based prostate tracking}
\label{sec:tracking}
\subsection{Coarse-to-fine strategies}

\begin{figure}
\centering
\includegraphics[width=0.9\textwidth]{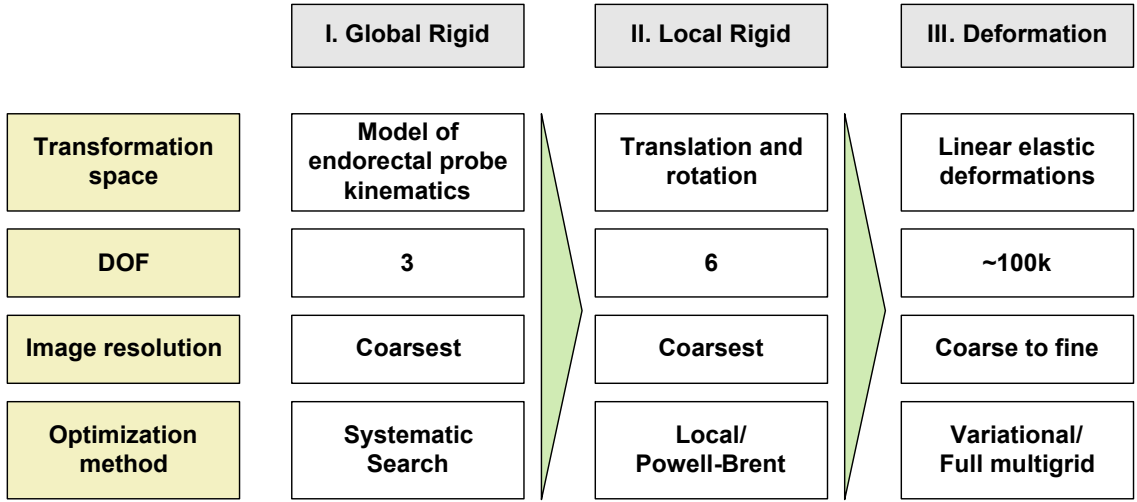}
\caption{Registration pipeline. The dimensionality of the transformation space and the image resolution are successively increased.}
\label{fig:pipeline}
\end{figure}

Non-linear registration of 3D US image streams is currently the most promising approach to perform organ tracking with deformation estimation. The principal challenges of image-based tracking systems are robust outcomes and computational efficiency. A technique to achieve both goals are coarse-to-fine registration strategies that successively increase the degrees of freedom (DOF) of the transformation space and the image resolution. An overview of this strategy is given in Fig.~\ref{fig:pipeline}.

\subsection{Coarse-to-fine transformation spaces}
We propose a 3-step registration pipeline (see Fig.~\ref{fig:pipeline}, upper row) for a purely image-based estimation of the rigid and residual elastic transformations between the prostate imaged in a reference volume and in a tracking volume. At each step, the number of degrees of freedom of the transformation space is increased. This strategy stems from the observation that image registration with high dimensional transformation spaces requires, in general, a fairly good initial estimate in order to converge to the physically correct solution. If initial estimates of the required quality are not available beforehand, they need to be computed on a lower dimensional transformation space on which the distance measure is less exposed to local minima. Ideally, if it is small enough, the search space can be explored systematically, in order to find all major local minima.

In the presented approach, three search spaces are explored using a registration pipeline. The first step of the pipeline consists of a systematic exploration of a 3 DOF model of the probe kinematics, that integrates rectal and image formation constraints. This model excludes solutions that are not plausible in relation to the prostate. The systematic search yields a set of points near strong local minima that are investigated in the second step of the pipeline. Here, a local optimization is performed on each minimum using a classic 6 DOF rigid transformation space. Finally, the best result is retained as the initial estimate for the third registration step, which estimates the elastic deformation.

\subsection{Coarse-to-fine image resolutions}
The second coarse-to-fine strategy operates on the image resolution. While the first two registration steps are executed at very coarse resolution levels of a Gaussian image pyramid, elastic registration descends to finer levels to estimate local details like the deformation caused by a needle insertion. Large parts of the transformation are therefore computed on very coarse levels, not only boosting the registration speed, but also giving the algorithm a more global perspective at early stages of the process.

\begin{figure}
\centering
\subfigure[]{\includegraphics[width=0.36\textwidth]{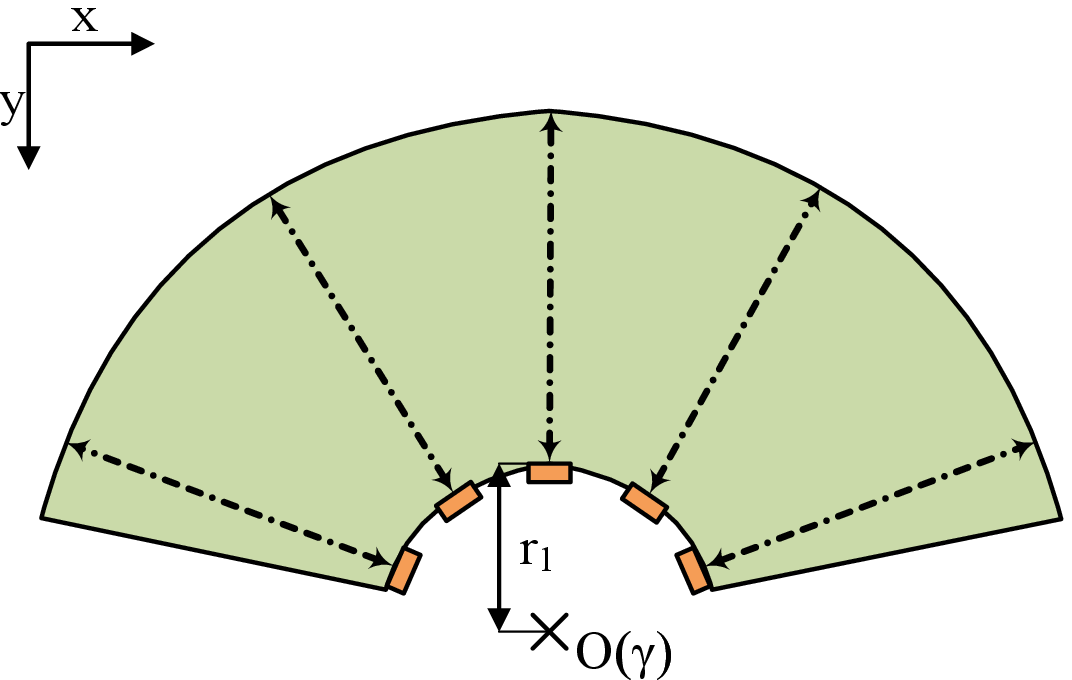}}\hfill
\subfigure[]{\includegraphics[width=0.36\textwidth]{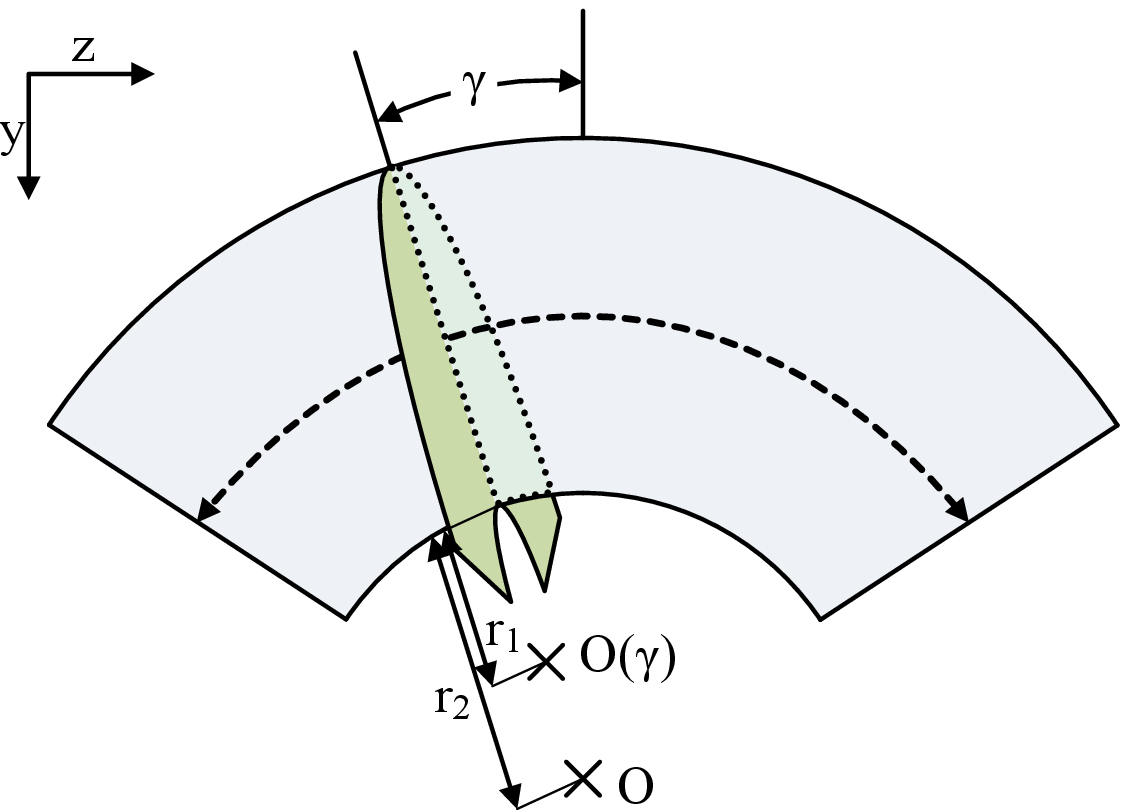}}\hfill
\subfigure[]{\includegraphics[width=0.26\textwidth]{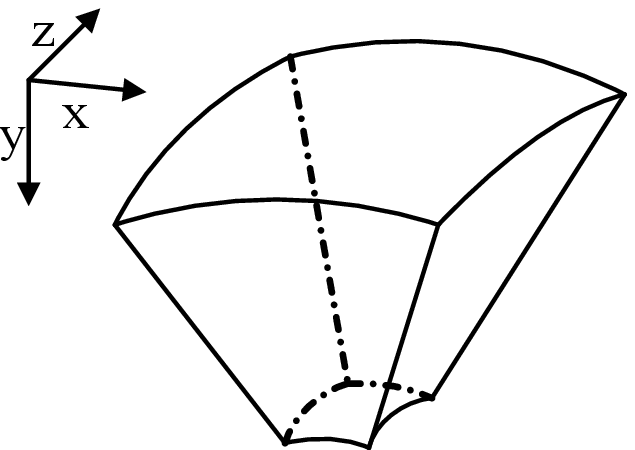}}
\caption{Geometry of end-fire, volume-swept 3D US view cone. Fig.~(a) depicts the 2D transducer array plane, (b) the motor sweep plane, and (c) gives a 3D view of the scanned volume.}
\label{fig:torus}
\end{figure}

Special attention is paid to the construction of the multi-resolution image pyramid. The volume covered by the end-fire US beam corresponds to a section of a torus (cf. Fig.~\ref{fig:torus}). The beam borders in Cartesian space are therefore non-trivial and a complex mask is required to differentiate voxels that carry information from background voxels. The latter must be ignored during image processing to avoid biases. However, most image processing (e.g. gradient computation and Gaussian blurring) is performed on the neighborhood of a voxel, i.e. it cannot be carried out if the neighborhood is incomplete. While this is a minor issue for high resolution images, it poses a major problem on coarser levels of an image pyramid. The entire border is lost with each down-sampling, and the ratio of border voxels to inner voxels increases. As a consequence the information containing volume decreases exponentially with each resolution level and the multi-resolution approach would be limited to a small number of levels.

\begin{figure}
\center
\subfigure{\includegraphics[width=0.2\textwidth]{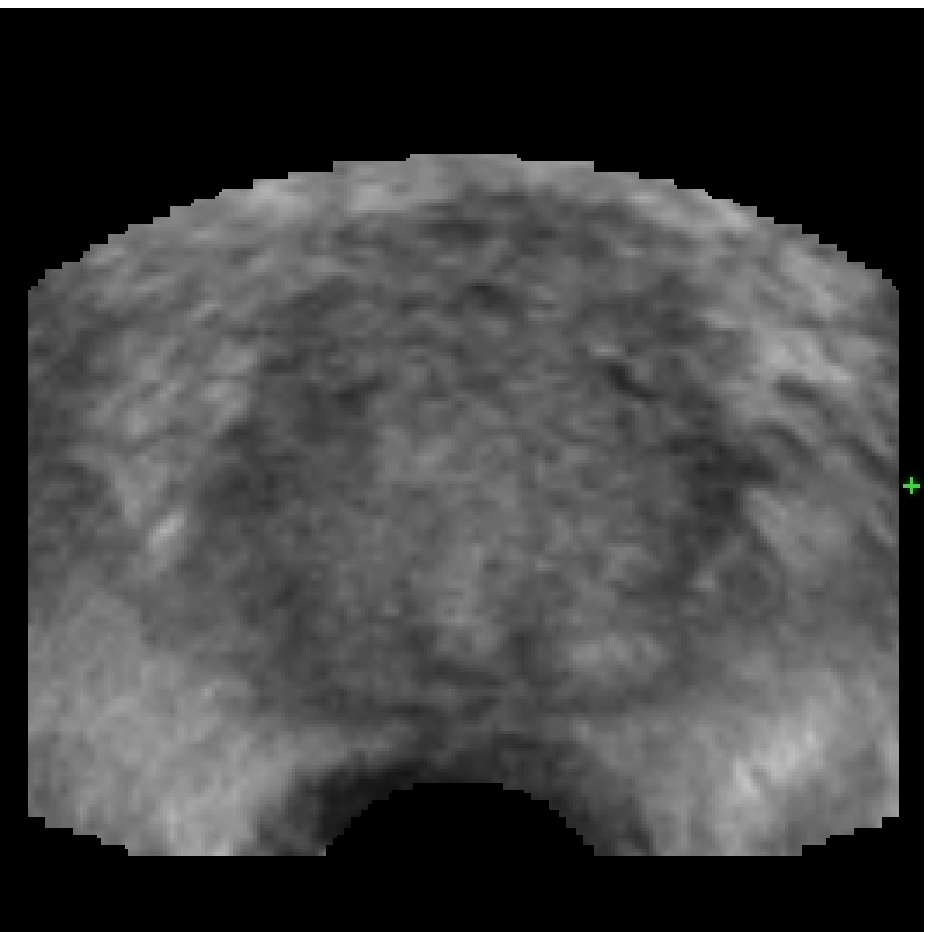}}\hspace{.03\textwidth}
\subfigure{\includegraphics[width=0.2\textwidth]{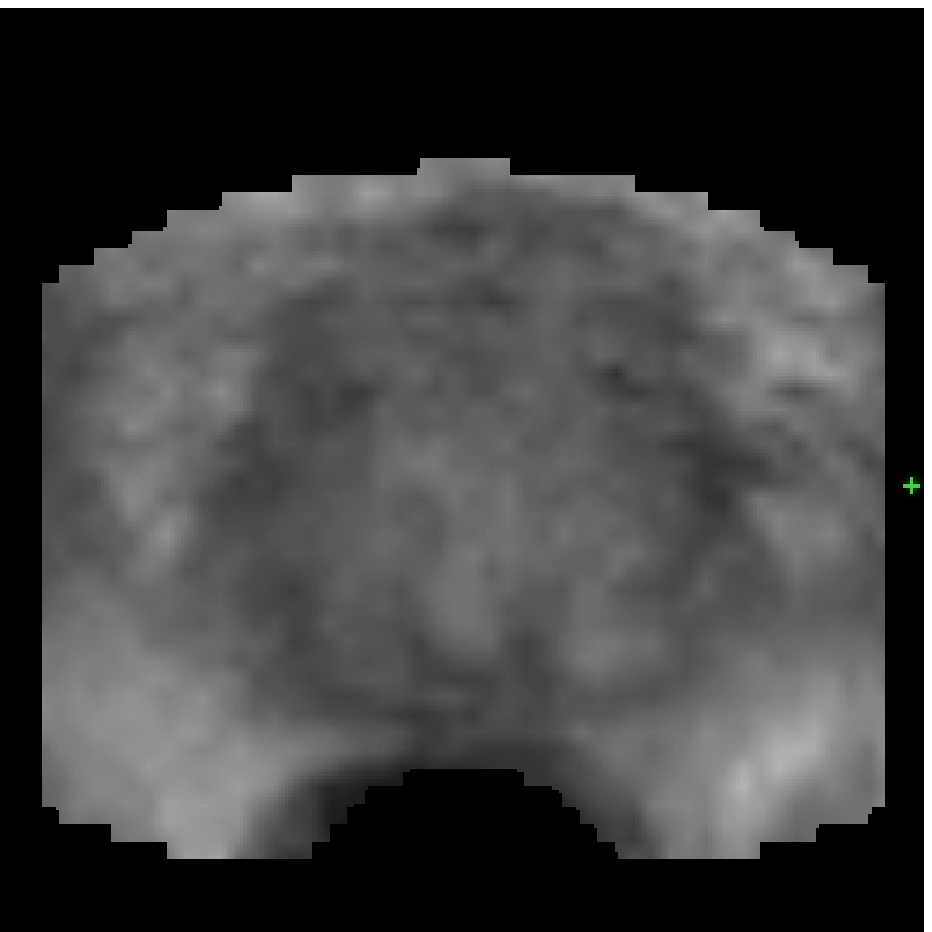}}\hspace{.03\textwidth}
\subfigure{\includegraphics[width=0.2\textwidth]{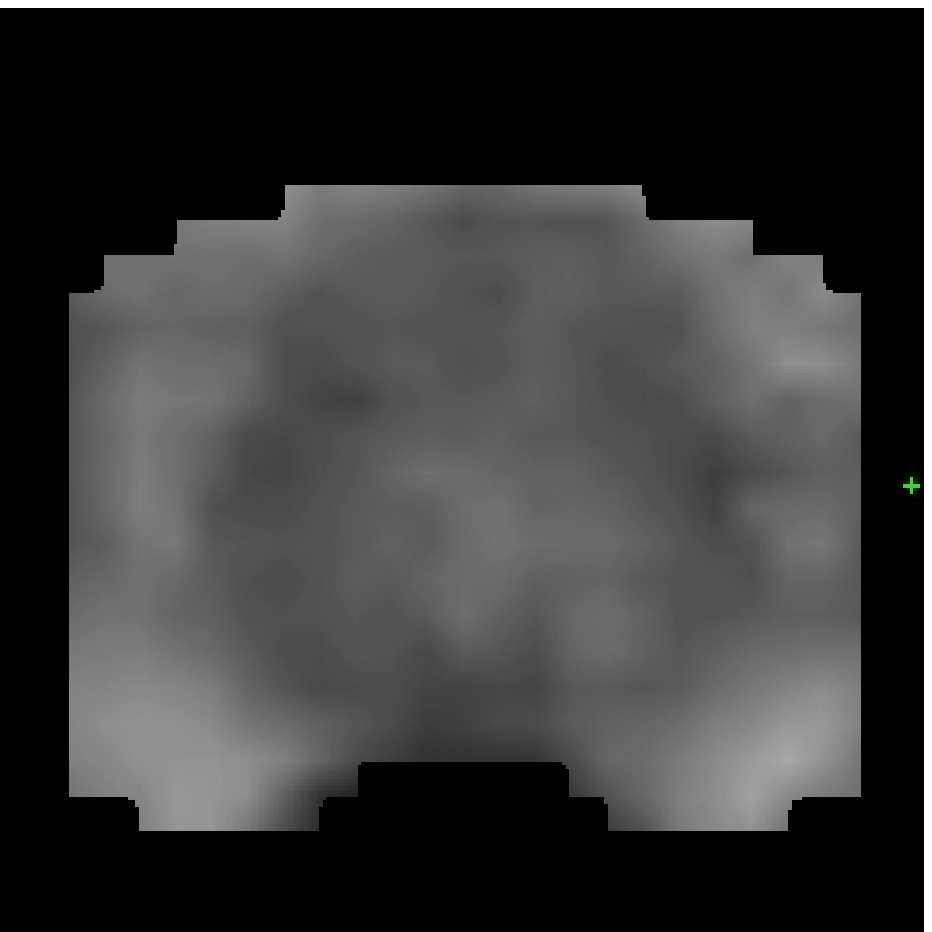}}\hspace{.03\textwidth}
\subfigure{\includegraphics[width=0.2\textwidth]{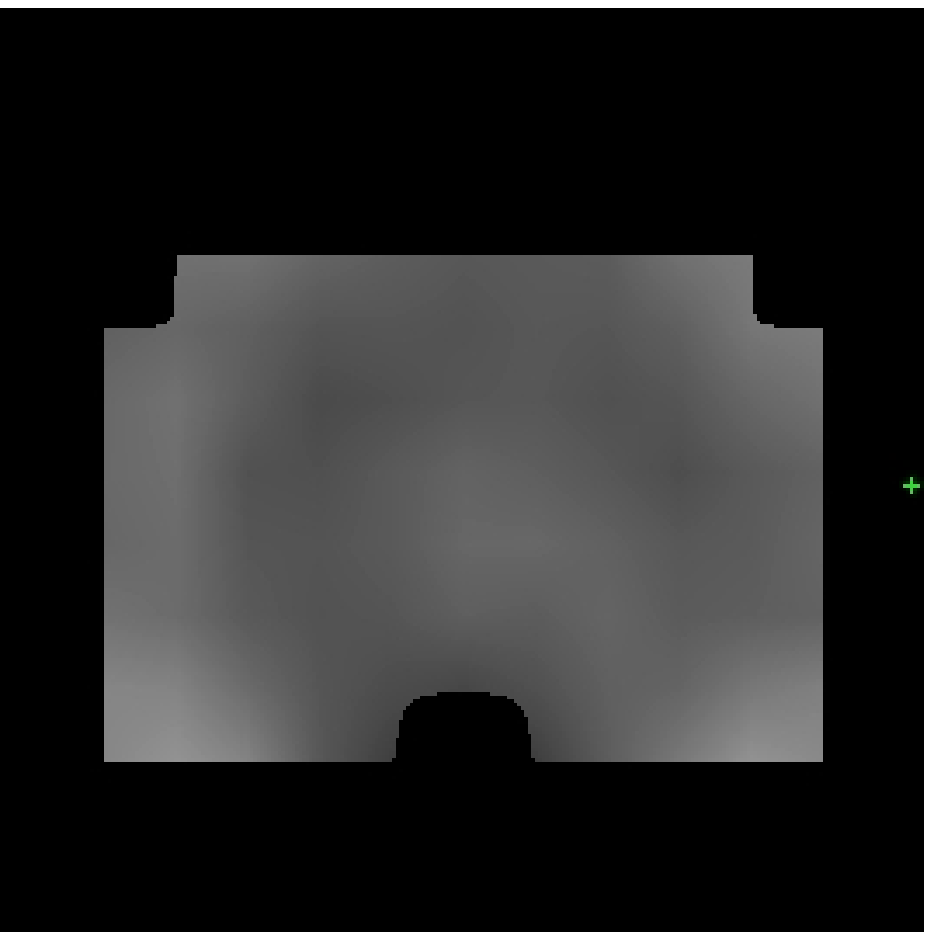}}\\
\subfigure{\includegraphics[width=0.2\textwidth]{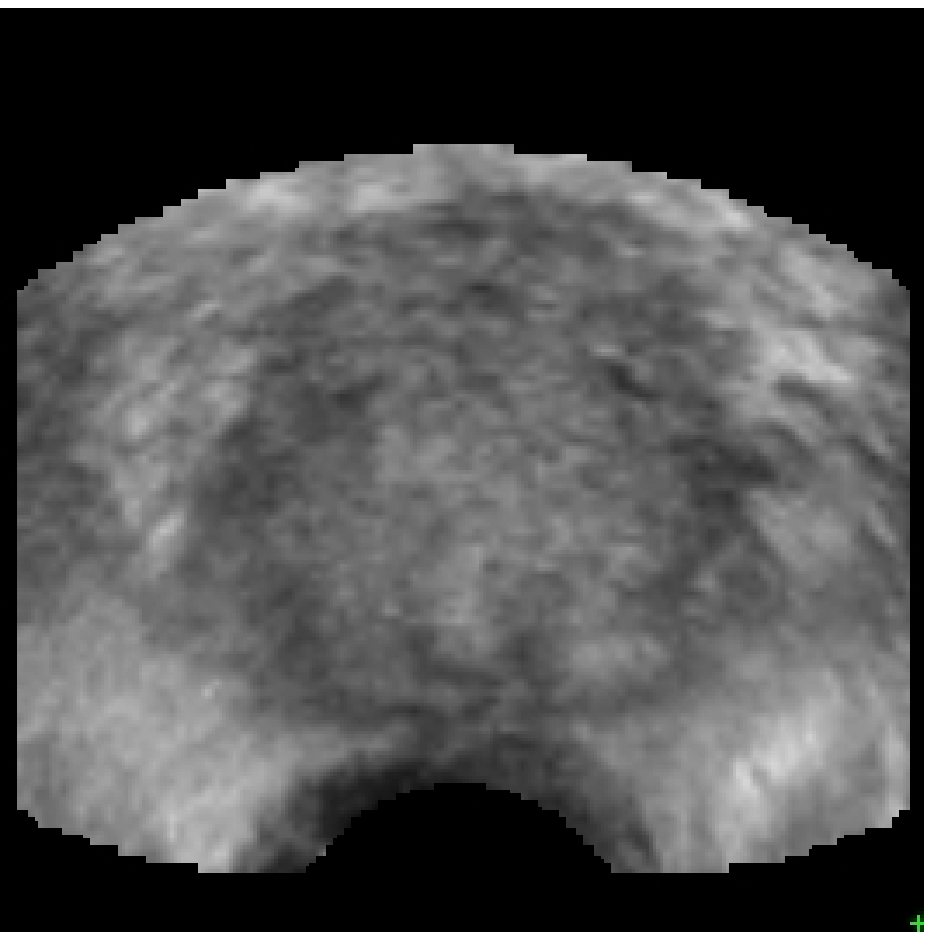}}\hspace{.03\textwidth}
\subfigure{\includegraphics[width=0.2\textwidth]{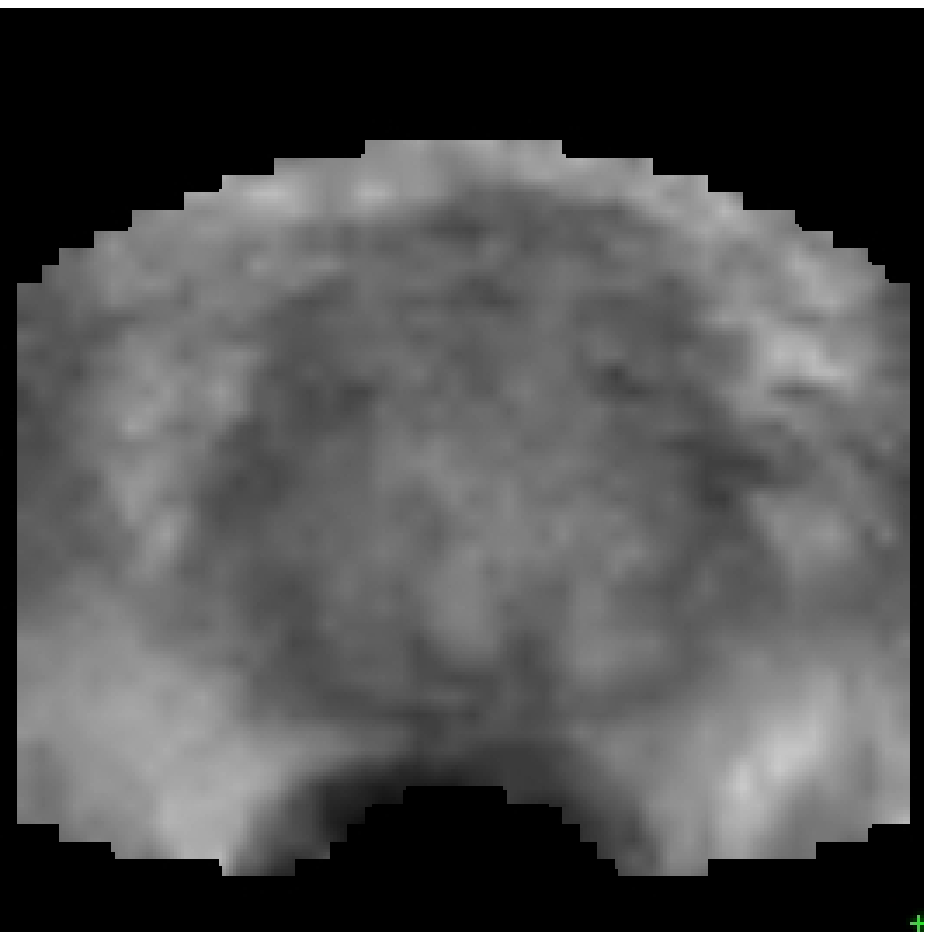}}\hspace{.03\textwidth}
\subfigure{\includegraphics[width=0.2\textwidth]{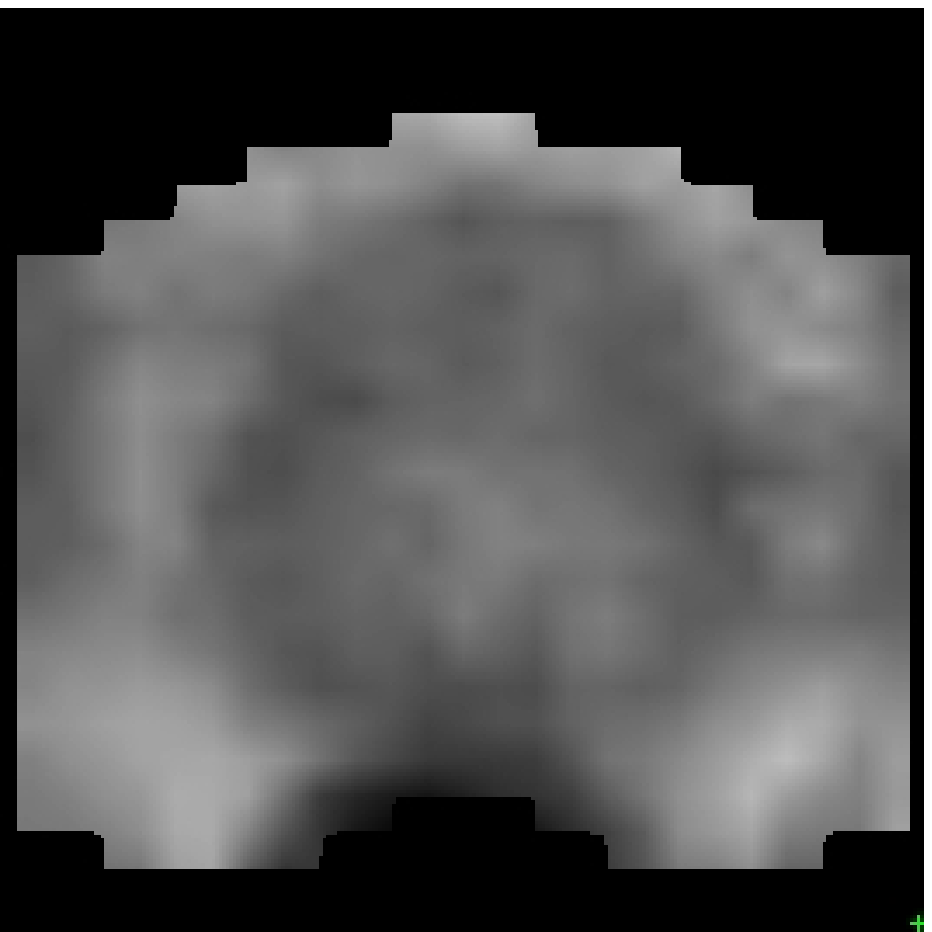}}\hspace{.03\textwidth}
\subfigure{\includegraphics[width=0.2\textwidth]{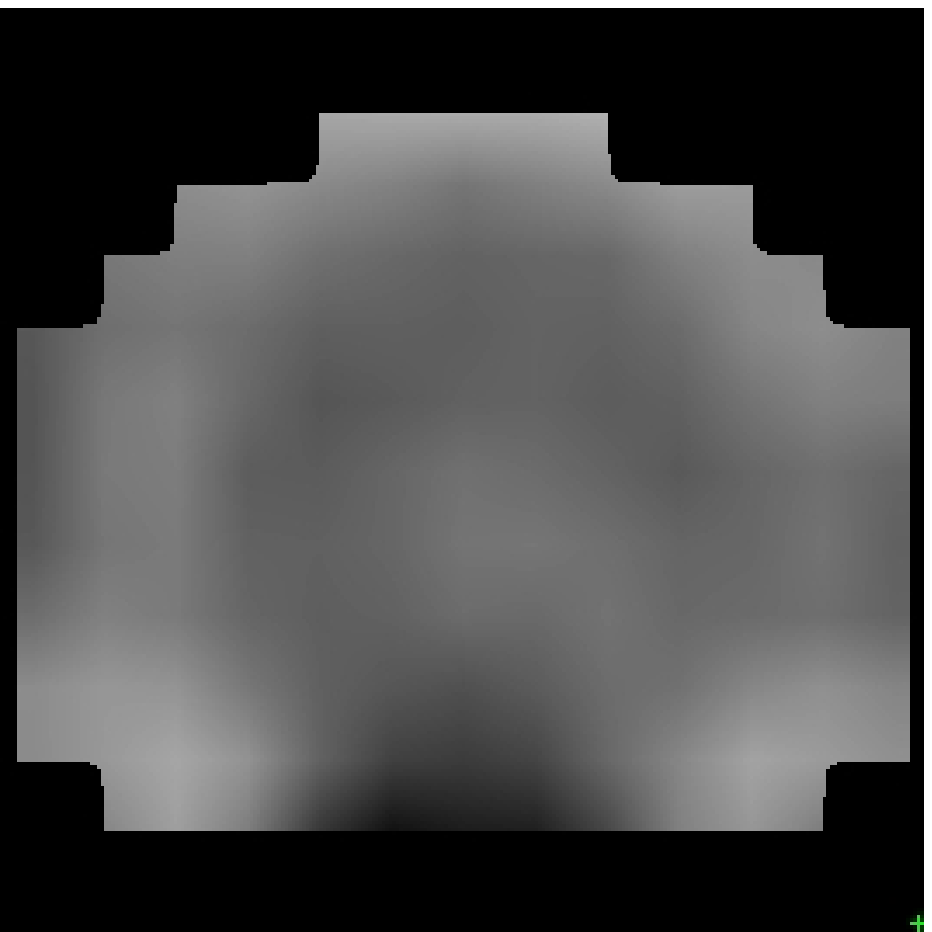}}
\caption{Information loss and image extrapolation. The upper row shows cuts through a multi-resolution pyramid of volumes constructed without information loss containment, from finest (left) to coarsest (right). Note that the volume of information is significantly reduced on the coarsest level (upper row, right-most). The lower row shows a multi-resolution pyramid built with the presented image extrapolation technique. The volume of information remains almost constant.}
\label{fig:multirespyramid}.
\end{figure}

To cope with this important issue, a first order border extrapolation is performed on the mask borders of a pyramid level $k$ before computing level $k+1$. Let $I:\mathbb{R}^3\rightarrow\mathbb{R}$ be the image, $p \in \mathbb{N}^3$ a voxel position and $q \in \mathbb{N}^3$ a neighboring voxel of $p$. Finally, let $d \in \mathbb{N}^3$ be the distance vector $d=q-p$. Then, the Taylor expansion of $I(xd+p), x\in\mathbb{R}$ at point 0 yields, after solving for $I(p)$,
\begin{eqnarray}
I(p)&=&I(xd+p)-x[\frac{d}{dx}I(xd+p)](0)-\sum_{n=2}^{\infty}\frac{x^n}{n!}[\frac{d^n}{dx^n}I(xd+p)](0)\nonumber\\
&\approx&I(xd+p)-x(I(d+p)-I(p)).\nonumber
\end{eqnarray}
This expression is discretized by setting $x=2$, which is the smallest number for which the discrete data prolongation is meaningful (it follows that two consecutive voxels in direction $d$ are needed for an estimation). We get, therefore,
\begin{eqnarray}
I_q(p)&=&2I(p+d)-I(p+2d)=2I(q)-I(2q-p).
\end{eqnarray}
The final intensity is the mean of $I_q(p)$ for all $q \in \mathcal{N}_{26}(p)$ for which the computation is possible, where $\mathcal{N}_{26}(p)$ is the 26-neighborhood of $p$. If less than a third of the directions can be computed, the voxel is discarded. Fig.~\ref{fig:multirespyramid} shows that the volume containing information remains almost constant with this approach.

\section{Rigid registration}

\subsection{Parametric optimization framework}
The first two steps of the registration pipeline (cf. Fig.~\ref{fig:pipeline}) determine the rigid transformation between the prostate in the reference and the tracking images. Both steps operate on low-dimensional search spaces that can be efficiently solved with a parametric formulation of the optimization problem. We look for a set of parameters $\theta^\ast$ in a parameter space $\Theta \subset \mathbb{R}^d$ such that
\begin{equation}
\theta^\ast[I_1, I_2; \varphi] = \argmin{\theta \in \Theta}\mathcal{D}[I_1, I_2\circ\varphi(\theta)],
\end{equation}
where the $I_i:\mathbb{R}^3\rightarrow\mathbb{R}$ are images and $\varphi(\theta,x) : \Theta\times\mathbb{R}^3\rightarrow\mathbb{R}^3$ defines a spatial transformation in function of $\theta$ and $x \in \mathbb{R}^3$. Finally, $\mathcal{D}[I_1, I_2]$ is an image distance measure. In the presented approach, the Pearson correlation coefficient\footnote{Also known as \textit{cross correlation} or simply  \textit{correlation coefficient}.}
\label{eqn:pearson}
\begin{equation}
\mathcal{D}_{CC}=\frac{\mbox{Cov}(I_1, I_2)}{\sqrt{\mbox{Var}(I_1)\mbox{Var}(I_2)}}
\end{equation}
is used as image distance measure for parametric optimization.

\subsection{Global rigid registration with endorectal probe kinematics}
\label{sec:kinematics}
The presence of local minima in the large rigid search space make it difficult to use local optimization methods (e.g. gradient descent or Powell-Brent, \cite{Brent73Algorithms,NumericalRecipes92}) directly. The objective of the first registration step is therefore to estimate the position of the US probe relative to the organ to find a transformation close to the solution. Instead of using magnetic or mechanic tracking systems, a model of the kinematics of the probe under physiological and image acquisition constraints was designed. The idea is to profit from the following observations:
\begin{enumerate}
\item The probe head must be in contact with the thin rectal wall in front of the prostate during image acquisition. This is a necessary condition to obtain an image of the gland.
\item The anal sphincter heavily constrains probe motion in the rectum.
\end{enumerate}

\begin{figure}
\subfigure[]{\includegraphics[width=0.32\textwidth]{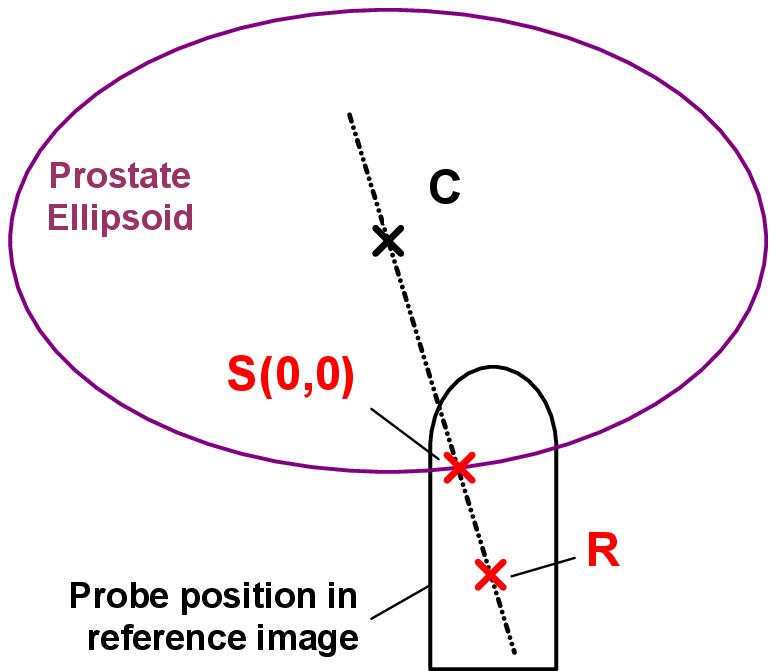}}
\subfigure[]{\includegraphics[width=0.32\textwidth]{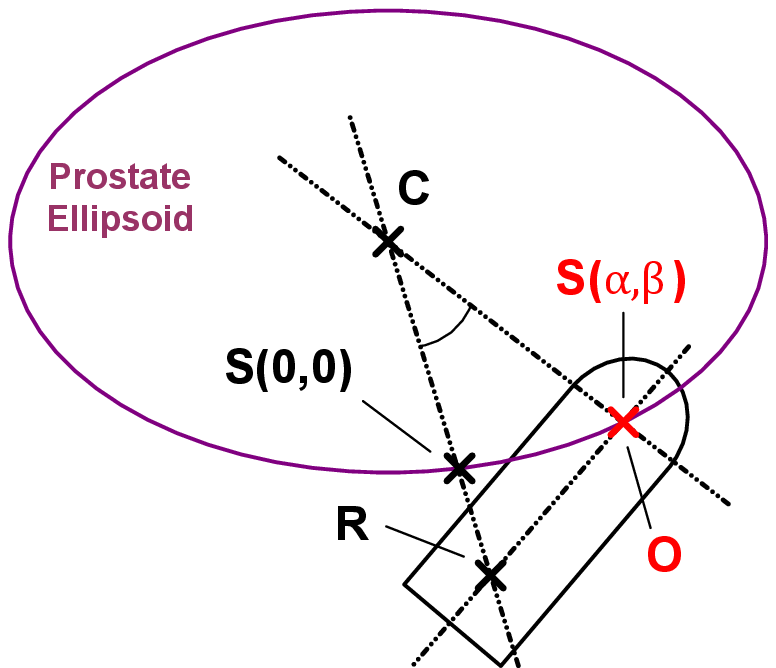}}
\subfigure[]{\includegraphics[width=0.32\textwidth]{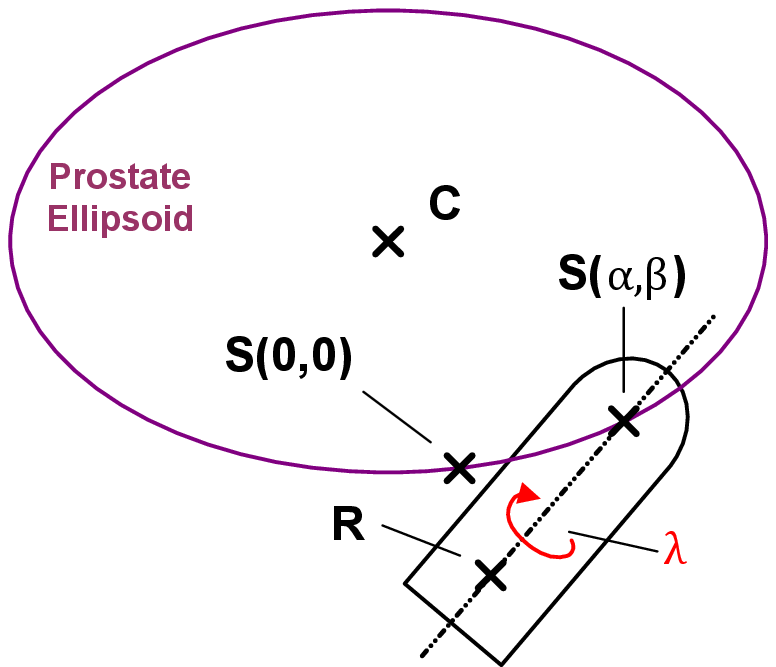}}
\caption{Model of probe head kinematics under endorectal and prostate image formation constraints. Fig.~(a) shows the construction of the origin $S(0,0)$ of the surface discretization, which corresponds to the intersection point of a line defined by the rectal fixed point $R$ and the ellipsoid center $C$ with the ellipsoid. Fig.~(b) illustrates the exploration of the surface with the probe using the spherical angles $\alpha$ and $\beta$. The US origin $O$ is placed on the surface of the ellipsoid. Fig.~(c) finally models the rotations around the axis of the US probe.}
\label{fig:kinematics}
\end{figure}

This makes it possible to define a kinematic model of probe head motion with respect to the prostate surface. It is assumed that the center of the probe head lies on the principal axis of the probe (the \textit{probe axis}). It is further assumed that the probe axis always lies on a hypothetical rectal fixed point $R\in\mathbb{R}^3$ which approximates the sphincter constraints. Finally, an approximate estimation of the prostate surface in the reference image is needed. The role of the surface estimation is comparable to that of a region of interest (ROI), it therefore does not need to be very accurate. We thus opted to use a simple ellipsoid, which is manually defined by the clinician on the first volume he acquires. This step consists in the placement of an axis-aligned rectangular bounding box around the prostate, which can be done with a few mouse clicks. All positions for which the US origin $O$ lies on the surface approximation and the probe axis lies on $R$ are considered as plausible probe positions, see also Fig.~\ref{fig:kinematics}a and b. The contact points can be defined with two parameters $\alpha$ and $\beta$ using a spherical representation of the ellipsoid. A third parameter $\lambda$ is needed to model the probe rotations around the probe axis, see Fig.~\ref{fig:kinematics}c.

In the presented approach it is further assumed that the probe head position in the image and the probe head radius are known. The precision of the fixed point $R$ is not crucial, and it is possible to use the same fixed point $R$ for all patients. It is an approximation that stems from a learning set that was semi-automatically registered: for a given patient $i$ of the learning set, $R^i$ was defined as the point with minimal distance to the set of probe axes projected into the reference volume after registration. The mean of the $R^i$ is an approximation of $R$ that works well in practice (cf. Sec.~\ref{sec:experiments}, where the accuracy of the model is evaluated).

The resulting 3D transformation space is small enough for a systematic exploration with an acceptable computational burden. To cope with variations in probe pressure, i.e. the insertion depth of the probe into the rectal tissues, the model is explored at five different depths in the probe axis direction at steps of 5~mm. The parameter range is configured such that probe tilt angles of up to 30° and arbitrary rotations around the probe axis are considered. The systematic search is performed on very coarse resolution levels (4 resolution reductions of factor $2^3$) to increase the performance. 

\subsection{Local rigid registration}

The five probe positions with optimal $\mathcal{D}_{CC}$ are used as start positions for multiple rigid registrations. The distance measure is optimized on this 6 DOF search space using a fast local method (Powell-Brent \cite{Brent73Algorithms,NumericalRecipes92}). The transformation $\varphi(o,\omega;x): x\mapsto R_\omega x+o$ models the rigid space, with the origin $o\in\mathbb{R}^3$ and the three Euler angles $\omega\in\mathbb{R}^3$ that define the rotation matrix $R_\omega$. The origin $o$ is set to the center of the ellipsoidal prostate approximation described in Sec.~\ref{sec:kinematics}.

The goal of this step is to find the local minima of the distance measure $\mathcal{D}_{CC}$ in the neighborhood of the initial estimates, which makes it possible to make a robust final choice between them. The estimation with minimal distance is retained. This step can be executed on a fairly coarse level since the finer details will be considered in the following elastic estimation step. The rigid registration can hence be performed rapidly.


\section{Elastic registration}

\subsection{Framework for non-linear registration}

Image-based deformation estimation can be formulated as an optimization process of a local distance measure. Let $I_1, I_2:\mathbb{R}^3\rightarrow\mathbb{R}$ be images, $\varphi:\mathbb{R}^3\rightarrow\mathbb{R}^3$ the deformation function and the functional $\mathcal{D}[I_1,I_2;\varphi]$ a measure of the distance between $I_1$ and $I_2\circ\varphi$. In contrast to parametric approaches that use basis functions to build the deformation function, we will follow a variational approach \cite{valadez02thesis,Modersitzki04numerical,cachier02thesis} and define $\varphi(x)=x+u(x)$, where $u:\mathbb{R}^3\rightarrow\mathbb{R}^3$ is assumed to be a diffeomorphism. The deformation could then be estimated by solving the optimization problem
\begin{equation}
\varphi^\ast=\argmin{\varphi}\left(\mathcal{E}[I_1,I_2;\varphi]\right),
\end{equation}
where the registration energy $\mathcal{E}$ simply corresponds to $\mathcal{D}$.
Straightforward minimization of a distance measure, however, yields poor results in general due to countless local minima, in particular in the presence of noise, partial object occlusion and other imperfections in the image data. Unfortunately, US is a particularly noisy modality, which makes 3D US based deformation estimation vulnerable to local misregistrations. This problem can be addressed by integration of a priori models of the expected deformation. This can be done implicitly by adding further energy terms to the objective function. In this work, inverse consistency and elastic regularization energies are added. 

\subsection{Inverse consistency constraints}
In non-linear image registration, the forward estimation that minimizes $\mathcal{E}[I_1,I_2;\varphi]$ does not generally yield the inverse of the backward estimation that minimizes $\mathcal{E}[I_2,I_1;\psi]$, i.e. $\varphi \circ \psi \neq Id$ with $Id:\mathbb{R}^3\rightarrow\mathbb{R}^3,x\mapsto x$. Introduction of Zhang's inverse consistency constraint \cite{zhang06consistent} 
\begin{equation}
\mathcal{I}[\psi;\varphi]=\int_\Omega||\psi \circ \varphi - Id||_{\mathbb{R}^3}^2\ dx
\end{equation}
as additional energy penalizes solutions that lead to inconsistent inverse transformations, where $\Omega \subset \mathbb{R}^3$ is the registration domain in image space. Estimation of the forward and the backward deformations is coupled by an alternating iterative optimization
\begin{eqnarray}
\label{invconsistency}
\varphi^{k+1}=\argmin{\varphi}\left(\mathcal{E}[I_1,I_2;\varphi]+\mathcal{I}[\psi^{k};\varphi]\right),\\
\psi^{k+1}=\argmin{\psi}\left(\mathcal{E}[I_2,I_1;\psi]+\mathcal{I}[\varphi^{k};\psi]\right).
\end{eqnarray}
Concurrent estimation with mutual correction reduces the risk of local misregistrations.

\subsection{Elastic regularization}
The deformation of the prostate caused by probe pressure is mostly elastic, which justifies the introduction of the linearized elastic potential \cite{Modersitzki04numerical} 
\begin{equation}
\mathcal{E}[\varphi]=\mathcal{E}[u+Id]=\int_\Omega \frac{\mu}{4}\sum^3_{j,k=1}\left(\partial_{x_j}u_k+\partial_{x_k}u_j\right)^2+\frac{\lambda}{2}(\mbox{div}\ u)^2\ dx
\end{equation}
as additional energy, where $\lambda$ and $\mu$ are the Lamé coefficients, which are related to Poisson's coefficient $\nu$ and Young's modulus $E$ via the equations
\begin{equation}
E=\frac{(3\lambda+2\mu)\mu}{\lambda+\mu} \mbox{   and   } \nu=\frac{\lambda}{2(\lambda+\mu)}. \nonumber
\end{equation}

Note that the linear elastic potential prevents the estimation of strong deformations when operating with non-physical, fractional forces derived from local image dissimilarities. This choice was made to avoid strong misregistrations when $I_1$ and $I_2$ contain local differences for which the image distance metric is not invariant. It is, however, possible to use for example the less restrictive fluid regularization to compute complete forces, and to apply linear elastic regularization in a second step. The low target registration error (TRE) of rigid registration (cf. Sec.~\ref{sec:experiments}) indicates however that the gland rarely gets deformed more than 10$\%$, which is the reason why we did not further investigate this approach.

\subsection{Image distance measure}

\begin{figure}
\subfigure[]{\includegraphics[width=.25\textwidth]{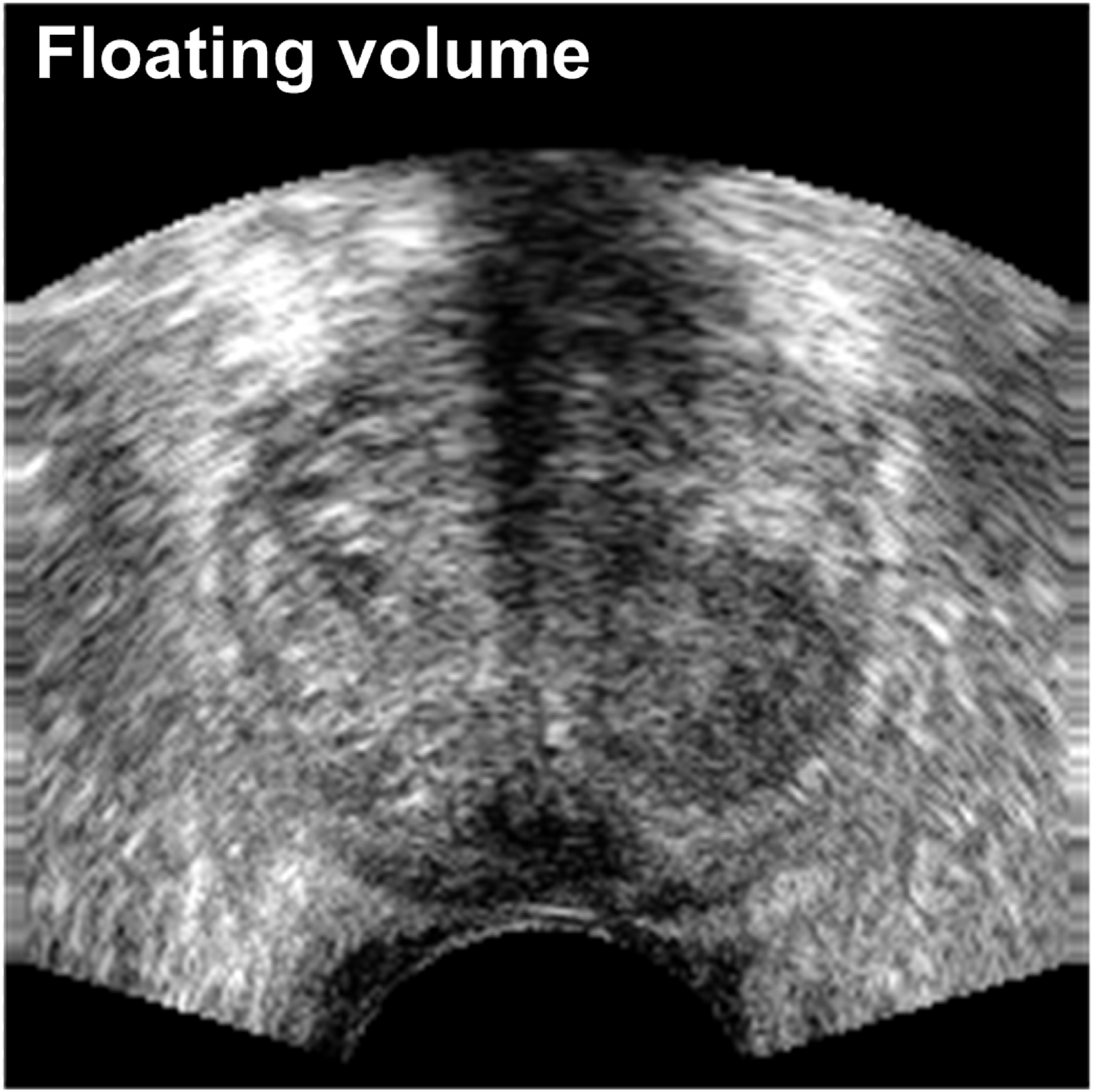}}\hfill
\subfigure[]{\includegraphics[width=.25\textwidth]{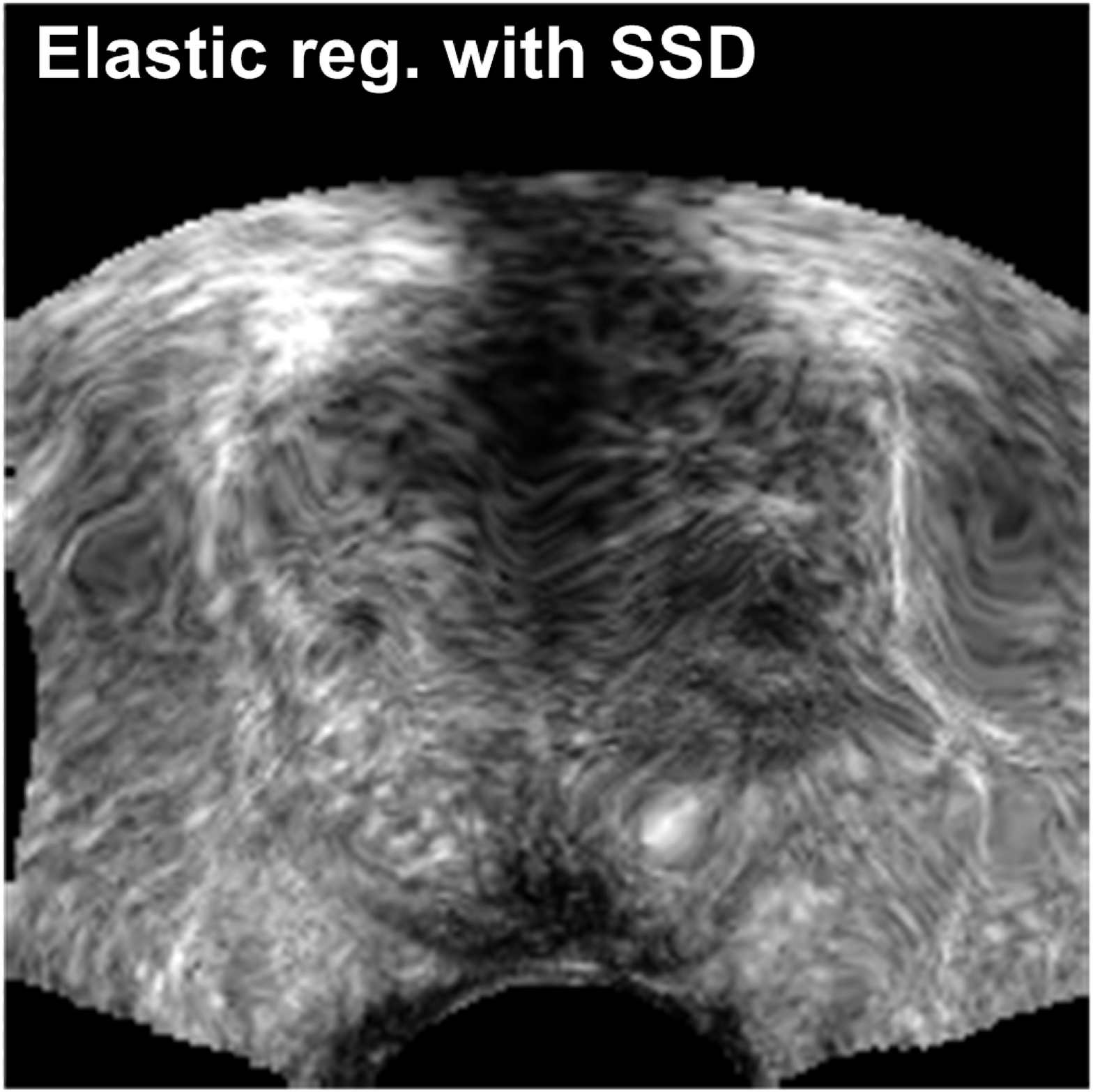}}\hfill
\subfigure[]{\includegraphics[width=.25\textwidth]{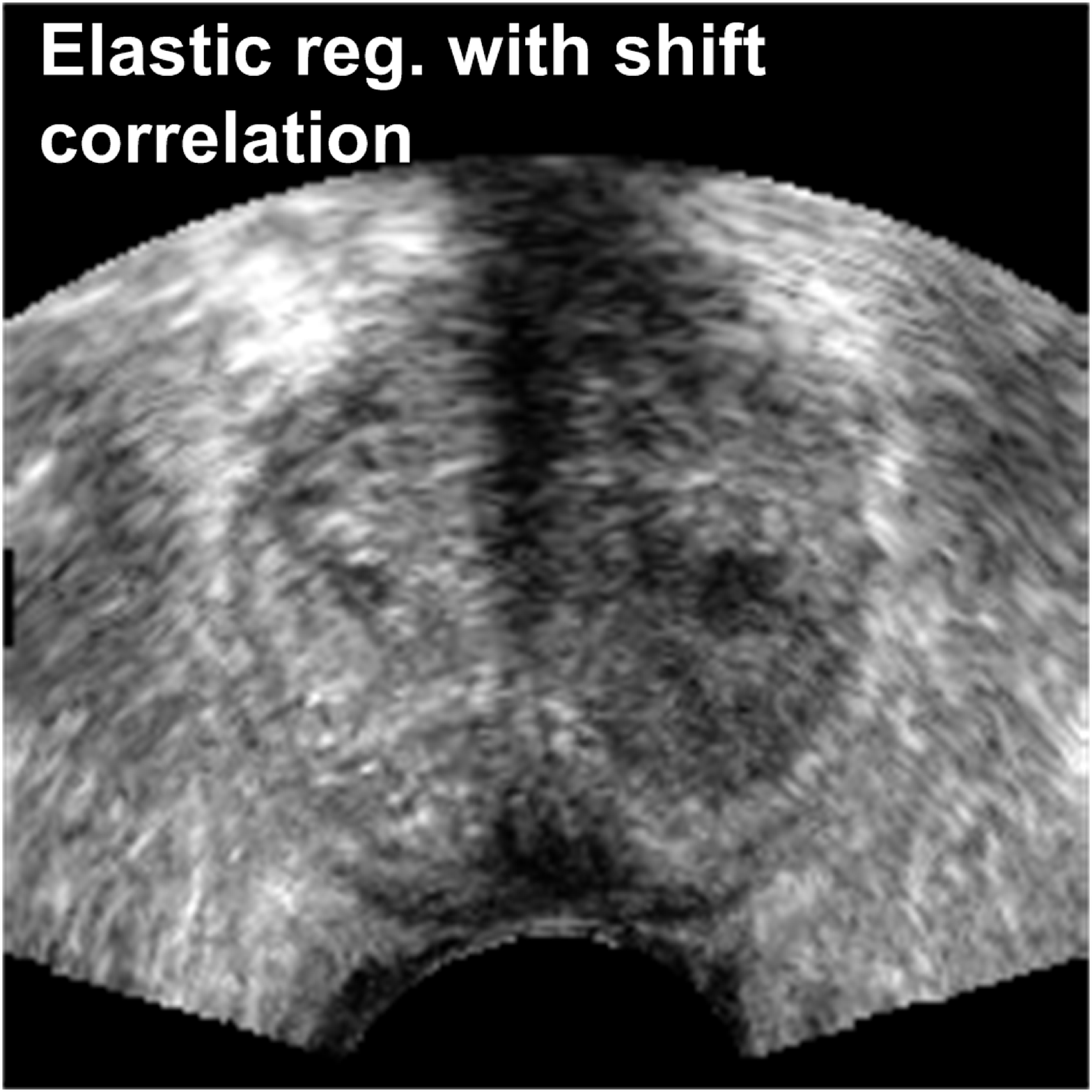}}\hfill
\subfigure[]{\includegraphics[width=.25\textwidth]{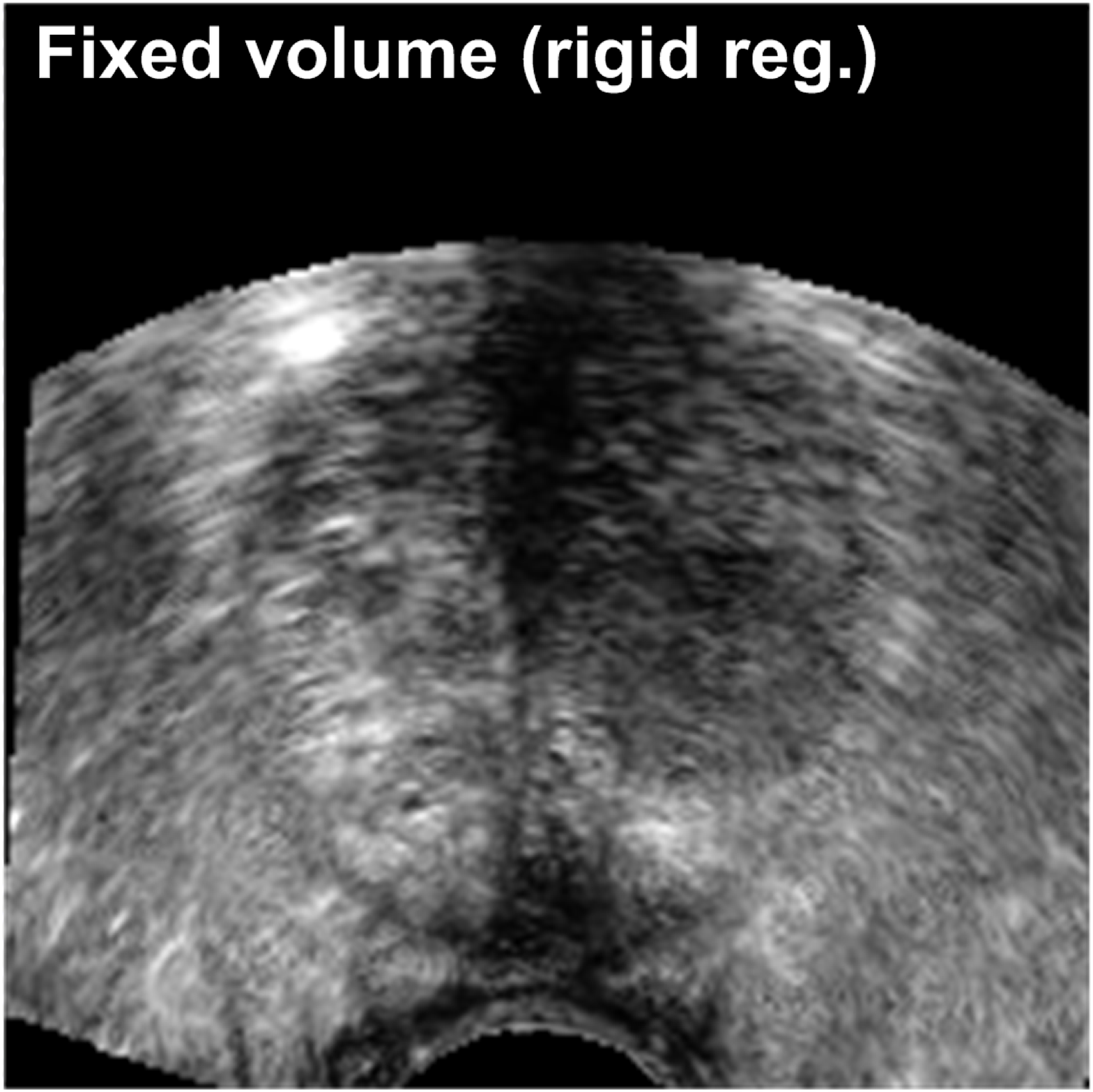}}
\caption{Intensity shift correlation model. Fig.~(a) shows the floating volume, Fig.~(d) the fixed volume (after rigid registration), Fig.~(b) the elastic registration with the SSD correlation model and Fig.~(c) shows the result with the intensity shift correlation model and inverse consistency constraints. All other parameters were identical for both registrations. The SSD driven registration is incorrect because of various local intensity shifts that are caused by the difference in probe pressure between the acquisition of the fixed and the floating images. The intensity shift correlation model correctly handles this problem and converges to the physically correct result.}
\label{fig:shift}
\end{figure}

As for rigid registration, the image distance measure is the driving energy of the deformation estimation process. While a global measure is used for rigid registration, deformation estimation requires a local measure to capture the transformation details. The degree of locality of a measure is in general determined by the size of its convolution kernel. The more local the measure is, the more details can be captured on a given resolution level. In consequence, the number of resolution levels that can be used for deformation estimation, and hence the computation time, depends heavily on the locality of the measure. On the other hand, the measure must also be able to cope with modality-specific image variations like for example intensity changes when the ultrasound beam angle changes with respect to an imaged surface. Experiments on patient data have shown that the sum of squared distances (SSD), which has a minimal kernel size, is a poor distance measure for deformation estimation in US images. Local intensity changes are frequent due to changing US beam angles with respect to the tissues and probe pressure variations. The Pearson correlation that was used for rigid registration requires, however, a considerably larger kernel size to achieve statistical power, in particular if its complex derivative is used. The generalized correlation ratio introduced by Roche et al. \cite{roche01USMRI} allows us to derive a more appropriate correlation-based distance measure:
\begin{equation}
\mathcal{D}_{CR}[I_1,I_2;f]=\tfrac{1}{|\Omega|}\sum_{x\in\Omega}(I_1(x)-f(I_2(x)))^2,
\end{equation}
where the function $f:\mathbb{R}\rightarrow\mathbb{R}$ models the intensity mapping. When a linear relationship $f=\alpha I+\beta$ is assumed, the measure corresponds to the Pearson correlation $\mathcal{D}_{CC}$ used for rigid registration (cf. Eqn.~\ref{eqn:pearson}). Note that $\mathcal{D}_{CC}$ estimates the optimal $\alpha$ and $\beta$ implicitly from the image data. Removing $\alpha$ leads to a correlation measure with only one intrinsic DOF, i.e.
\begin{equation}
\mathcal{D}_{CR}[I_1,I_2;I(\cdot)+\beta]=\frac{1}{|\Omega|}\sum_{x\in\Omega}(I_1(x)-I_2(x)-\beta)^2.
\end{equation}
The parameter $\beta$ can be directly estimated on the difference image using a Gaussian convolution $\mathcal{G}_\sigma(x)$. Introducing the registration transformation $\varphi$ and switching to the continuous domain yields
\begin{equation}
\mathcal{D}_{SC}[I_1, I_2;\varphi]:=\int_\Omega (I_1(x)-I_2\circ\varphi(x)-(I_1-I_2\circ\varphi)\conv(x))^2\ dx.
\end{equation}
This measure requires less data to achieve statistical power compared to the two-dimensional Pearson correlation, but it is more general than the identity-assuming SSD with its zero DOF. It is invariant to low-frequency intensity shifts between the compared images, which is why we called it \textit{shift correlation} $\mathcal{D}_{SC}$. The invariant frequency range is controlled by the standard deviation $\sigma$ of $\mathcal{G}$. If $\sigma$ gets smaller, the cropped range gets larger, and the registration convergence rate decreases and may even stall if only high frequency noise is left. When used with a multi-resolution solver on a Gaussian pyramid (cf. Sec.~\ref{sec:solver}), which implicitly performs a low-pass filtering of the intensity variations on coarse resolutions, this approach transforms to a band-pass filtering on varying frequency bands. In this configuration it is sufficient to chose relatively small standard deviations, without risking registration inefficiencies. Fig.~\ref{fig:shift} illustrates the performance of the shift correlation model combined with the inverse consistency constraint.


\subsection{Solver}
\label{sec:solver}
Combination of the energy terms yields the alternating system
\begin{eqnarray}
\label{solver1}
\varphi^\ast &=&\argmin{\varphi}\left(\mathcal{D}_{SC}[I_1,I_2,\varphi]+\mathcal{E}[\varphi]+\mathcal{I}[\psi;\varphi]\right),\\
\label{solver2}
\psi^\ast &=&\argmin{\psi}\left(\mathcal{D}_{SC}[I_2,I_1,\psi]+\mathcal{E}[\psi]+\mathcal{I}[\varphi;\psi]\right).
\end{eqnarray}
An iterative two-step minimization scheme is used to solve both objective functions. The Euler-Lagrange equations of Eqn.~\ref{solver1} and \ref{solver2} are rewritten as a fixed point iteration
\begin{eqnarray}
\label{partial1}
\frac{\varphi^{k+1}-\varphi^{k}}{\delta t} = \mathcal{L}[\varphi^{k}] + f_{\mathcal{D}_{SC}}[I_1,I_2;\varphi^{k}]+f_\mathcal{I}[\psi^{k};\varphi^{k}],\\
\label{partial2}
\frac{\psi^{k+1}-\psi^{k}}{\delta t} = \mathcal{L}[\psi^{k}] +
f_{\mathcal{D}_{SC}}[I_2,I_1;\varphi^{k}]+f_\mathcal{I}[\varphi^{k};\psi^{k}],
\end{eqnarray}
where $t \in \mathbb{R}$ controls the discretization granularity, and with the elliptic partial differential operator
\begin{equation}
\mathcal{L}[\varphi]=\mathcal{L}[u+Id]=\mu \Delta u + (\lambda + \mu)\nabla \mbox{div}\ u,
\end{equation}
which is obtained from the Gâteaux-derivative of $\mathcal{E}$ at $\varphi$, cf. \cite{Modersitzki04numerical}. The Gâteaux derivative of $\mathcal{I}$ at $\varphi$ yields the force term
\begin{equation}
f_\mathcal{I}[\psi;\varphi]=(\nabla \psi)\circ\varphi\cdot(\psi\circ\varphi-Id).
\end{equation}
Finally, the Gâteaux derivative of $\mathcal{D}_{SC}$ at $\varphi$ provides us the image-based forces

\begin{align}
f_{\mathcal{D}_{SC}}[I_1,I_2;\varphi]=&-\left(I_{1} - 2I_1\conv + I_1\conv\conv - \right.\\ \nonumber
&\left.I_2\circ\varphi + 2I_2\circ\varphi\conv - I_2\circ\varphi\conv\conv\right)\cdot(\nabla I_2)\circ\varphi.
\end{align}
To avoid the expensive double convolution we use the approximation
\begin{equation}
\label{eqn:simplif}
I_k\conv\conv\approx I_k\conv,
\end{equation}
which simplifies the gradient to
%
\begin{align}
\label{eqn:shgrad}
f_{\mathcal{D}_{SC}}[I_1,I_2;\varphi]=&-\left(I_1-I_1\conv-I_2\circ\varphi+I_2\circ\varphi\conv\right)\cdot(\nabla I_2)\circ\varphi.
\end{align}
Note that (\ref{eqn:shgrad}) can also be obtained by interpreting $\beta$ as an independent function $\beta:\mathbb{R}^3\rightarrow\mathbb{R}$, and by applying the SSD distance measure on the image pair \{$I_1(x), I_2(\varphi(x))+\beta(x)$\}. The gradient obtained from the Gâteaux derivative of this measure at $(\varphi, \beta)$,
\begin{equation}
f=-(I_1-I_2\circ\varphi+\beta)\cdot{\choose{(\nabla I_2)\circ\varphi}{1}},
\end{equation}
can be used for an alternating estimation of $\varphi$ and $\beta$. If $\beta$ is estimated from the image instead of computing it with the gradient, i.e. by setting $\beta(x)=I_2\ast\mathcal{G}_\sigma(x)-I_1\circ\varphi\ast\mathcal{G}_\sigma(x)$, we get again (\ref{eqn:shgrad}). The approximation (\ref{eqn:simplif}) can hence be analytically justified by using an alternative formulation of the similarity measure (see \cite{cachier02thesis}, p. 98, for a similar discussion on the CC).

Finally, an iterative algorithm is used to estimate the displacement fields:
\begin{algorithmic}[1]
\While{not converged}
\State compute $f_{\mathcal{D}{SC}}[I_1,I_2;\varphi^{k}]$ and $f_\mathcal{I}[\psi^{k};\varphi^{k}]$
\State compute $f_{\mathcal{D}{SC}}[I_2,I_1;\psi^{k}]$ and $f_\mathcal{I}[\varphi^{k};\psi^{k}]$
\State solve Eqn. \ref{partial1} for $\varphi^{k+1}$
\State solve Eqn. \ref{partial2} for $\psi^{k+1}$
\EndWhile
\end{algorithmic}
The forces are considered as constants for the resolution of the PDEs \ref{partial1} and \ref{partial2}, and the forward and the backward estimation mutually correct themselves at each force update. The PDEs are solved using Red-Black Gauss-Seidel relaxation \cite{briggs08multigrid,NumericalRecipes92} and the full multigrid strategy is used to obtain a linear computational complexity \cite{brandt84localmodeanalysis,briggs08multigrid}. Convergence is achieved if the difference of the $L_2$-norm of the sum of all local forces is below a threshold for both the forward and the backward estimation between two iterations. Note that the algorithms iterate until convergence at each level, deviating thus from the classical multigrid scheme. This is necessary since relaxation is performed on fractional forces. Fixed edges and bending side walls are used as boundary conditions on the registration domain \cite{Modersitzki04numerical}. 

%
%
%
%
%
\subsection{Parameter settings}

A recurring problem in variational image registration frameworks is the proper scaling of the energy terms. In the presented approach, line searches in the inverse gradient direction are performed within a limited perimeter in order to obtain locally exact solutions for the image distance and the inverse consistency forces. The forces obtained are then summed together, i.e. they are equally weighted but not averaged. The compressibility constraint of the linear elastic tensor poses some problems in the presented framework: since we are operating with fractional forces, it behaves like a strong shape constraint and registration will stall when it is enabled from the beginning. We therefore allow the deformation estimation to converge with the compressibility constraint disabled (Poisson's $\nu = 0$), and turn it on only in a second run where local over-stretching and squeezing is corrected ($\nu = 0.48$). Young's modulus $E$ has no physical meaning in this approach since fractional displacements are used as artificial forces. We set both $E$ and the PDE discretization $\delta t$ to 0.5. The final displacement $\delta \varphi(x)^{k+1}$ that is added to $\varphi(x)^k$ is finally limited to $||\delta \varphi(x)||<1$ to ensure that no jumps over intensity barriers are possible. This constraint greatly improves the overall numerical stability.


\section{Clinical application}

\begin{figure}
\centering
\includegraphics[width=0.6666\textwidth]{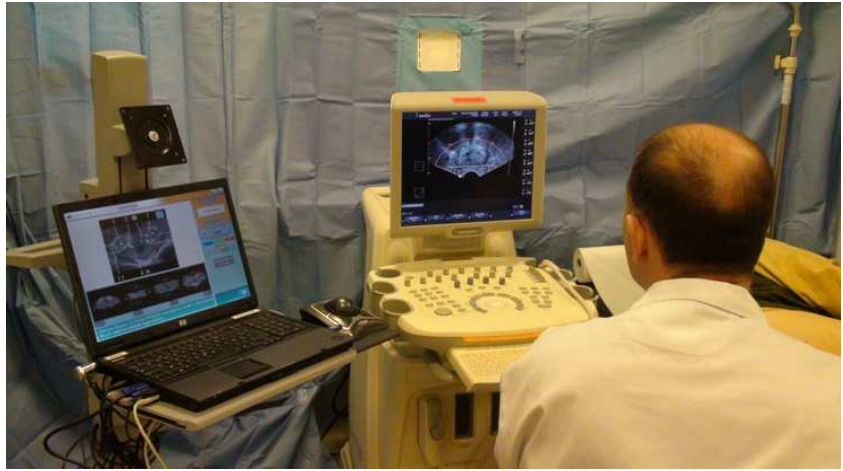}
\caption{Application prototype. The figure shows a prototype of the biopsy assistance application during a biopsy session.}
\label{fig:urostation}
\end{figure}

\begin{figure}
\centering
\includegraphics[width=1\textwidth]{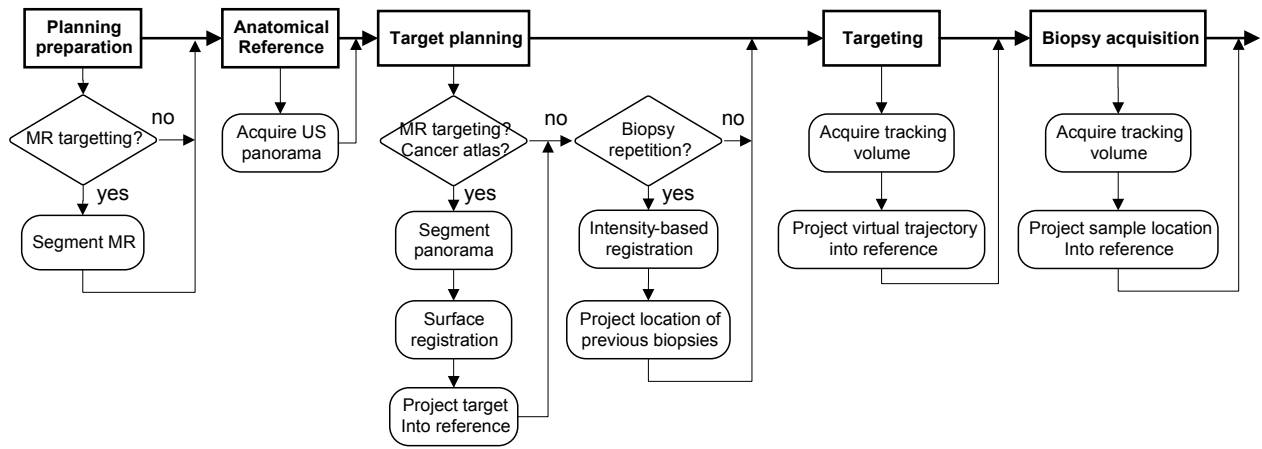}
\caption{Clinical flow chart of the biopsy application. Biopsy planning and acquisition phase.}
\label{fig:flowchart}
\end{figure}

\begin{figure}
\centering
\includegraphics[width=0.7\textwidth]{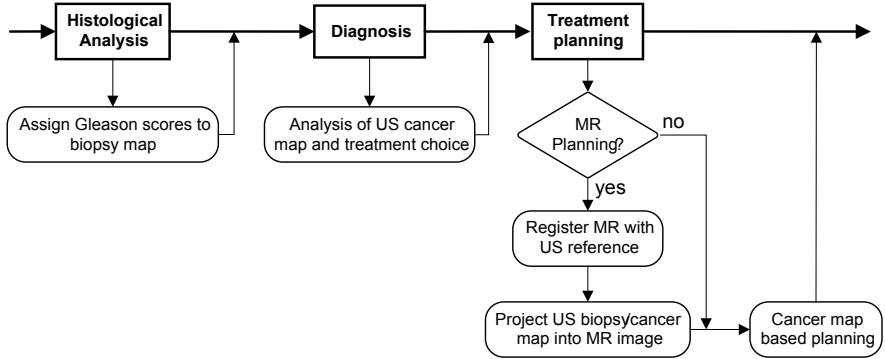}
\caption{Clinical flow chart of the biopsy application. Histological analysis, diagnosis and therapy planning phase.}
\label{fig:flowchartpost}
\end{figure}

In the previous sections we proposed a tracking method for prostate motion using 3D TRUS. In this section a clinical application for prostate biopsy tracking and guidance, based on this method, will be presented. The primary objective of this application is to combine clinical accuracy with an intuitive user interface and a lightweight clinical work-flow. In particular, care was taken such that no logistical and interventional overhead is added to the current standard clinical procedure. The application aims to provide solutions for the clinical issues of TRUS prostate biopsies enumerated in \ref{sec:clinicalissues}. This includes the ability to target suspicious lesions identified on MR images, to visualize locations sampled during a previous biopsy session, and to provide interventional biopsy maps as well as post-interventional cancer maps. Fig.~\ref{fig:flowchart} gives a summary of the clinical work-flow for biopsy acquisition, while Fig.~\ref{fig:flowchartpost} describes the post-biopsy work-flow from histological analysis to therapy planning. The different steps are described in detail in the following sections. An illustration of the clinical setup is given in Fig.~\ref{fig:urostation}.


\subsection{Panoramic volume as anatomical reference}
\label{sec:panorama}

The first part of the clinical protocol is performed a few minutes prior to biopsy collection, and consists in the acquisition of the anatomical reference. Unfortunately, a single 3D volume of the prostate does not typically contain the entire gland because of the pyramidal form of the 3D US beam that cuts the gland at its lateral borders and, in particular, near the probe head where most tissue samples are taken. For this reason, three volumes are acquired from different positions, elastically registered and compounded into what we name a \textit{panorama volume}. The bounding ellipsoid required for the kinematic model is defined by the clinician after the acquisition of the first volume.

\subsection{Biopsy planning}

The second part of the protocol consists in the definition of non-ultra\-sound biopsy targets in the anatomical reference. This step needs to be simple and rapid since it is not convenient for the clinician to interact with the computer while holding a probe, and the patient discomfort increases with the duration of the rectal penetration. The clinician therefore first identifies the targets, then he inserts the probe and acquires the reference volume, and finally the software registers the targets with the reference volume. If the target registration requires user interaction, the probe can be held in place with an articulated arm in order to free the physician's hands.

\begin{figure}
\centering
\hfill
\subfigure[]{\includegraphics[width=0.25\textwidth]{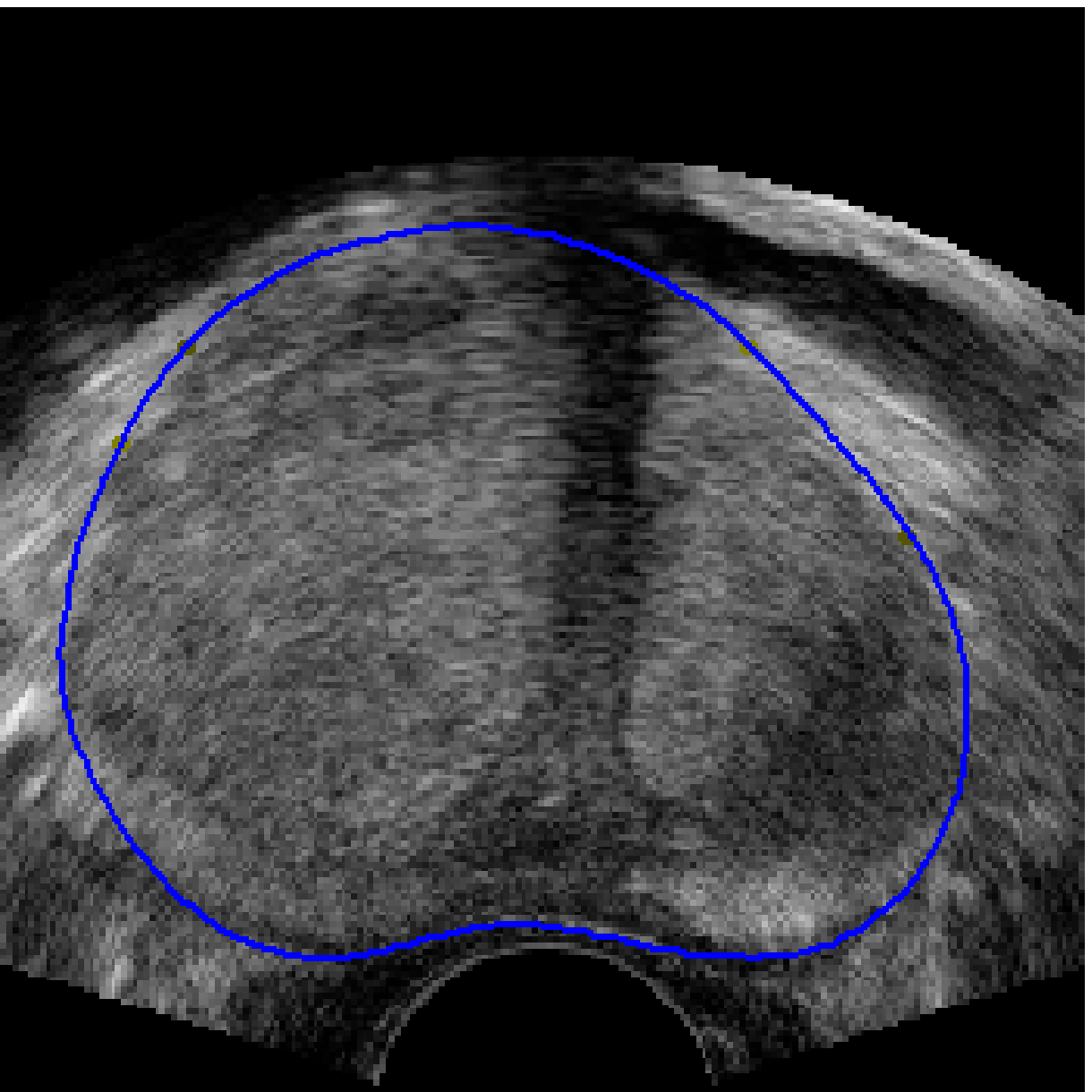}}\hfill
\subfigure[]{\includegraphics[width=0.25\textwidth]{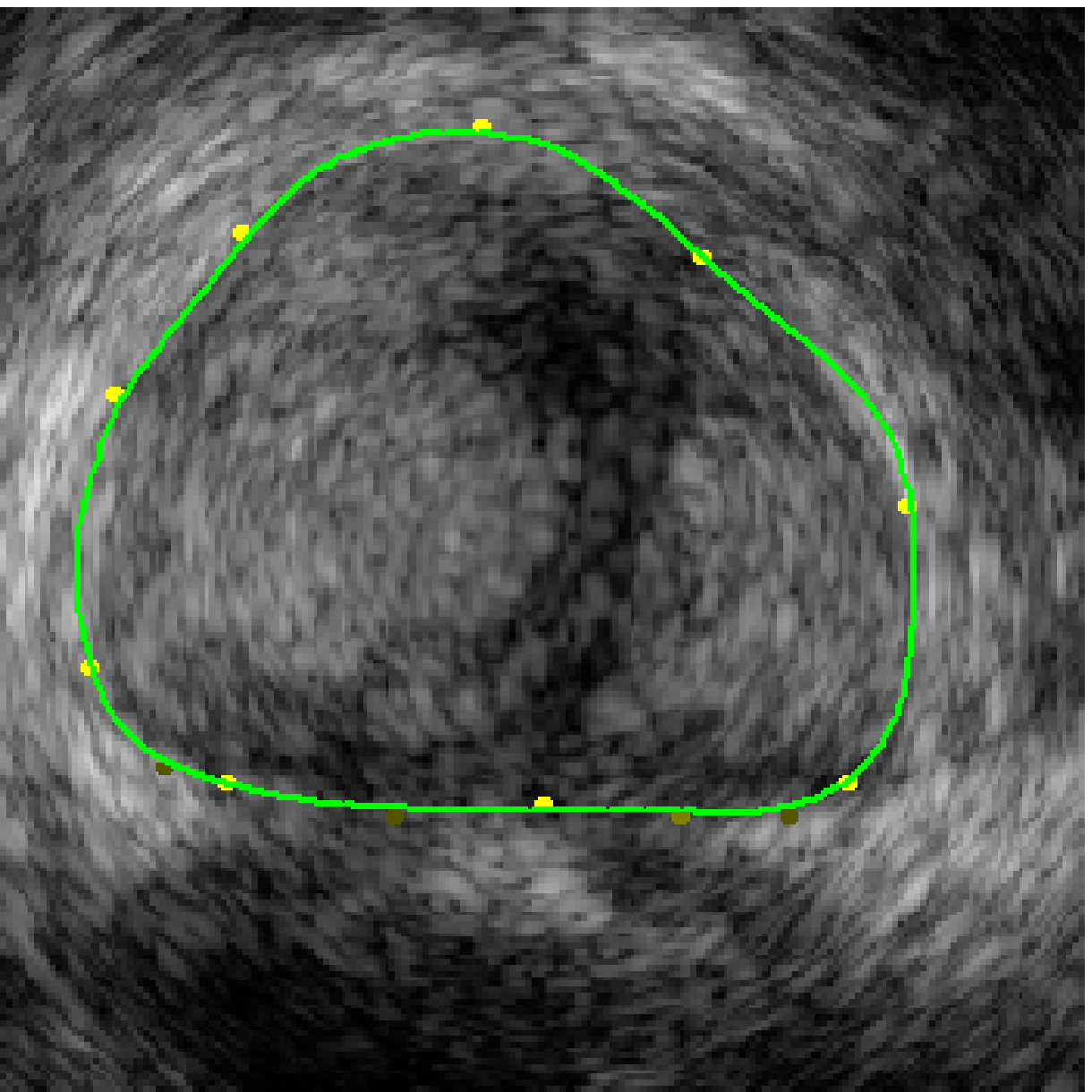}}\hfill
\subfigure[]{\includegraphics[width=0.25\textwidth]{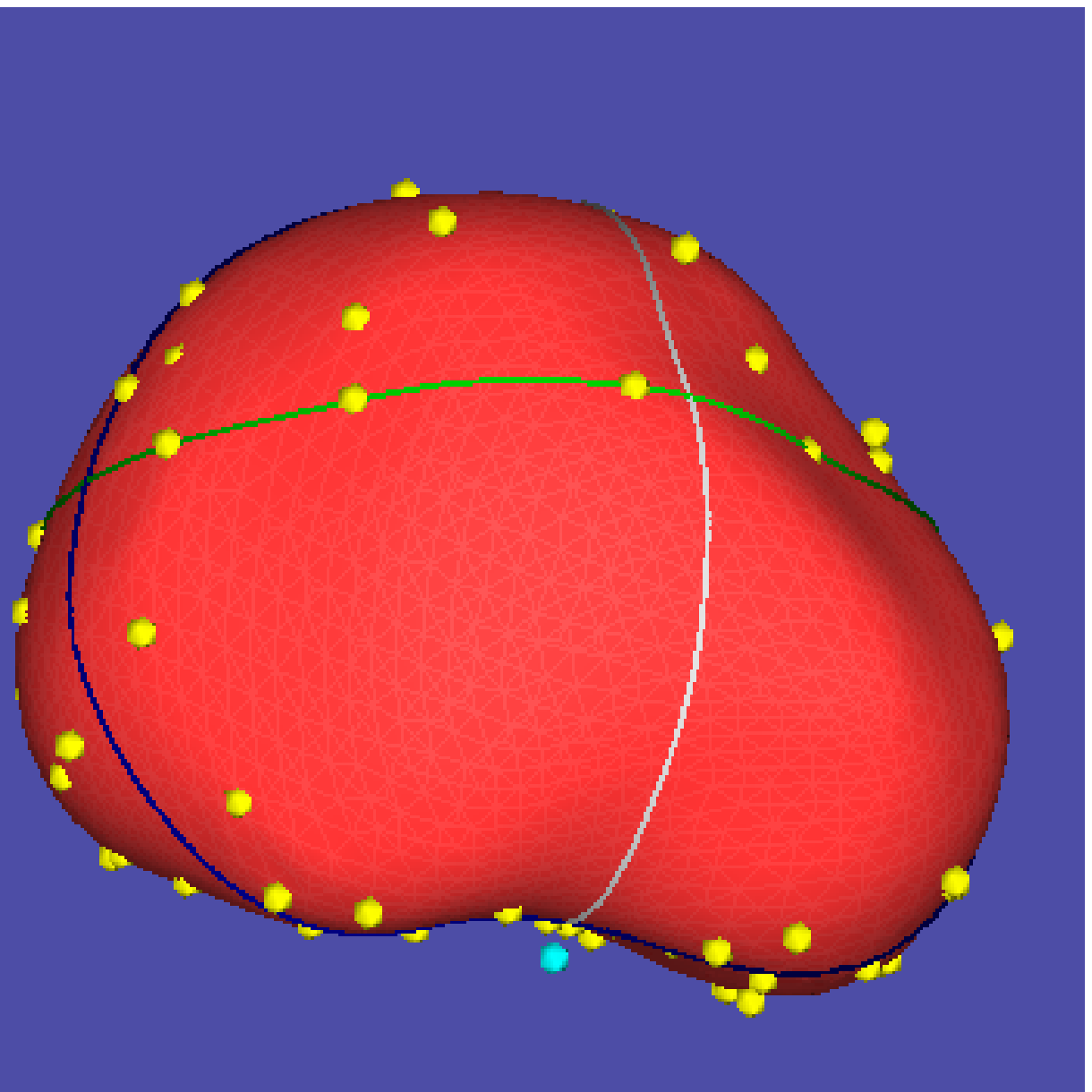}}\hfill
\subfigure[]{\includegraphics[width=0.25\textwidth]{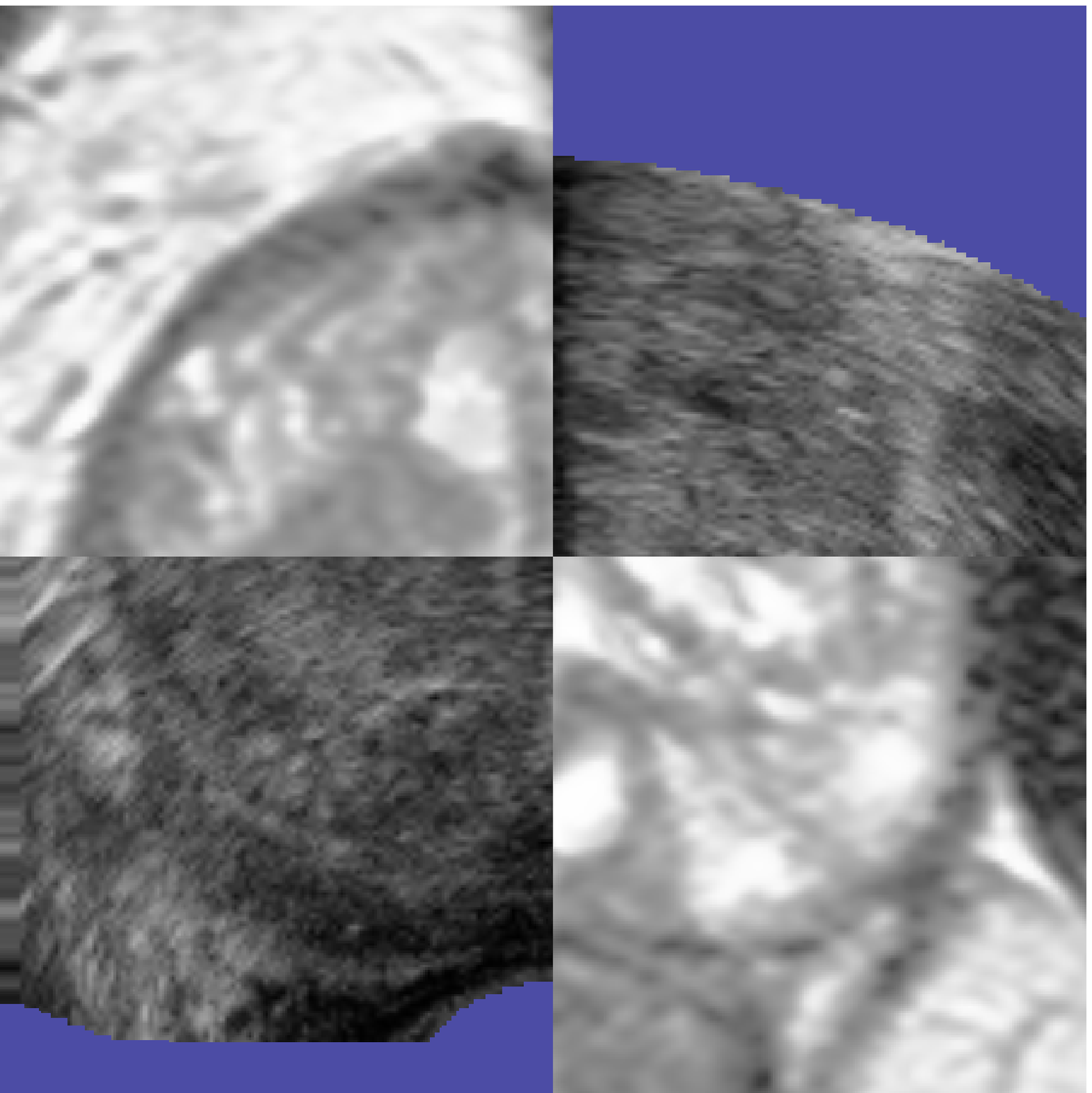}}
\caption{Planning phase. Fig.~(a) shows the transversal and (b) the coronal semi-automatic segmentation result. Fig. (c) shows the 3D representation of the segmentation including the manually segmented points (yellow). Fig.~(d) finally shows the result of non-linear surface registration between the MR image and the US reference.}
\label{fig:planning}
\end{figure}

In the concrete case of MR lesion targeting, the segmentation of the prostate in the US reference, required for MR-US registration, needs to be performed rapidly for the same reasons. We use the algorithm proposed by Martin et al. in \cite{martin10isbi} to perform MR to 3D TRUS prostate volume fusion. This algorithm uses the MR image prostate segmentation as shape prior for a fully automatic detection of the prostatic capsule in 3D US volumes. The MR shape is automatically obtained with a probabilistic atlas and a spatially constrained deformable model, using the method presented by Martin et al. in \cite{martin10medphys}, cf. Fig.~\ref{fig:planning}. After MR-US fusion, the segmented MR target can be projected into the anatomical reference of the tracking system. The application also provides the facility to project biopsy maps into an MR image, using the same techniques.

In the case of the repetition of a biopsy series, it is interesting to know the exact location of previously acquired samples. This is performed by registering the anatomical reference of the previous intervention with the newly acquired reference using the elastic tracking algorithm presented in Sec.~\ref{sec:tracking}. However, we do not yet have enough data on repeated biopsies to prove the viability of this approach. The time lapse between two series can be important, and the organ can be altered in between, for instance because of tumor growth. Also, the involved imaging hardware can be different or differently configured, which could yield dissimilar images. If purely image-based registration should lack stability, a mixed image and surface-based registration will most certainly be sufficient.

Finally, it could be interesting to indicate targets with a high probability of carcinoma presence. This can be performed by registering a statistical cancer atlas, such as that developed by Shen et al. \cite{shen04atlas}, with the prostate shape segmented in the reference volume. This could lead to a better statistical sampling and could potentially reduce the number of biopsy acquisitions per session, see also the optimal protocol for transperineal biopsies developed by Shen.

\subsection{Biopsy acquisition}

After registration and projection of the targets, the biopsy core acquisition phase is started. Current 3D US scanners can acquire 0.5-5 volumes per second, depending on the resolution. However, continuous 3D volume streams cannot be processed at the same frame-rate with the current implementation of the prostate tracking algorithm. Also, the higher frame-rates of 2D US are visually more pleasing. As a consequence, 2D US is used for needle placement. Single 3D volumes are acquired only when positional information about the targeted or collected sample is required. Volume acquisition can be triggered out of 2D US mode with a foot pedal. The clinician's hands are free for tasks like holding the probe and the biopsy gun. The volumes are automatically transferred to the application via TCP/IP and DICOM and after reception they are automatically registered. The puncture path that corresponds to the probe position when the volume was acquired is then projected into the reference volume and visualized with the planned targets. This makes it possible to validate and correct the puncture trajectory before firing the biopsy gun.

\begin{figure}
\centering
\hfill
\subfigure[]{\includegraphics[width=0.33\textwidth]{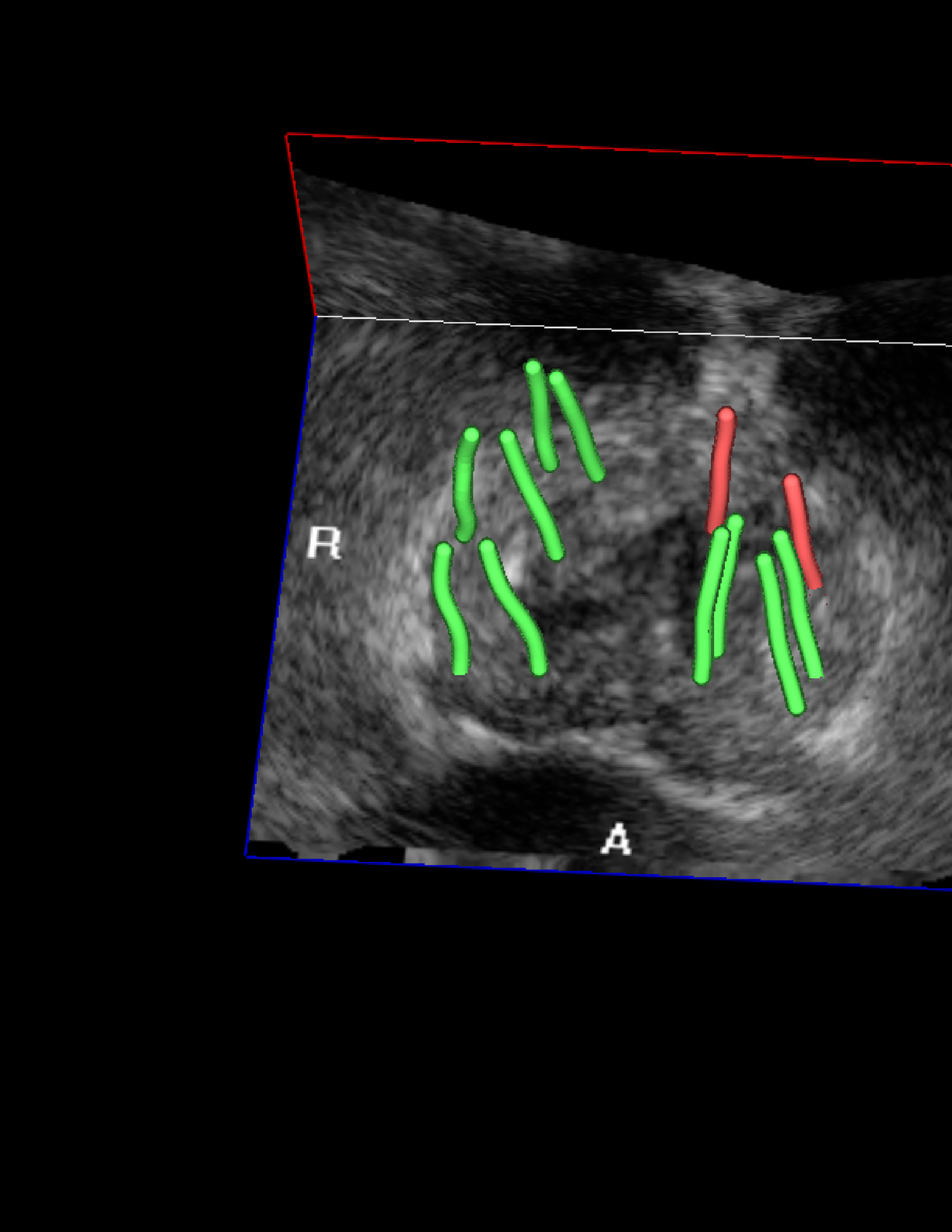}}\hfill
\subfigure[]{\includegraphics[width=0.33\textwidth]{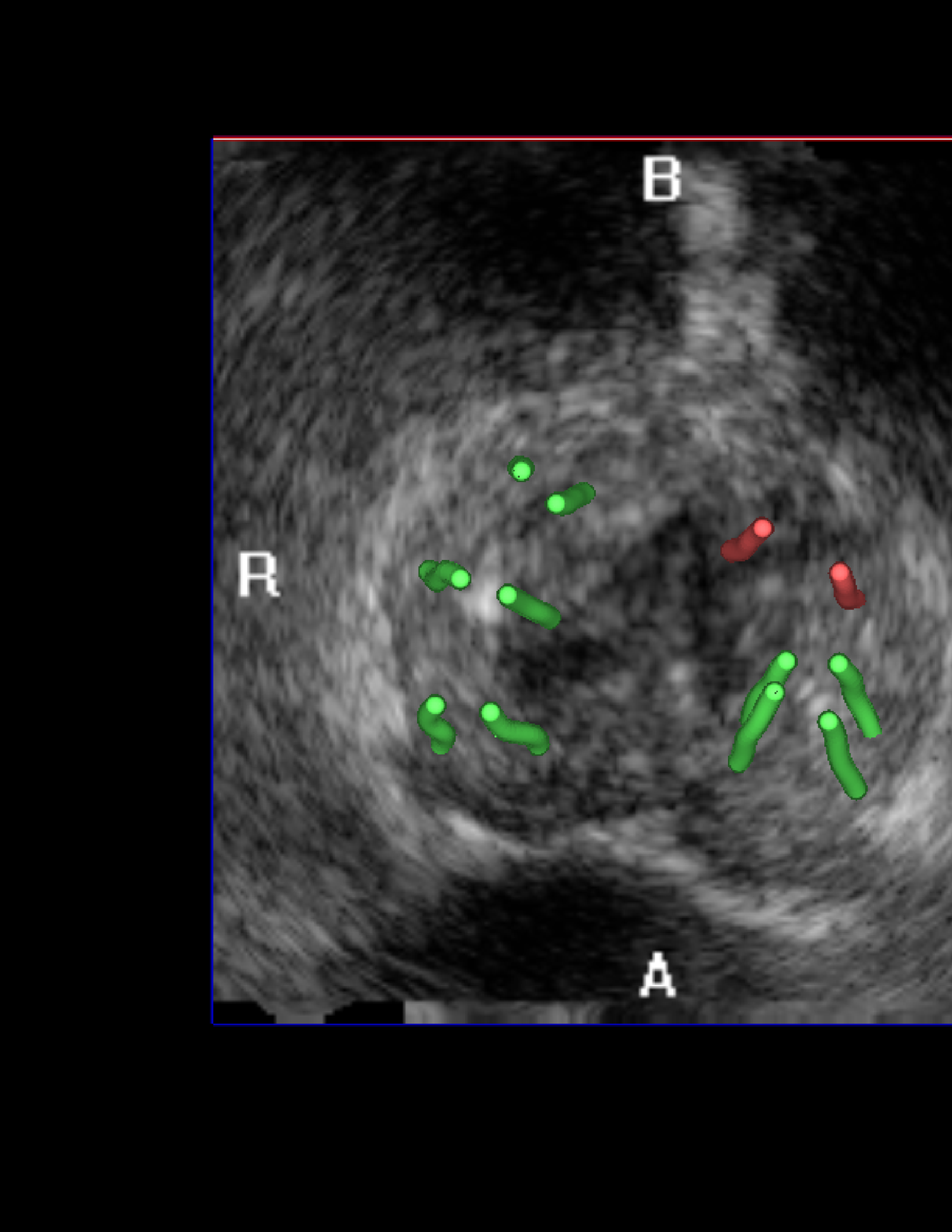}}\hfill
\subfigure[]{\includegraphics[width=0.33\textwidth]{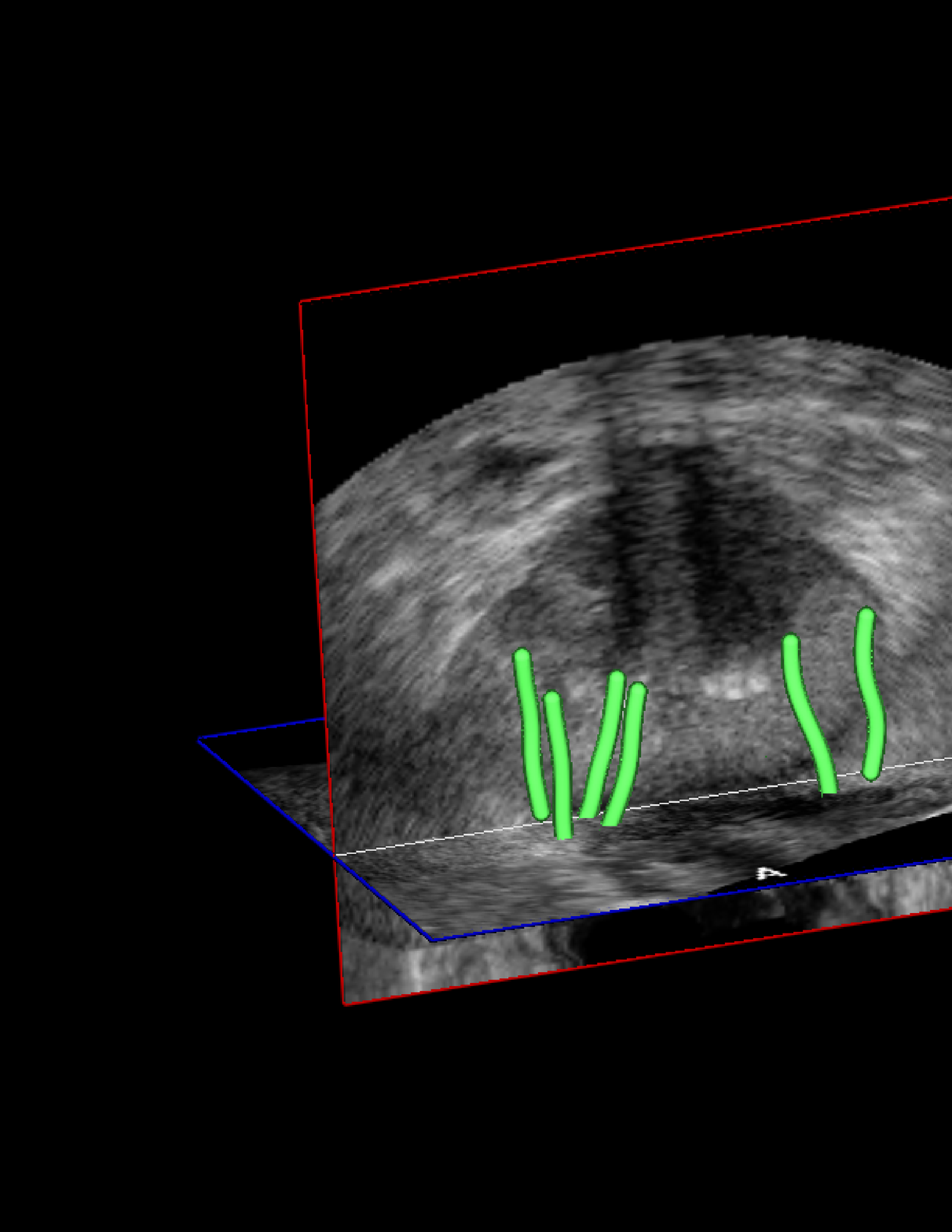}}
\caption{Biopsy maps. Fig.~(a)-(c) show different views of a biopsy map with color-coded Gleason score.}
\label{fig:biopsymap}
\end{figure}

When the needle is correctly placed, the tissue sample is collected by triggering the spring needle gun. Before removing the needle, the clinician acquires a US volume containing the needle image. The needle is then automatically segmented, if present. It is projected into the reference anatomy, giving the clinician an immediate feed-back about the sampling position with respect to previous biopsies. The biopsy distribution can hence be assessed during the intervention and additional samples can be collected, if necessary. A typical interventional biopsy map is given in Fig.~\ref{fig:biopsymap}.

\subsection{Multi-modal biopsy and cancer maps}

After acquisition, the samples are sent to the pathologist for histological examination. The Gleason score, a visual tissue grading scale that is correlated with the aggressiveness of the carcinoma, is determined for each sample and entered into the biopsy application. A color code is used to visualize the cancer grade of a sample in the biopsy maps. This provides an instant overview over the cancer distribution for diagnosis and treatment planning.

With the presented system it is possible to localize the biopsy samples accurately. This capacity could help to improve existing treatment strategies and might be useful for experimental focal therapy approaches. It can be interesting, for example, to project the cancer map onto a MR planning volume. This can be done using the same approach as the MR target projection onto the US reference.

\section{Experiments and results}
\label{sec:experiments}

The tracking system was tested on real patient images that were acquired during biopsy sessions at the Pitié Salpétrière hospital, Paris, France. The patient study was approved by the ethical committee of the Pitié Salpétrière and was performed with the consent of each patient. The volumes were acquired with a GE Voluson equipped with an endorectal RIC5-9 probe. Three volumes were acquired a couple of minutes before biopsy acquisition for the creation of a panorama volume that contains the entire gland (see \ref{sec:panorama}). Interventional volumes were acquired after each biopsy shot with the needle left inside. Besides these points, the standard clinical protocol was not altered. Registrations were performed on a standard PC with 4GB of RAM, 3GHz processor frequency and two cores. The acquired volumes had a resolution of 199$^3$ voxels. A 5-level multiresolution pyramid was used for registration, the resolution on the coarsest level being 13$^3$ voxels.

\subsection{Kinematic model}
\begin{figure}
\subfigure[]{\includegraphics[width=0.32\textwidth]{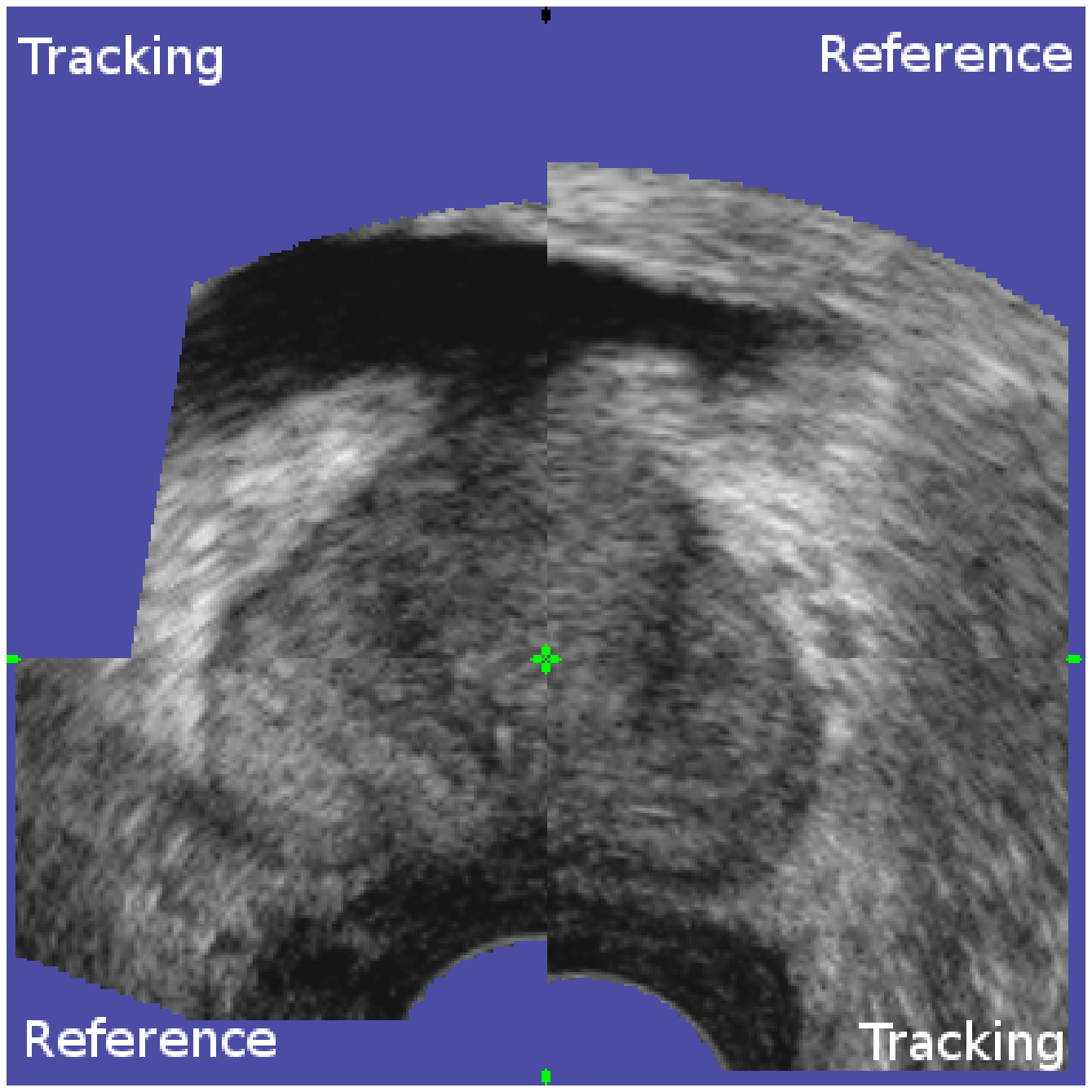}}\hfill
\subfigure[]{\includegraphics[width=0.32\textwidth]{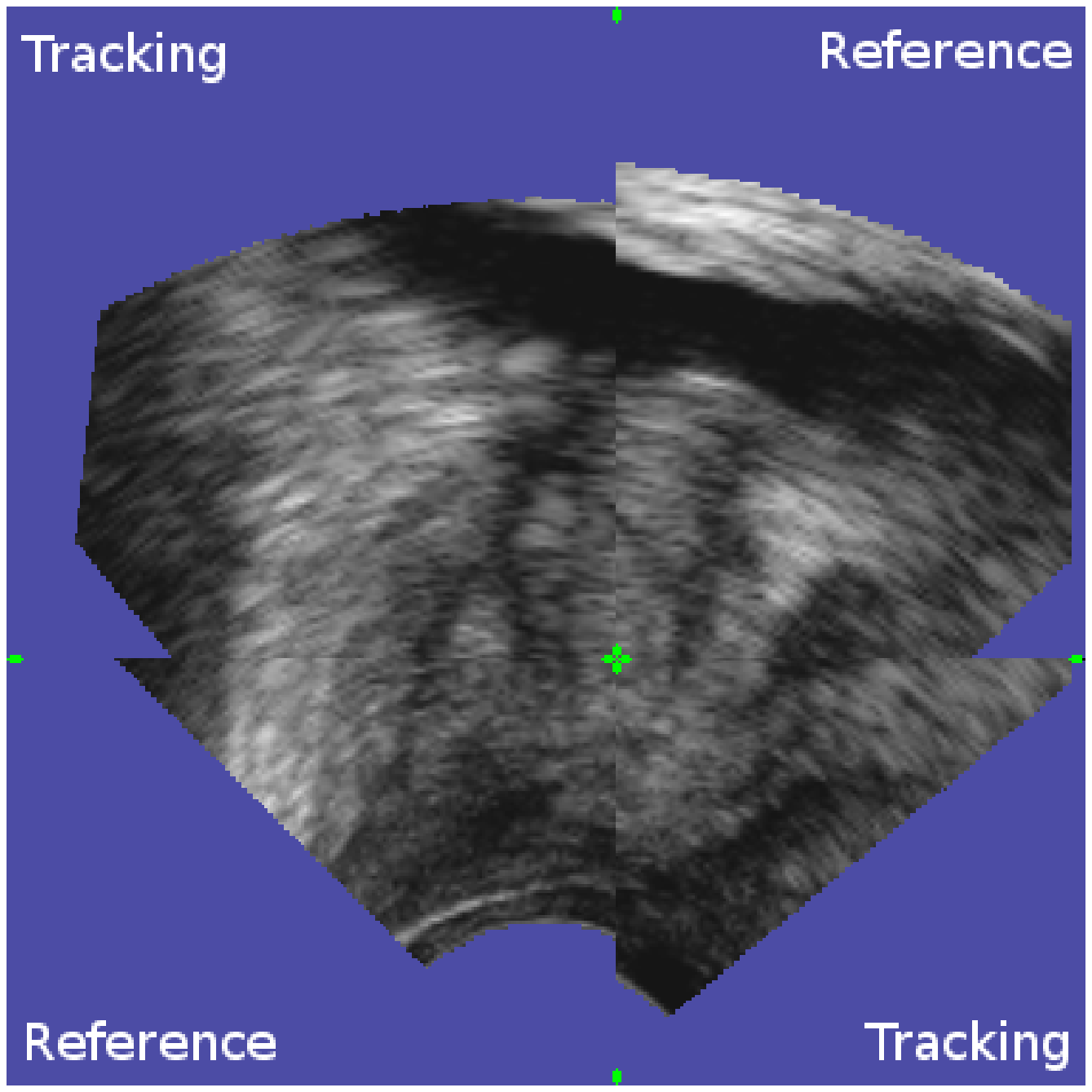}}\hfill
\subfigure[]{\includegraphics[width=0.32\textwidth]{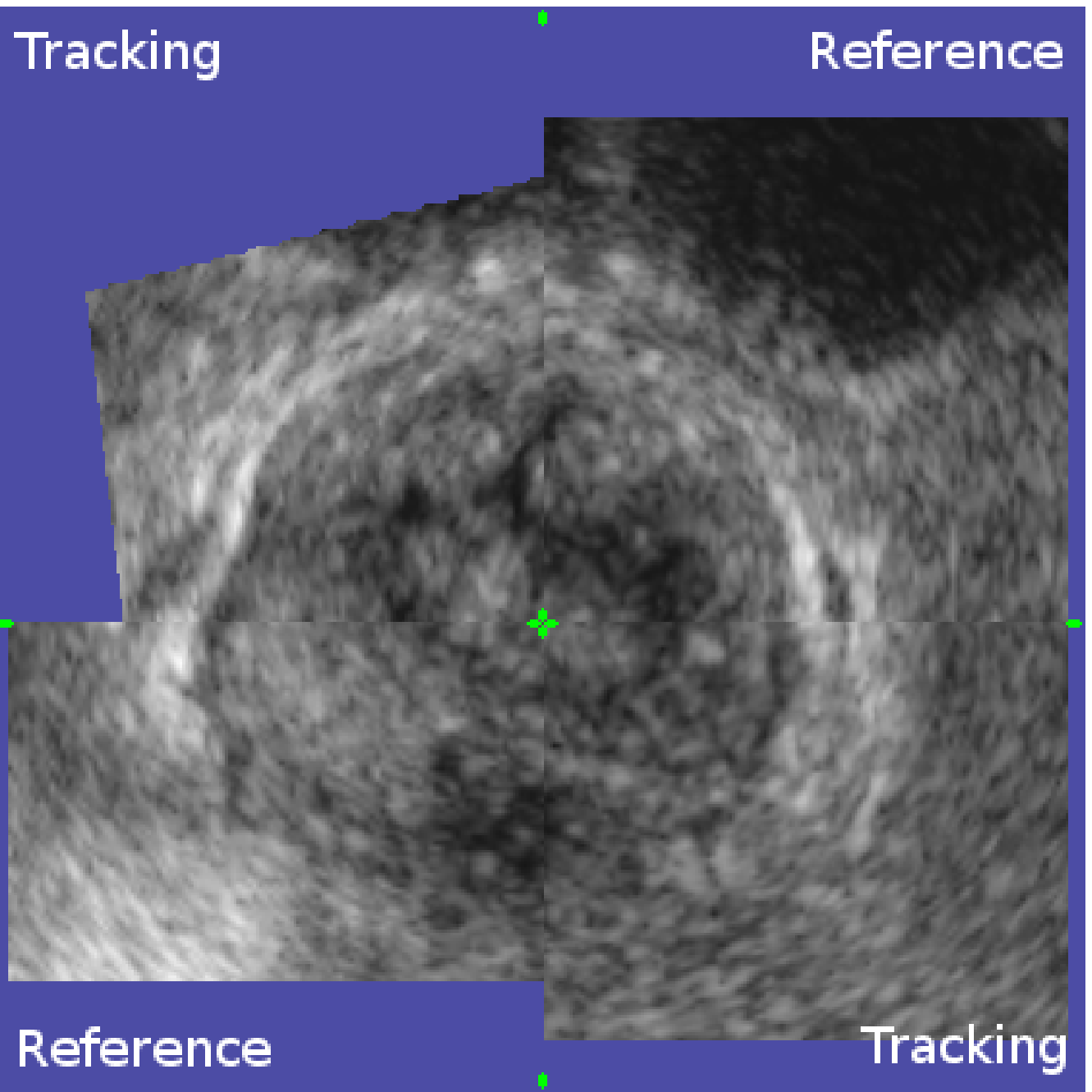}}
\caption{Viewer for visual validation. The viewer shows transverse (a), sagittal (b) and coronal cuts (c) simultaneously. The registration volumes are overlaid: the upper-left and lower-right corners of each view show the tracking volume, while the lower-left and upper-right corners show the reference volume. The green cut point is the 3D point where the three image planes intersect. The user can move the cut point freely in each view, allowing him to explore the entire volume (changing the point in one view changes the spatial position of the other two views).}
\label{fig:volumeviewer}
\end{figure}

The first experiment evaluates the capacity of the kinematic probe movement model to estimate a probe position sufficiently close to its real position with respect to the gland. For this test, 786 registrations of the biopsy volumes of 47 patients were performed with the corresponding panorama volumes. The registrations were then visually validated either by the clinician immediately after biopsy acquisition or off-line by one of the authors. This was carried out using a volume viewer that allows to overlay and to explore the reference and the tracking volume after application of the registration transformation (see Fig.~\ref{fig:volumeviewer}). Note that the accuracy of a large and randomly chosen subset of the registrations classified as valid was evaluated to 0.8$\pm$0.5~mm, cf. the accuracy study presented in Sec.~\ref{sec:tre}, which indicates that only few registrations were falsely identified as correct. In this study, 769 (97.8$\%$) volumes were classified as valid, and 17 (2.2$\%$) were classified as failures. The 17 failures occurred with volumes that did not contain enough information about the prostate, i.e. the tracked object was literally "out of view". This was caused by inadequate US depth or probe pressure (prostate capsule not visible) in 11 cases, partial probe contact with the rectum in 1 case, extremely lateral volumes containing only a small part of the prostate in 4 cases and an incomplete panorama in 1 case. Note that the failures were not caused by patient movements.

\begin{figure}
\subfigure[]{\includegraphics[width=0.32\textwidth]{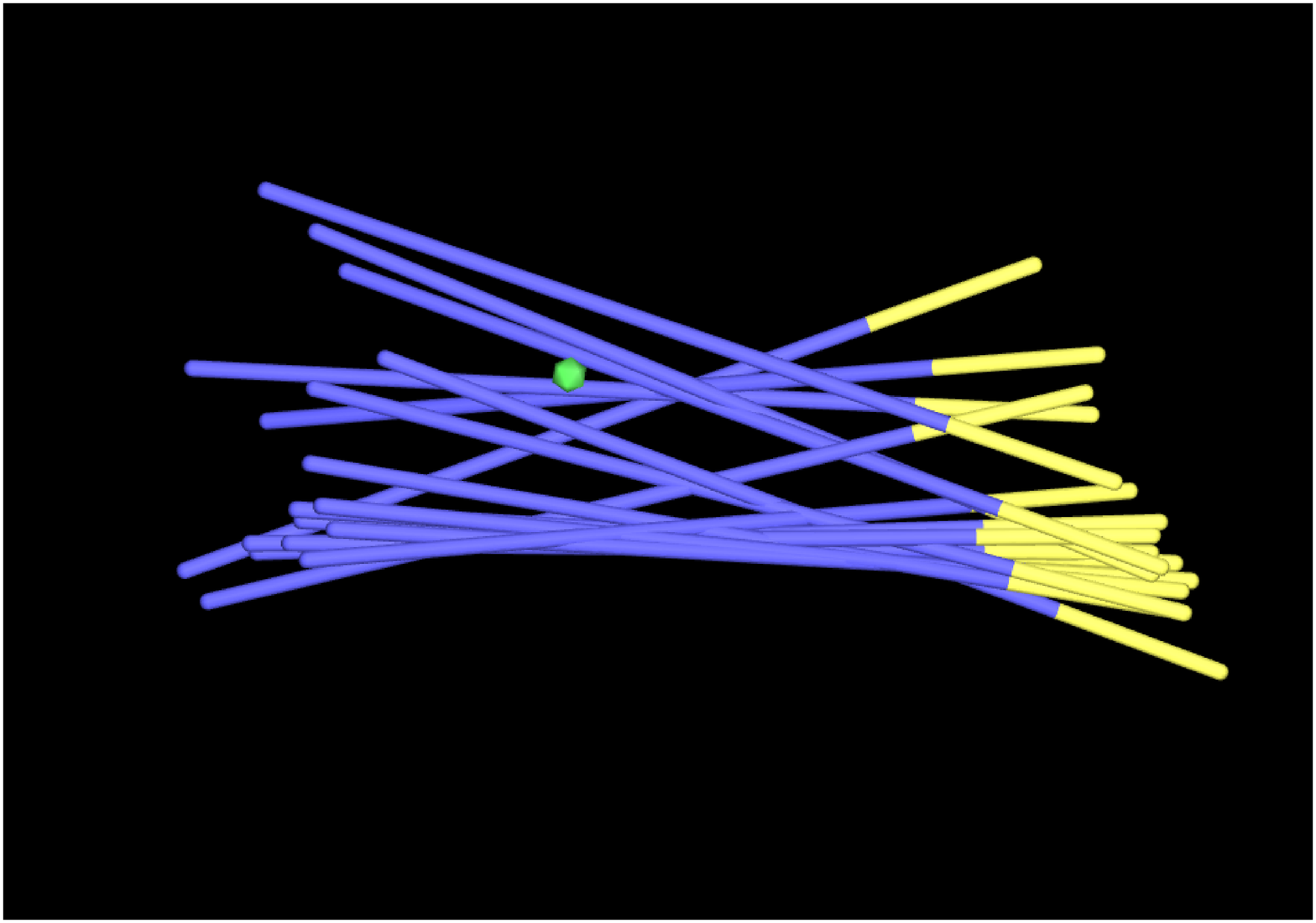}}\hfill
\subfigure[]{\includegraphics[width=0.32\textwidth]{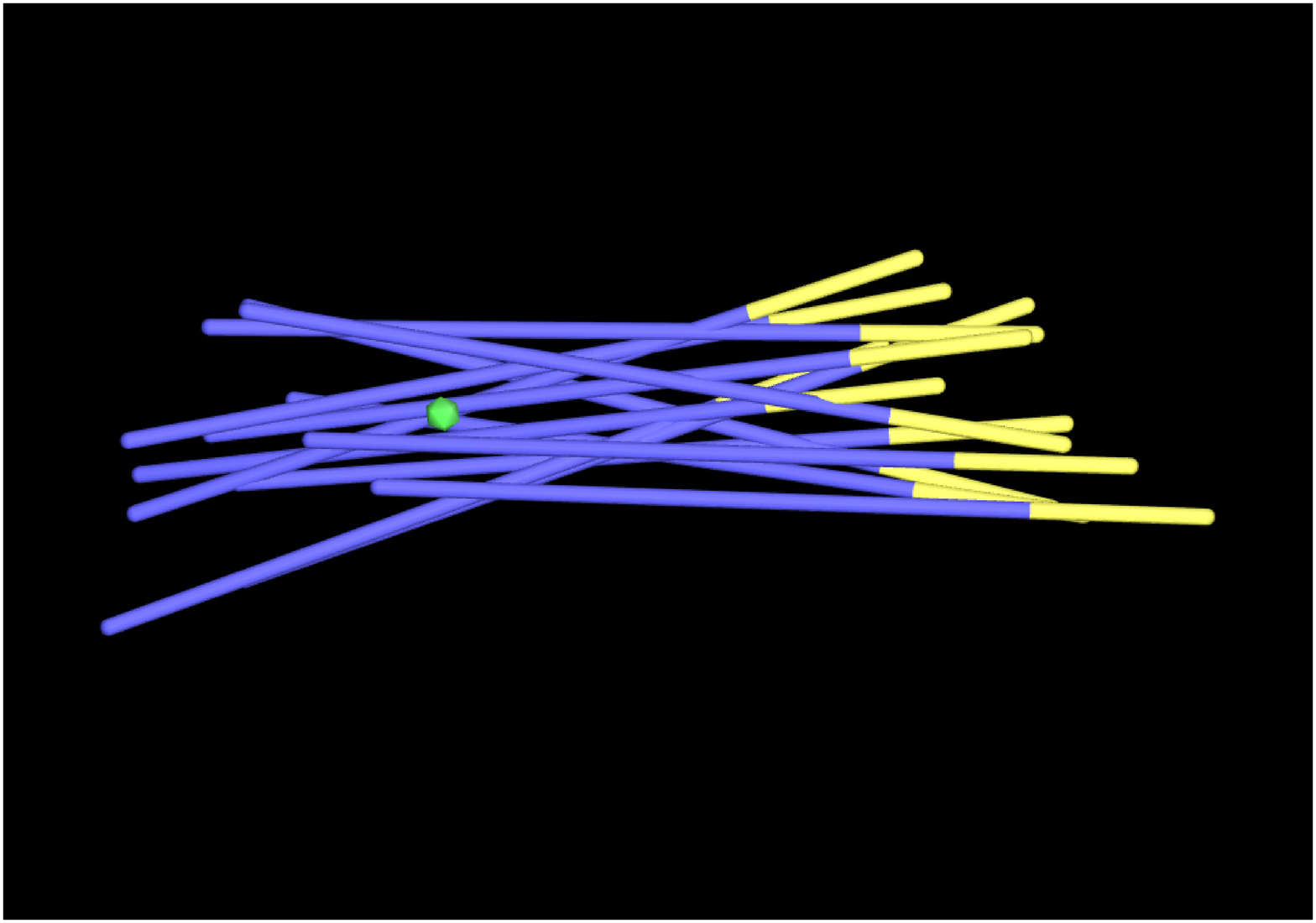}}\hfill
\subfigure[]{\includegraphics[width=0.32\textwidth]{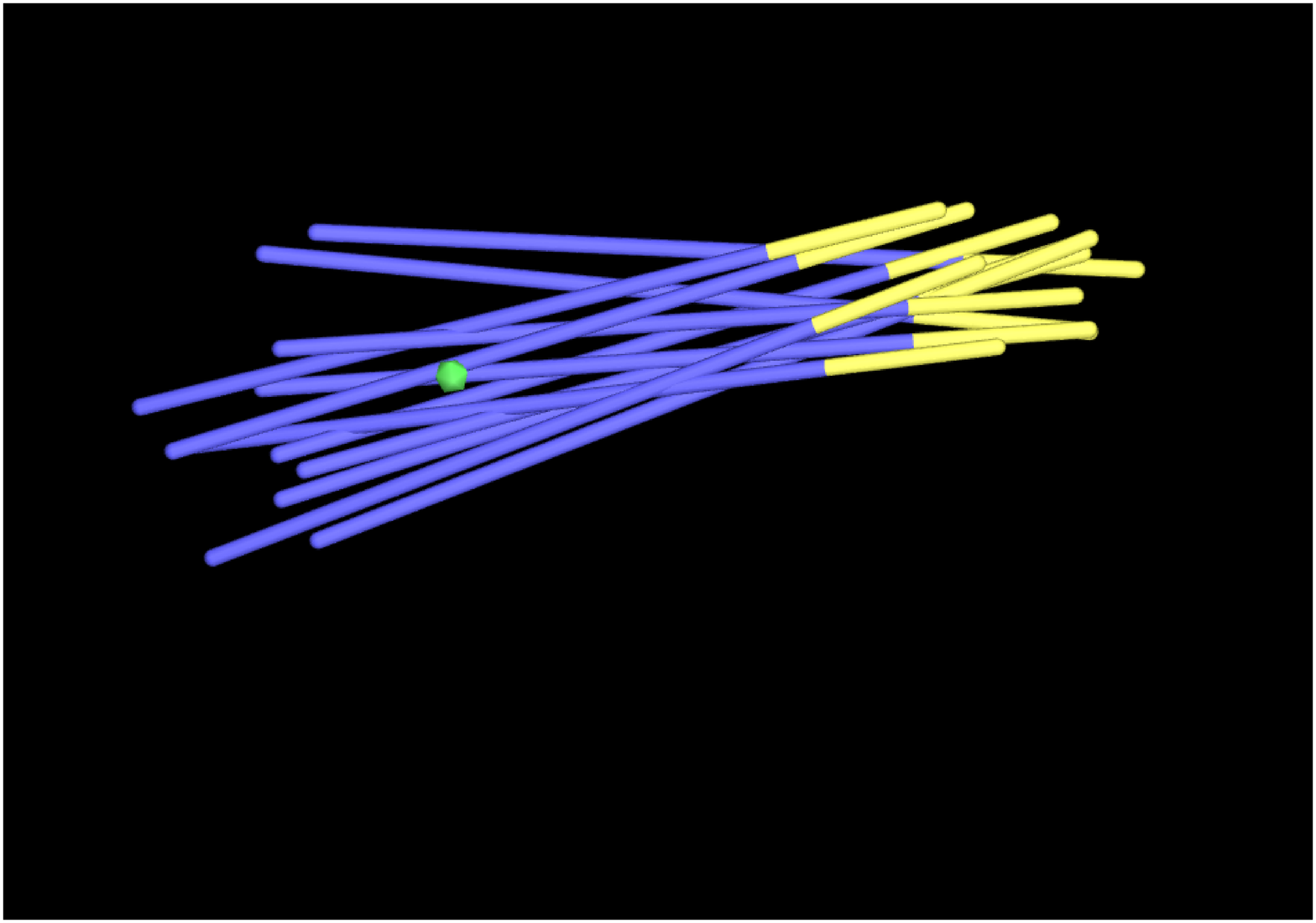}}
\caption{Probe positions. The Fig. (a) to (c) show the probe positions during biopsy acquisition for three different patients, projected into panorama space after registration. The interface between blue and yellow tubes corresponds to the position of the transducer origin, the yellow tips correspond to the contact point of the probe with the rectum. The green sphere is the a priori fixed point. One can see that the angular error of the probe position predicted by the kinematic model stems from the fact that the fixed point model of the rectal sphincter is rather approximate.}
\label{fig:typicalprobeposition}
\end{figure}

A second experiment studies the role and performance of the kinematic model (pipeline step 1 in Fig.~\ref{fig:pipeline}) with respect to local rigid registration (pipeline step 2). We therefore compute the angular and the Euclidean distances of the best transformation predicted by the model from the transformation estimated with local rigid registration to see how accurate predictions of step 1 are and how much the second step improves these predictions. The best prediction of the model is defined as the transformation that is closest to the transformation produced by step 2. The distance between two transformations is defined as the root mean square (RMS) Euclidean distance of the 6 intersection points of the probe head with the canonical coordinate axis, centered at the transducer origin, after respective application of the transformations. The study was performed on 269 successful volume registrations stemming from 14 biopsy sessions. The ellipsoid surface was discretized using a 16$\times$16 grid, and the 360° rotational space around the probe axis was discretized using 24 steps. The model was evaluated at five different depths. The mean Euclidean distance of the transducer array origin was 2.0$\pm$1.5~mm, and the mean angular distance was 9.7$\pm$5.1°. Concerning the five different depths at which the model was evaluated, 15$\%$ of the best predictions were found at an offset of -10~mm, 23$\%$ at -5~mm, 35$\%$ at 0~mm, 21$\%$ at +5~mm and 6$\%$ at +10~mm. The model hence predicts the position of the transducer array quite precisely, which indicates that it is relatively robust with respect to the accuracy of the user-defined ellipsoid. Note that evaluating multiple depths can compensate ellipsoid placement errors in the probe axis direction. The angular predictions are less accurate, which stems from the resolution of the rotational space around the probe axis of 15° and from the fact that using an a priori fixed point as a model for the constraints exerted by the rectal sphincter is rather simplistic. After rigid registration, the average distance of the probe axis from the a priori fixed point was 4.3$\pm$3.1~mm. Fig.~\ref{fig:typicalprobeposition} shows typical probe positions during biopsies with respect to the fixed point for three different patients. In conclusion, the results given in this paragraph reflect the trade-off between accuracy and speed of the kinematic model. The role of the kinematic model is to initialize local rigid registration efficiently. However, to achieve clinically acceptable accuracy, local registration is mandatory.

\subsection{Target registration error}
\label{sec:tre}
\begin{figure}
\subfigure[]{\includegraphics[width=0.24\textwidth]{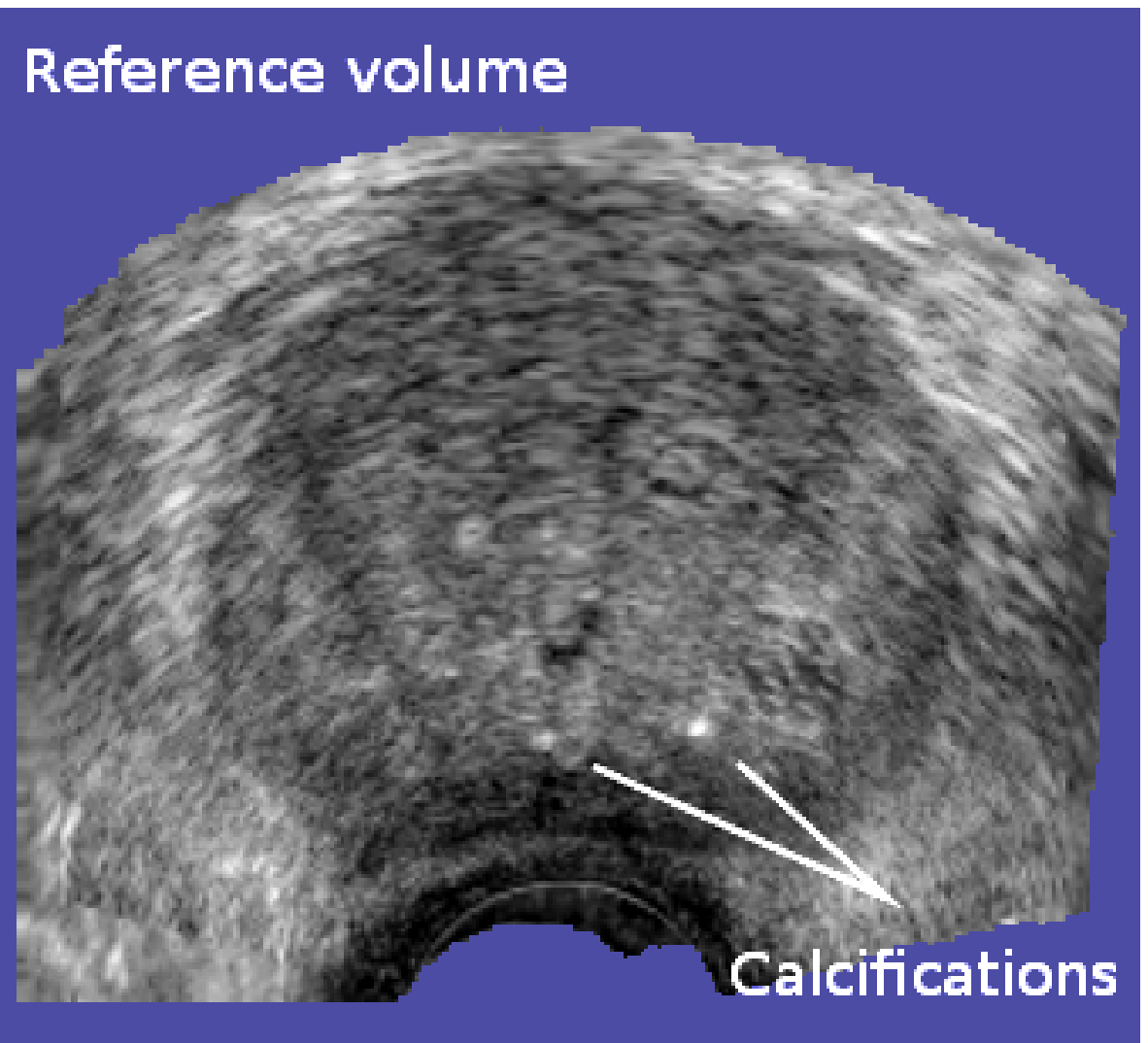}}
\subfigure[]{\includegraphics[width=0.24\textwidth]{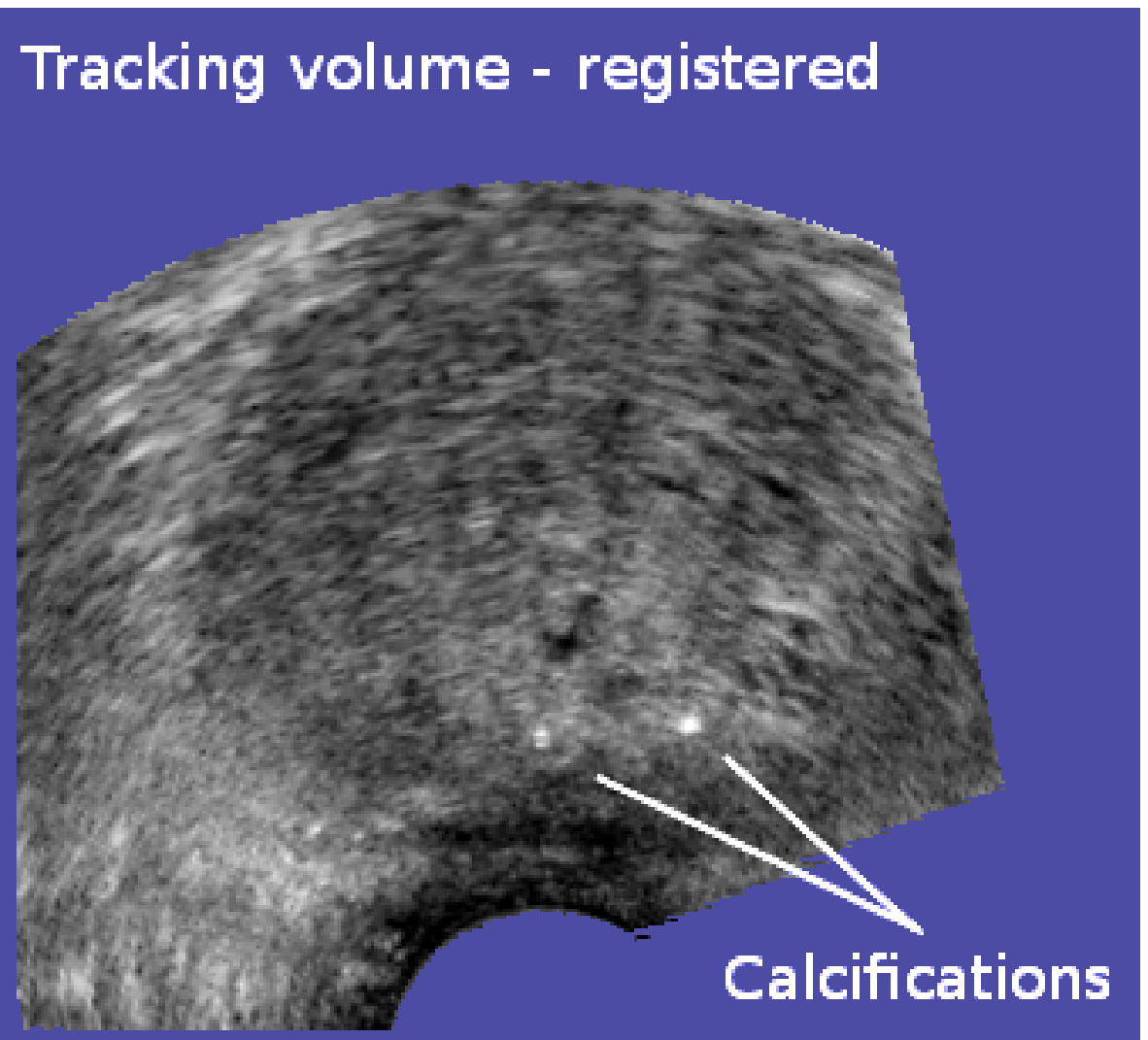}}
\subfigure[]{\includegraphics[width=0.24\textwidth]{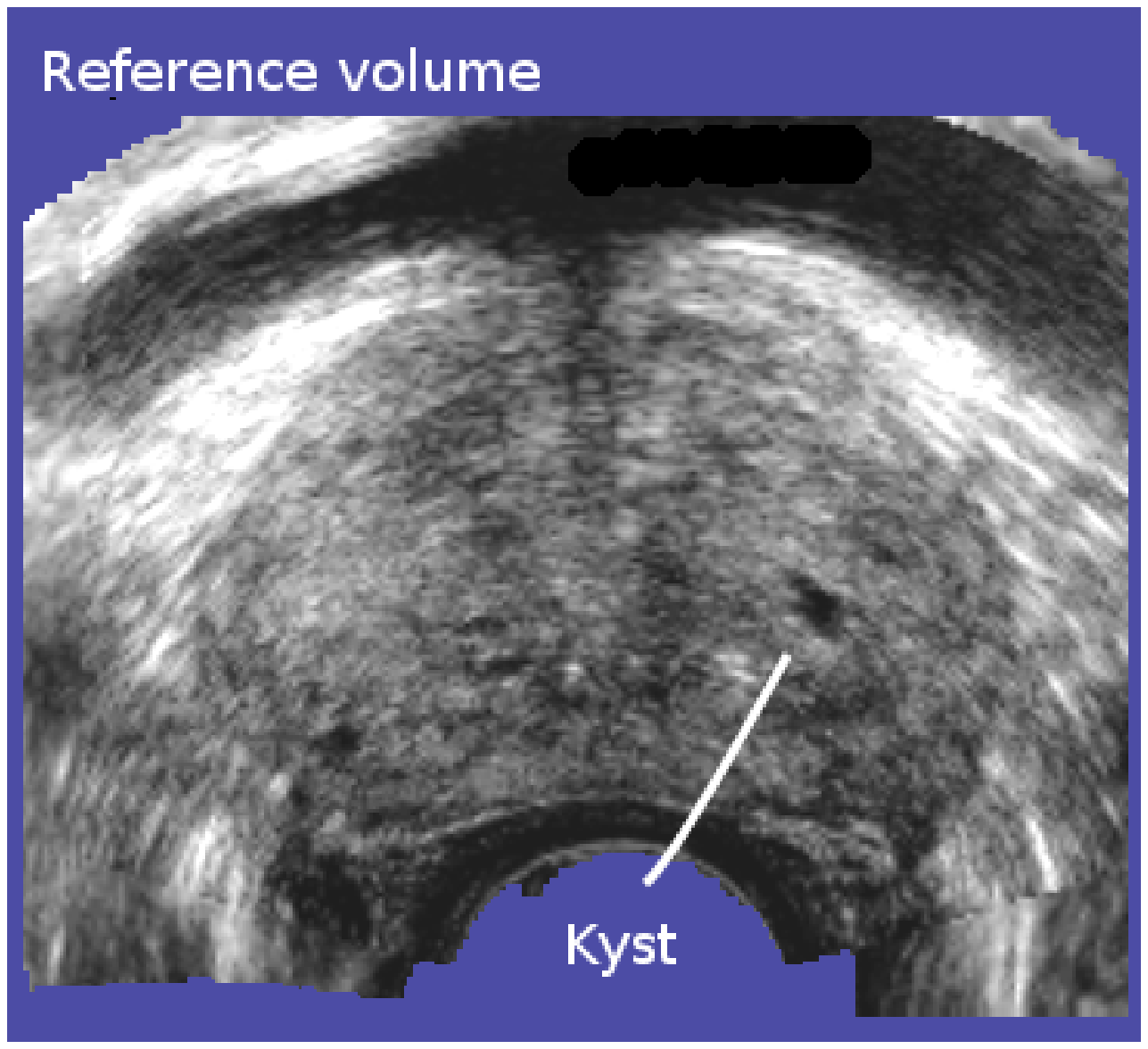}}
\subfigure[]{\includegraphics[width=0.24\textwidth]{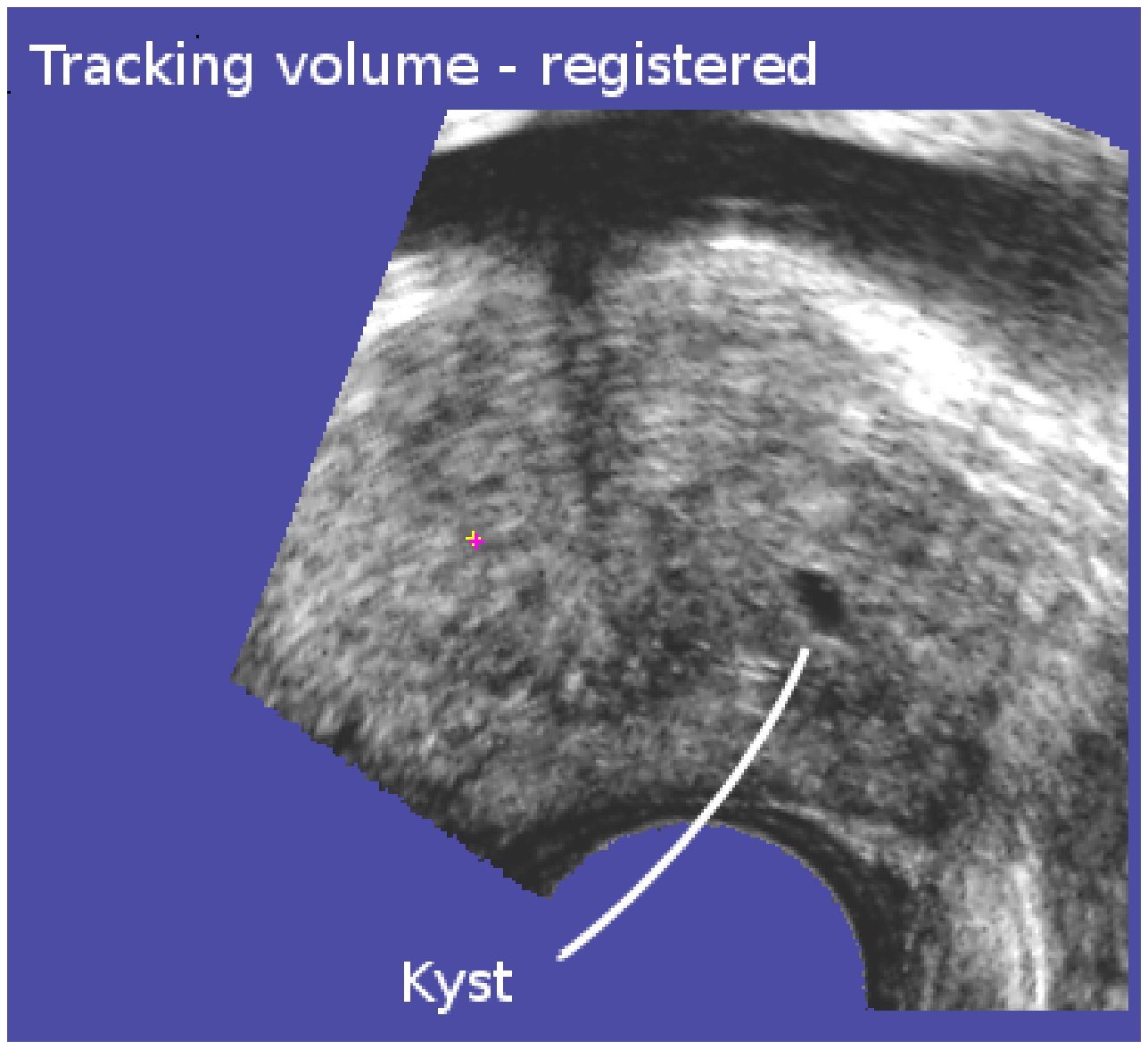}}
\caption{Fiducials. Fig.~(a) and (b) show corresponding calcifications in the tracking and the panorama volume, Fig.~(c) and (d) illustrate a kyst.}
\label{fig:fiducials}
\end{figure}

\begin{figure}
\centering
\includegraphics[width=0.3333333333333333333\textwidth]{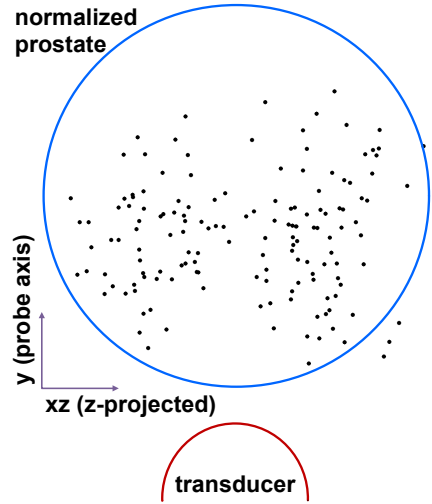}
\caption{Fiducial distribution. This figure shows the distribution of the 147 fiducials that were segmented in the panorama volumes in normalized space (unit circle). Normalization was performed by scaling the bounding ellipsoids on the unit sphere, followed by a projection of the fiducials onto the transverse mid-plane of the sphere.}
\label{fig:fiducialdist}
\end{figure}

This third study evaluates the accuracy of 687 registrations stemming from 40 patients that were classified as correct. Fiducials were first manually segmented in the panorama volumes, and then in the registered tracking volumes if they were visible. Finding unambiguously identifiable fiducials in the panorama and the majority of the tracking volumes was a challenging task. Segmentation consisted in the definition of the barycenter of a structure. Small structures like calcifications or kysts were preferred since their limited volume facilitates manual barycenter definition and hence reduces the fiducial localization error (FLE), see Fig.~\ref{fig:fiducials}. We did not segment (parts of) the prostatic capsule since it is difficult to identify anatomically corresponding points on surfaces in US images, i.e. the FLE of the point distance measure would have increased. Surface-distance measures on the other hand underestimate the tissue-correspondence error because they are insensitive to on-surface misalignments. In total, 147 reference fiducials were segmented in the 40 panorama volumes (3.7 fiducials per volume), and 1889 corresponding fiducials were segmented in the 687 tracking volumes (2.7 fiducials per volume). The statistical power of the simple descriptive statistics that were used in this study is ensured by the total number of 1889 evaluated samples, i.e. the small number of samples per registration pair is compensated by the large number of evaluated registrations. The FLE was estimated to 0.35$\pm$0.19 mm via multiple segmentations of the reference fiducials. The anatomical distribution of the reference fiducials is illustrated in Fig.~\ref{fig:fiducialdist}. It is important to note that the fiducial segmentations are not used by the registration algorithm; the fiducial registration error is hence a valid estimator of the target registration error (TRE).

To evaluate the TRE, the distance between the centers of corresponding fiducials were measured using the Euclidean distance
\begin{equation}
\epsilon(P, k)=||F_{I_0}^P(k)- \varphi^\ast[I_0, I_m]\circ F_{I_m}^P(k)||_{\mathbb{R}^3},
\end{equation}
where $P$ is the patient, $F_{I_0}^P(k)$ is the $k$th fiducial in the panorama image $I_0$ and $F_{I_m}^P(k)$ is the anatomically corresponding fiducial in the tracking image $I_m$.

The results for both rigid and elastic registration are given in Tab.~\ref{tab:accuracy}. Rigid registration yields a TRE of 1.4$\pm$0.8 mm while the TRE after deformation estimation is 0.8$\pm$0.5 mm. The mean error in the worst decile is 3.0~mm for rigid registration, which is reduced to 2.0~mm after deformation estimation. The average computation time of rigid registration is about 2~s versus about 7~s for deformation estimation. We also evaluated the performance of the deformation estimation without inverse consistency constraints (TRE of 1.0$\pm$0.7 mm) and when using standard SSD instead of the proposed intensity-shift compensating metric $\mathcal{D}_{SC}$ (TRE of 1.6$\pm$1.5 mm), everything else being identical. Both techniques improve thus accuracy. SSD-based registration yields results worse than rigid registration, which proves that this measure is inadequate for deformation estimation in US images. On the other hand, shift correlation significantly improves the rigid registration result while its computation time is still reasonable. It is hence an interesting alternative to more complex similarity measures.

\begin{table}
\centering
\begin{tabular}{l@{\hspace{10pt}}r@{\hspace{10pt}}|@{\hspace{10pt}}c@{\hspace{10pt}}c@{\hspace{10pt}}c@{\hspace{10pt}}c@{\hspace{10pt}}}
~&	                        & mean      & RMS  & mean dist. & execution \\
&		                      & distance  & distance & (worst decile) & time  \\
&                         & [mm] & [mm] & [mm] & [s]\\[3pt] \hline &&&&&\\[-7pt]
\footnotesize{1}& unregistered              & 13.8$\pm$7.9 & 17.0 & 17.9 & - \\
\footnotesize{2}& rigid                     & ~1.4$\pm$0.8 & 1.5 & ~3.0 & 2.1 \\
\footnotesize{3}& deformation               & ~0.8$\pm$0.5 & 0.9 & ~2.0 & 6.8 \\[3pt] \hline &&&&&\\[-7pt]
\footnotesize{4}& def. w/o inv. cons.       & ~1.0$\pm$0.7 & 1.2 & ~2.7 & 6.4 \\
\footnotesize{5}& def. w std SSD            & ~1.6$\pm$1.5 & 2.2 & ~5.2 & 4.3 \\
\end{tabular}
\caption{Accuracy study. Line 2) and 3) present TRE and registration time of the rigid and deformation estimations. Line 4) shows the result for the deformation estimation without inverse consistency constraints, and line 5) for the deformation estimation using standard SSD instead of shift correlation as similarity measure. The mean error in the `worst decile' was computed on the 10$\%$ of fiducials with the largest TRE.}
\label{tab:accuracy}
\end{table}

\subsection{Rigid vs. elastic registration}
\begin{figure}
\centering
\includegraphics[width=0.8\textwidth]{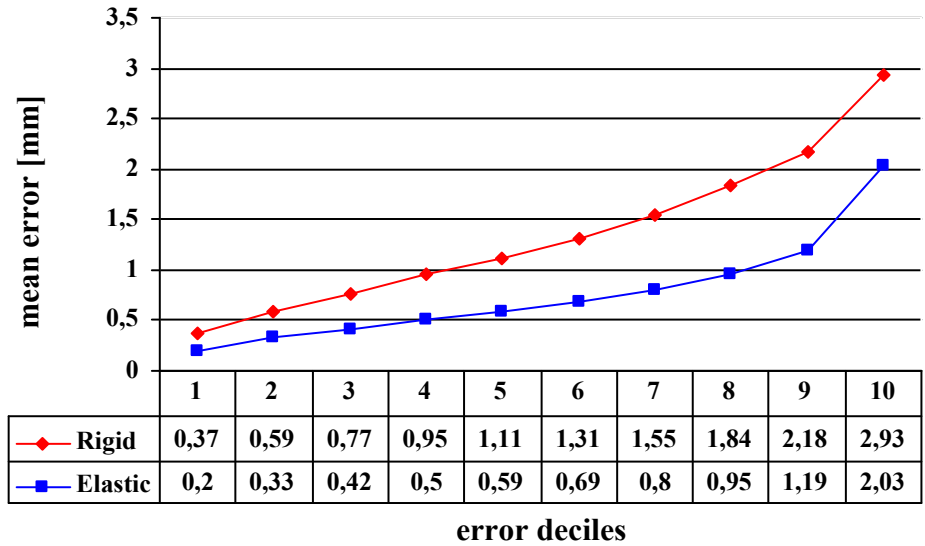}
\caption{Rigid vs. Elastic. The Fig. shows the error deciles after rigid and elastic registration.}
\label{fig:rigid-vs-elastic}
\end{figure}

\begin{figure}
\subfigure[]{\includegraphics[width=.25\textwidth]{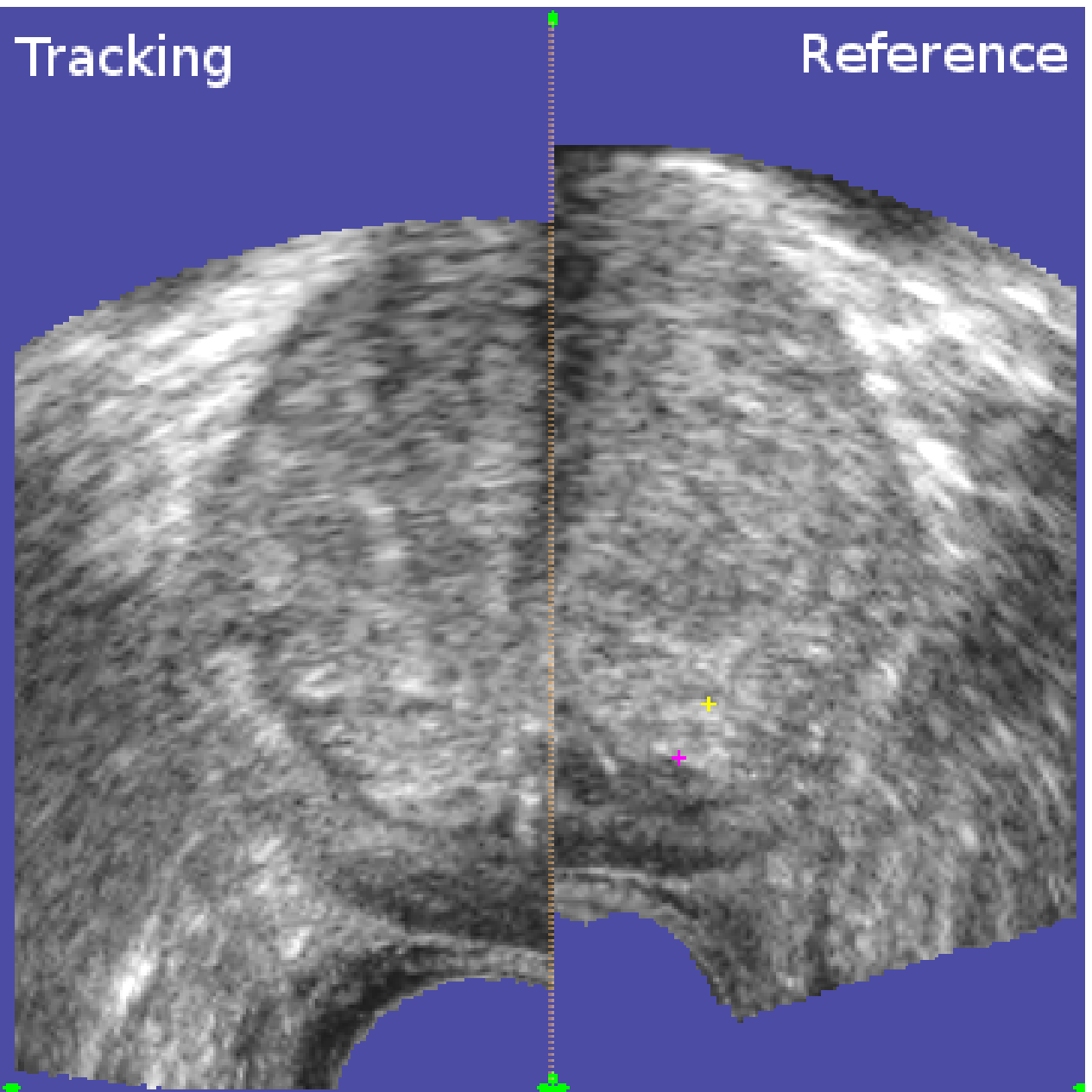}}\hfill
\subfigure[]{\includegraphics[width=.25\textwidth]{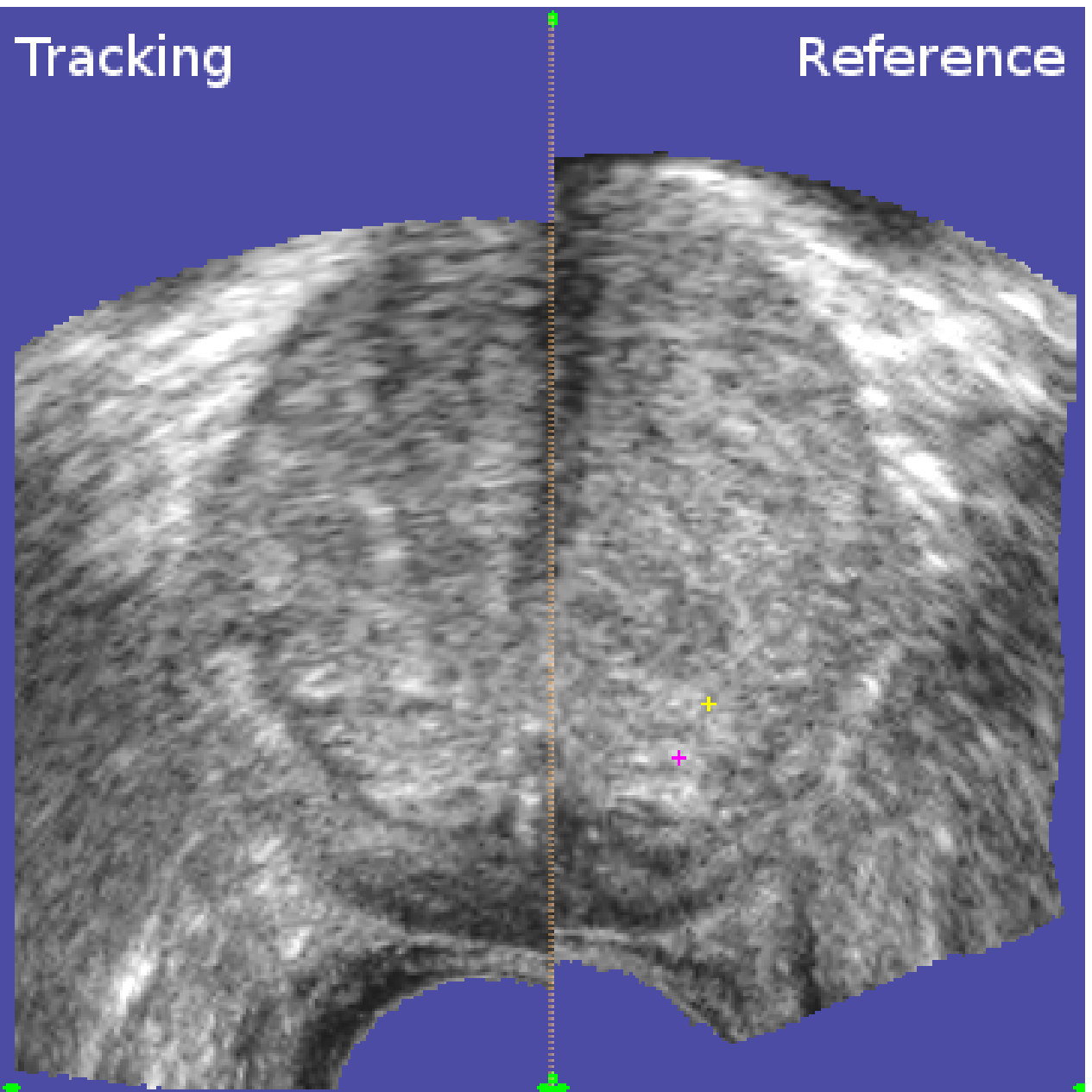}}\hfill
\subfigure[]{\includegraphics[width=.25\textwidth]{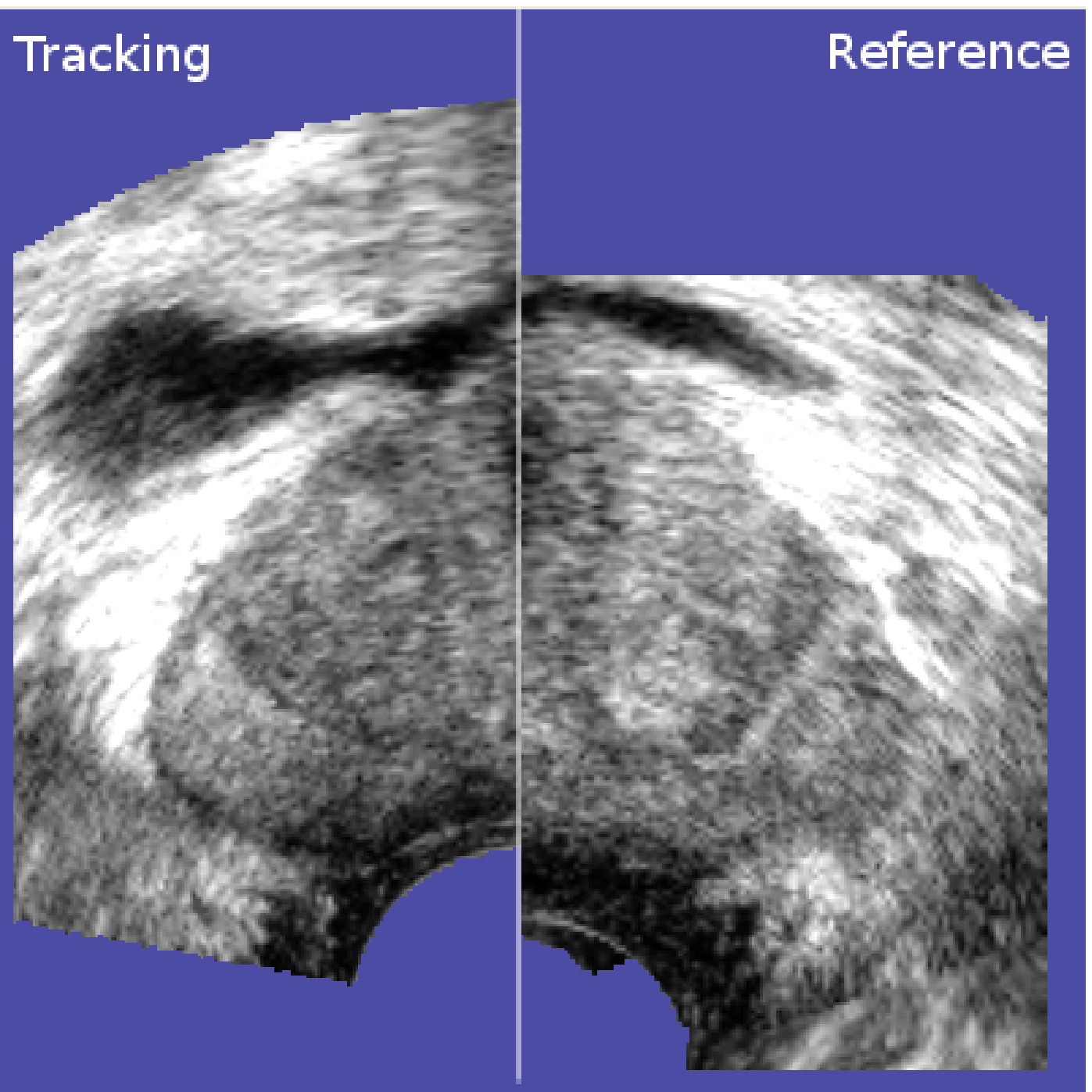}}\hfill
\subfigure[]{\includegraphics[width=.25\textwidth]{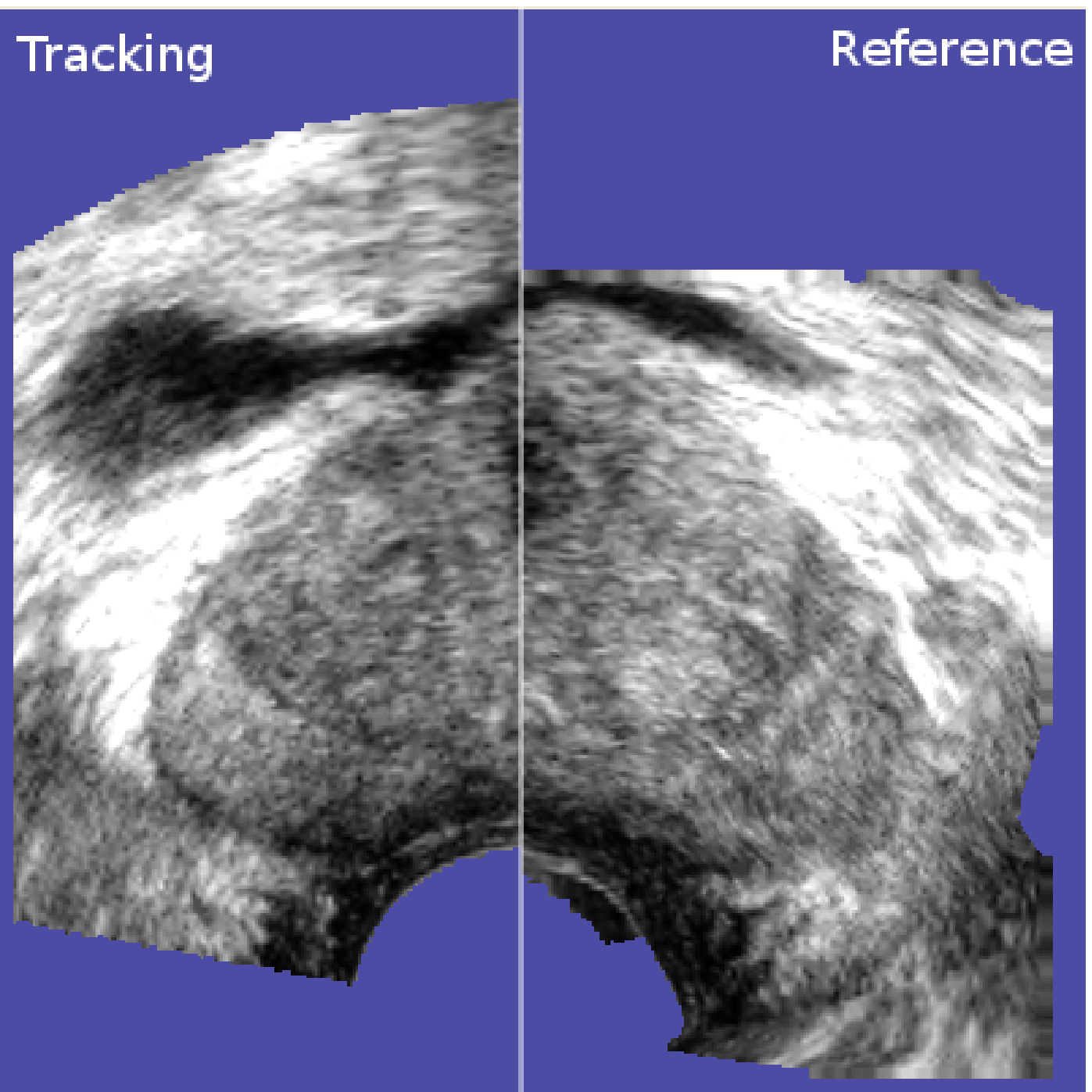}}
\caption{Rigid vs. Elastic. Fig.~(a) and (c) show typical rigid registration results, the left half of each image showing the tracking image, and the right half showing the reference. In both cases, different probe pressure was applied when the volumes were acquired. The non-linear compression cannot be corrected by the rigid registration and local mismatches subsist in particular near the probe head. Fig.~(b) and (d) show the result of the deformation estimation, which corrects the local errors.}
\label{fig:elasticresults}
\end{figure}

A comparison of the mean decile TRE of rigid and elastic registration was carried out to evaluate the clinical relevance of deformation estimation, see Fig.~\ref{fig:rigid-vs-elastic}. Elastic registration improves TRE in all deciles about at least one third. In absolute values, it yields the strongest benefit in the worst decile, where the average error is decreased by 1 mm to 2 mm. The error curve of rigid registration indicates that the gland is barely deformed in the majority of the volumes that were analyzed. However, in about 20$\%$ to 30$\%$ of the volumes stronger deformations can be observed that can be reduced with elastic registration. The cross correlation between the TRE and the fiducial distance from the transducer is -0.09 for rigid registration, i.e. these two variables are uncorrelated. The cross correlation between the TRE and the fiducial distance from the prostate center is 0.17 for rigid registration, i.e. the error at the capsule is of similar extent than the error at the center. The cross correlation between the TRE and the fiducial distance to the needle tip is 0.02, and between the TRE and the fiducial distance to the entire needle trajectory is 0.0. We therefore did not measure significantly stronger errors near the needle than elsewhere. This observation has to be, however, taken with care since it is possible that including all fiducials to measure a local phenomenon (the volume of the needle is very small compared to the volume of the prostate), together with the sparse number of fiducials per volume, might compromise the statistical power of this particular test. A visual illustration of the deformation estimation is given in Fig.~\ref{fig:elasticresults}.

\section{Discussion}
\label{sec:techdiscuss}

We have presented a 3D TRUS based tracking system for prostate motion that occurs during transrectal biopsy acquisition. The presented approach differs from existing systems in that it uses a motorized US transducer for 3D volume acquisition and in that
it is purely image-registration based. A kinematic model of the rectal probe motion is used to find the position of the probe with respect to the gland, and not in operating room coordinates, as it is the case with approaches that use probe tracking hardware. A registration framework capable of estimating the rigid and the residual elastic motion of the prostate was presented. A simple but effective local similarity measure was introduced to drive the elastic registration process as efficiently as possible. Special extrapolation techniques were proposed to cope with information loss problems in multi-resolution image registration. A clinical application was built on the top of the tracking system, allowing the computation of precise biopsy and cancer maps, MR-TRUS target projection, MR biopsy maps for therapy planning, and rudimentary support for guidance towards non-ultrasound targets. In this section, we will discuss the proposed system.

\subsection{Target registration error}
\label{sec:trediscussion}

A prostate tumor is considered as being clinically significant if it has a volume of at least 0.5 cc \cite{ahmed07nature,mozer09review}. If the shape of the tumor resembles a sphere, this corresponds to a radius of roughly 5~mm. Let us further assume that the error is normally distributed. Based on these assumptions, Karnik et al. claim that a tracking TRE of 2.5~mm (RMS) yields a probability of 95.4$\%$ that the registered targets will lie inside the clinically significant 5~mm radius \cite{karnik10medphys}. However, this is only valid if there are no other sources of error in the clinical application. 

If the clinician wants to be guided towards non-US targets, for example suspicious lesions in MR images, he has to segment and register them with the US reference volume. We can assume that MR target segmentation, MR-TRUS registration and TRUS-TRUS registration are statistically independent. In that case, the total RMS error is
\begin{equation}
\epsilon=\sqrt{\sum_i{\epsilon_i^2}}.
\end{equation}
If we assume that manual MR segmentation is performed with an RMS error of 1~mm, MR-TRUS registration with an error of 2.5~mm \cite{martin10isbi,martin10medphys} and TRUS-TRUS registration also with 2.5~mm, we get a total RMS error of 3.7~mm. This corresponds to a 82$\%$ probability to hit the target.

However, we also have needle deflection, guide calibration errors, needle depth tracking errors, tissue deformation during needle insertion and the error of manual reproduction of the proposed trajectory. Unfortunately we do not know the extent of these errors. Let us optimistically assume that each of these factors add an additional RMS error of 1 mm. Then we get a total RMS error of 4.3 mm or a probability of about 75$\%$ to hit the target. This means that the sensitivity of MR-TRUS guided biopsies would correspond approximately to the sensitivity of a systematic sampling of the gland. With this error we would thus not gain anything.

It is therefore necessary to minimize the TRE of each part of the processing chain wherever possible. Reusing the assumptions of the previous paragraphs, the total RMS error of the presented system is 3.6 mm (RMS, 83$\%$ probability to hit the target) with elastic registration, and 3.8 mm (RMS, 81$\%$ probability to hit the target) with rigid registration. With elastic registration, the probability of missing the target is reduced by 25$\%$ compared to a tracking system with a TRE of 2.5 mm (RMS). If the TRE of the presented system could be reduced to 0.25 mm (RMS), which corresponds to the error of optical or magnetic tracking systems, the probability to hit the target would increase to 85$\%$. In other words, further reducing the tracking error by a factor of three improves the hit rate by only 2.5$\%$. For MR targeting, future efforts should thus be concentrated on the reduction of the MR-TRUS fusion error. 

Note that the error analysis is not the same for biopsy maps, where the total error is composed only of the TRUS-TRUS registration error and the needle segmentation error. If we assume, again, that a manual segmentation can be performed with an RMS error of 1 mm, the total error would be 1.34 mm (RMS). If the biopsy map is used for therapy planning, however, and if it is projected onto the therapy planning volume, the projection error adds to the total error.

In conclusion, to achieve a clinically satisfying total TRE, all sources of error must be minimized further. With the presented system, the most important sources of error seem to be MR-TRUS fusion and the human error, i.e. the capacity of the operator to reproduce a suggested trajectory. Future efforts should hence be concentrated on the optimization of these errors.

\subsection{Rigid vs. elastic tracking}

Elastic tracking is the preferred solution concerning the minimization of the total TRE of the system. However, it requires a three to four fold increase in computation time to reduce the risk of missing an MR target, estimated with the assumptions in Sec.~\ref{sec:trediscussion}, by only about ten percent. Other sources of error are predominant in the system. For biopsy maps, where the computation time is not an issue, and for the registration of the panoramas of two different sessions, elastic registration is the preferred solution.

It is interesting to have a look at the error distribution of rigid and elastic registration (cf. Fig.~\ref{fig:rigid-vs-elastic}). Elastic registration reduces the rigid error by almost 50$\%$ in the deciles 1-9. In the worst decile, the error reduction is 30$\%$. The error is reduced significantly in all deciles, which shows that the proposed non-linear registration method is robust. The rigid registration result is not degraded.

\subsection{Patient motion}

A considerable advantage of the proposed approach is that it is relatively robust with respect to patient movements. Some patients feel pain during biopsy acquisition, in which case hip displacements due to muscular contractions can occur. Other patients are not comfortably positioned and change their pose during the procedure. In the presented experiments, registration failures were not due to patient movements, but to a lack of prostate information in the tracking volume, which is comparable to a marker that is out of view of an optical tracking camera or to a magnetic sensor that is outside the magnetic field. This contrasts with systems that use probe tracking to initialize image-based registration, for which the patient/prostate can move outside the capture range of the registration algorithm. If this happens, these systems have to acquire a new anatomical reference volume, which is time consuming and requires transitive registration of the previous results. A nice side-effect of a purely software-based and free-hand tracking system is that less encumbering hardware is required in the operating room.

Patient movements can, however, still pose problems with the presented system when they occur during 3D TRUS acquisition, which leads to distorted volumes. The same applies when the operator moves the probe during volume acquisition. Patient and probe movements can also occur during registration, which is problematic when the system guides the clinician toward a target. This can be improved by reducing the registration time, cf. also the discussion in Sec.~\ref{sec:3Dtrus-guidance}.

\subsection{3D TRUS vs. 2D TRUS based tracking}
\label{sec:3Dtrus-guidance}
3D TRUS and 2D TRUS based prostate biopsy systems have both strengths and weaknesses. A large inconvenience of 3D US and deformation estimation is the volume acquisition time of 0.5-5~s (depending on the image quality) and the registration time of currently 7-8~s. While this is still sufficient to give the clinician an intra-operative feed-back, it is difficult to implement a system that can guide the clinicians to targets. It is conceptually straight-forward to parallelize the registration algorithm on specialized hardware like modern scalar processing units. In this case, an increase in speed by a factor of 20-30 seems to be realistic. However, real-time TRUS volume acquisition is currently a blocking issue for conveniently fast updates of the tracking position.

2D TRUS based tracking is less accurate and more sensitive to patient movements. When using 2D TRUS with image-based registration, there is a risk that lateral biopsies will not be correctly registered since only a small part of the image will contain the gland. The capture range of such a system is smaller than the range of a 3D TRUS based system. Another inconvenience of 2D TRUS is the less accurate and more time-consuming acquisition of the initial 3D volume that serves as anatomical reference. The impact of the volume reconstruction error cannot be fully evaluated with point fiducial based TRE evaluations. Phantom studies have shown distance errors of 3$\%$-4$\%$ \cite{bax08biopsyguidance, cool06reconstruction}; on real patients the accuracy is unknown. Perhaps a combination of 2D and 3D TRUS tracking would yield the best of both worlds.

\section{Conclusion}
The biopsy application that we implemented for clinical use gives the clinician an immediate and precise feed-back about the current biopsy distribution. The biopsy maps can be fused with the results of the histological analysis for diagnostics and therapy planning. The maps of a previous biopsy session can be projected on the current panorama image. Finally, the system enables guidance towards targets segmented on MR images. The system only minimally alters the current standard procedure and does not add significant logistical complexity to the intervention.

The cancer map module provides significant clinical value for diagnosis and therapy planning. It makes it possible to implement experimental focal therapy strategies based on the localization of positive samples. It is for example feasible to perform additional targeted biopsies in a second session around a positive sample to get a very precise idea of the cancer distribution. Inversely it is possible to confirm the absence of cancer locally with supplementary targeted biopsies. The possibility to project the cancer map into an MR image for determination of the shape of a tumor in a more sensitive imaging modality is also a very promising feature for prostate cancer diagnostic.

Guidance to predefined targets could increase the sensitivity of prostate biopsies. Suspicious lesions identified in MR volumes can be projected into the tracking images and hence enable them to be reached with high precision. If a biopsy series needs to be repeated, the clinician can avoid sampling previously examined tissues again, which leads to a higher coverage of the gland. A limitation of the system is, however, that it is not yet possible to perform guidance to targets in real-time. 

To date, the industrial version of the presented biopsy assistance system has been used on more than 200 biopsy sessions and has proven to work reliably in clinical practice. We are currently putting together the protocols for in-depth assessment of the clinical potential of the system. We are particularly interested in the impact of the system on the sensitivity of US guided prostate biopsies and biopsy repetitions. A preliminary study \cite{mozer09learning} showed that it is difficult for clinicians to reach the zonal targets of the clinical 12-core standard protocol accurately. Giving the clinician a visual feed-back about the sample distribution after the intervention lead already to a significant learning effect. We are also interested in the sensitivity and specificity of MR-targeted biopsies, which is an interesting option when the initial biopsy series was negative while the cancer suspicion could not be discarded.

In the longer term, it would be interesting to consider the usage of the prostate tracking system for therapy, knowing that a number of therapeutic interventions on the prostate are performed under endorectal US control (e.g. brachytherapy, HIFU, cryotherapy, ...). Focal therapy has the potential to lead to less side-effects than radical therapy, but it requires the accurate positioning of the therapeutic instrument to be safe, reliable and effective. US based prostate tracking can help to achieve this accuracy.
\section{Acknowledgements}

We would like to thank Antoine Leroy, PhD, KOELIS SAS, Grenoble, France, for assisting us in the coordination of our clinical experiments and for funding. We would also like to thank Grégoire Chevreau, MD, and Stéphane Bart, MD, and the entire urology department, led by Pr. Marc Olivier Bitker, of the Pitié Salpétrière hospital, Paris, France, for their participation in the biopsy volume acquisition project. We would furthermore like to thank Sébastien Martin, PhD, TIMC laboratory, Grenoble, France, for making his statistical prostate shape and deformation priors available for the segmentation module of the application prototype. Thanks also to Kévin Dowlut for his work on automatic needle detection and automatic registration validation. Final thanks go to the French Agence Nationale de la Recherche (TecSan grant), the French Ministry of Health (PHRC grant), the French Association Nationale Recherche Technologie (CIFRE grant) and to KOELIS SAS, Grenoble, France (CIFRE grant) for funding.



\bibliographystyle{elsarticle-harv}
\bibliography{tracking}



\end{document}